\documentclass[3p]{elsarticle}

\usepackage{lineno,hyperref}
\modulolinenumbers[5]

\usepackage{times}  
\usepackage{helvet} 
\usepackage{courier}  
\usepackage{graphicx} 

\usepackage{latexsym}

\usepackage{subfig}

\usepackage{abraces}

\usepackage[stable]{footmisc}

\usepackage{soul}
\usepackage[utf8]{inputenc}
\usepackage{amsmath}
\usepackage{booktabs}
\usepackage{amsthm}
\usepackage{amsfonts}
\usepackage[T1]{fontenc}

\usepackage{quoting}

\usepackage{fancybox}
\usepackage{amsfonts}

\usepackage{environ}
\usepackage{mathnotation}
\usepackage{multirow}

\usepackage{color}

\usepackage{xspace}
\usepackage{dashrule}
\usepackage[table,xcdraw,dvipsnames]{xcolor}
\usepackage[inline]{enumitem}

\usepackage{footmisc}

\newcommand{\dpi}{\mathit{dpi}}

\newcommand{\dt}{\mathcal{D}^*}
\newcommand{\md}{\mathcal{D}}
\newcommand{\mD}{\mathbf{D}}
\newcommand{\mC}{\mathbf{C}}

\newcommand{\ld}{\mathit{ld}}
\newcommand{\mo}{\mathcal{K}}
\newcommand{\mb}{\mathcal{B}}
\newcommand{\tax}{\mathit{ax}}
\newcommand{\Tp}{\mathit{P}}
\newcommand{\Tn}{\mathit{N}}
\newcommand{\tp}{\mathit{p}}
\newcommand{\tn}{\mathit{n}}
\newcommand{\mc}{\mathcal{C}}

\newcommand{\Queue}{{\mathbf{Q}}}

\newcommand{\mQ}{{\bf{Q}}}

\newcommand{\pr}{\mathit{pr}}
\newcommand{\bnd}{\mathit{bound}}
\newcommand{\node}{\mathsf{n}}
\newcommand{\valid}{\mathit{valid}}
\newcommand{\closed}{\mathit{closed}}
\newcommand{\childnodes}{\mathsf{Child\_Nodes}}

\newcommand{\sucs}{\mathsf{succ}}
\newcommand{\goal}{\mathsf{goal}}
\newcommand{\cost}{g}

\newcommand{\ppi}{\mathit{ppi}}

\usepackage{algorithm}
\usepackage[noend]{algpseudocode}
\algrenewcommand\algorithmicrequire{\textbf{Input:}}
\algrenewcommand\algorithmicensure{\textbf{Output:}}
\algrenewcommand\alglinenumber[1]{\tiny #1:} 
\algnewcommand{\IfThen}[2]{
	\State \algorithmicif\ #1\ \algorithmicthen\ #2
}

\usepackage[all]{xy}
\usepackage{tikz}

\newcounter{examplecounter}
\newenvironment{example}{
	\refstepcounter{examplecounter}%
	
	\vspace{7pt}
	\noindent\textbf{Example \arabic{examplecounter}}%
	\quad
}{
	
	\vspace{7pt}
}

\newtheorem{definition}{Definition}
\newtheorem{thm}{Theorem}
\newtheorem{lemma}{Lemma}

\makeatletter
\g@addto@macro\@floatboxreset\centering
\makeatother

\journal{ArXiv}

\begin{document}

\begin{frontmatter}


\title{RBF-HS: Recursive Best-First Hitting Set Search\tnoteref{t1}}
\tnotetext[t1]{This work is a technical report underlying the published paper ``Memory-Limited Model-Based Diagnosis'' \cite{rodler2022_memorylimited}. Earlier and 
	significantly 
	shorter versions of this work were accepted 
	at the \emph{Int'l Workshop on Principles of Diagnosis (DX-2020)} \protect\cite{rodler2020_rbfhs_dx} and at the \emph{Annual Symp.\ on Combinatorial Search (SoCS-2021)} \protect\cite{rodler2021_socs}.}

\author{Patrick Rodler}
\address{University of Klagenfurt, 9020 Klagenfurt, Austria\\
		patrick.rodler@aau.at}
\begin{abstract}
\small
Various model-based diagnosis scenarios
require the computation of the \emph{most preferred} fault explanations. Existing algorithms that are \emph{sound} (i.e., output only actual fault explanations) and \emph{complete} (i.e., can return all explanations), however, require exponential space to achieve this task. 
As a remedy, and to enable successful diagnosis both on memory-restricted devices and for memory-intensive problem cases, we propose two novel diagnostic search algorithms which build upon tried and tested techniques from the heuristic search domain. The first method, dubbed Recursive Best-First Hitting Set Search (RBF-HS), is based on Korf's well-known Recursive Best-First Search (RBFS) algorithm. We show that RBF-HS can enumerate an arbitrary predefined finite number of fault explanations in best-first order within linear space bounds, without sacrificing the desirable soundness or completeness properties. The second algorithm, called Hybrid Best-First Hitting Set Search (HBF-HS), is a hybrid between RBF-HS and Reiter's seminal HS-Tree. The idea is to find a trade-off between runtime optimization and a restricted space consumption that does not exceed the available memory. Notably, both suggested algorithms are \emph{generally applicable} to any model-based diagnosis problem, regardless of the used (monotonic) logical language to describe the diagnosed system and of the used reasoning mechanism. 

We ran comprehensive experiments on real-world benchmarks from the knowledge-based systems field, a domain where the features soundness, completeness, the best-first property as well as a general applicability are pivotal and where Reiter's HS-Tree is the predominantly used diagnostic search. The evaluation reveals that, when computing fault explanations minimal-cardinality-first, RBF-HS compared to HS-Tree reduces memory requirements substantially in most cases by up to several orders of magnitude, while 
also 
saving runtime in more than a third of the cases. When computing fault explanations most-probable-first, 
RBF-HS compared to HS-Tree tends to trade memory savings more or less one-to-one for runtime overheads. Whenever runtime overheads were significant, using HBF-HS instead of RBF-HS reduced the runtime to values comparable with HS-Tree while keeping the used memory reasonably bounded. 

Finally, the presented approaches are not restricted to diagnosis problems, but applicable to best-first hitting set computation in general, and thus 
have the potential to impact 
research and application domains beyond the frontiers of model-based diagnosis.
\end{abstract}


\end{frontmatter}


\section{Introduction}
\label{sec:intro} 
\emph{Model-based diagnosis} \cite{Reiter87,dekleer1987} is a popular, well-understood and domain-independent paradigm that has over the last decades found widespread adoption for troubleshooting systems as different as programs, circuits, physical devices, knowledge bases, spreadsheets, production plans, robots, vehicles, or aircrafts \cite{wotawa2010fault,DBLP:journals/jair/FeldmanPG10a,steinbauer2005detecting,Rodler2015phd,jannach2010toward,sachenbacher1998electrics,ng1990model,gorinevsky2002model,rodler_teppan2020,DBLP:conf/ieaaie/FelfernigMMST09}.
The principle behind model-based diagnosis is to model the system to be diagnosed by means of a logical knowledge representation language. 
Beside general knowledge 
about the system, this \emph{system description} includes a characterization of the normal behavior of all \emph{system components} relevant to the diagnosis task. 
Logical theorem provers can then be used to verify if the predicted system behavior---deduced from the system description under the assumption that all components work normally---is consistent with factual evidence (\emph{observations}) about the real system behavior. 
In case of an inconsistency, the goal is to find the abnormal components responsible for the observed system misbehavior. An (irreducible) set of components 
whose assumed abnormality 
makes the system description consistent with the observations is called a \emph{(minimal) diagnosis}. Typically, 
there are multiple minimal diagnoses for practical diagnosis problems, and it is an important issue to isolate the \emph{actual diagnosis}, which pinpoints the actually faulty components, from other spurious candidates.

Over the years, various diagnosis search methods have been suggested, e.g.,  \cite{jannach2016parallel,wotawa2001variant,lin2003computation,greiner1989correction,Shchekotykhin2014,rodler2018statichs,feldman2006two}. Motivated by different diagnosis scenarios and application fields, these algorithms feature greatly different properties. For instance, while some are designed to guarantee \emph{soundness} and \emph{completeness} (i.e., the computation of \emph{only} and \emph{all} minimal diagnoses), e.g., to ensure the localization of the actual diagnosis in critical applications (medicine \cite{rector2011getting}, aircrafts \cite{gorinevsky2002model}, etc.), others drop one or both of these properties, e.g., to allow for higher diagnostic efficiency \cite{Feldman2008,abreu2009low}. 
Since the computation of all (minimal) diagnoses is NP-hard \cite{Bylander1991},
all diagnosis searches have to focus on a (computationally feasible) subset of the diagnoses in general. This subset is commonly referred to as the \emph{leading diagnoses} \cite{dekleer1991focusing_prob_diag}, and usually defined as the best minimal diagnoses according to some preference criterion such as minimal cardinality or maximal probability. 
Algorithms which enumerate diagnoses in order of preference are called \emph{best-first}. 
One of the 
most general sound, complete and best-first algorithms in literature is Reiter's seminal HS-Tree \cite{Reiter87,Rodler2015phd,greiner1989correction}, because it is independent of the used (monotonic)
system description language and of the used theorem prover. The advantage of this generality is a \emph{broad and flexible applicability} of the search algorithm over a wide range of diagnosis application domains. For example, in the field of knowledge base or ontology debugging, diagnosers have to deal with a myriad of different logics that are used to model and solve problems in various domains while achieving a trade-off between inference complexity and logical expressivity.
The development of different (suitably adapted) diagnostic search techniques for all these cases would be hardly realizable. General algorithms like HS-Tree, on the other hand, can work with any of these logics and related theorem provers out of the box.

Traditional (sound and complete) best-first diagnosis search methods require an exponential amount of memory. The reason is that all paths in a search tree must be stored in order to guarantee that the best one is expanded in each iteration. 
This can prevent the application of best-first searches to a range of model-based diagnosis scenarios which, e.g., \emph{(a)}~pose substantial memory requirements on the diagnostic methods, or \emph{(b)}~suffer from too little memory. 
One example for (a) are problems involving high-cardinality diagnoses, e.g., when two systems are integrated and a multitude of errors emerge at once
\cite{Shchekotykhin2014,meilicke2011alignment}. Manifestations of (b) are frequently found in today's era of the Internet of Things (IoT), distributed or autonomous systems, and ubiquitous computing, where low-end microprocessors, often with only a small amount of RAM, are incorporated into almost any device. Whenever such devices should perform (self-)diagnosing actions \cite{klein1999evaluating,williams1996model}, memory-limited diagnosis algorithms are a must \cite{williams2007conflict,zoeteweij2008automated}.

As a remedy, we introduce 
two general diagnostic search algorithms that require either linear or {(quasi-)}restricted
 memory while featuring all above-mentio-ned desirable search properties. In particular, our \textbf{contributions} are: 
\begin{itemize}[noitemsep]
	\item We propose \emph{Recursive Best-First Hitting Set Search (RBF-HS)}, a novel diagnostic search drawing on ideas used by Korf in his well-known \emph{Recursive Best-First Search (RBFS)} algorithm \cite{korf1992linear}.
	\item We show that RBF-HS can compute an arbitrary predefined finite number of minimal diagnoses in a sound, complete and best-first way within linear memory bounds, and that it can be generally applied to arbitrary diagnosis problems as per Reiter's theory of model-based diagnosis \cite{Reiter87}.
	\item We generalize RBF-HS, which acts on the maxim to use as little memory as possible, by integrating it with HS-Tree to a hybrid search method that remains sound, complete and best-first. The basic rationale behind this search, dubbed \emph{Hybrid Best-First Hitting Set Search (HBF-HS)}, is to initially run 
	HS-Tree as long as sufficient memory is still available (\emph{optimize time}), and to then switch to RBF-HS to minimize the additional used memory (\emph{optimize space}) in order to avoid running out of memory and preserve problem solvability. 
\end{itemize}
Beside thorough theoretical complexity and correctness analyses, we present extensive empirical evaluations of the proposed techniques on real-world diagnosis cases where we demonstrate the broad applicability of our approaches on problems formulated in various logics with high expressivities and hard reasoning complexities beyond NP-complete. The \textbf{main experiment results} are: 
\begin{itemize}[noitemsep]
	\item \emph{Minimal cardinality first:} When computing minimal diagnoses in ascending order of cardinality, we find that RBF-HS, compared to HS-Tree, \emph{(1)}~exhibits significant memory savings as opposed to no more than marginal runtime losses in most cases, 
	where savings increase with increasing problem complexity,
	\emph{(2)}~saves \emph{both} memory \emph{and} runtime in more than a third of the cases, 
	\emph{(3)}~scales to large numbers 
	of computed leading diagnoses and to problems involving high-cardi-nality minimal diagnoses, and 
	\emph{(4)}~in the rare cases where runtime overhead was significant, using HBF-HS instead of RBF-HS reduced the runtime to values comparable with HS-Tree while keeping the used memory reasonably bounded.     
	\item \emph{Maximal probability first:}  When computing minimal diagnoses in descending order of probability, we find that RBF-HS tends to trade memory savings more or less one-to-one for runtime overheads (which has well-understood theoretical reasons that we discuss). Again, HBF-HS turns out to be a reasonable remedy to cut down the runtime while complying with practicable memory bounds. 
\end{itemize}
The \textbf{organization of the paper} is as follows. 
We repeat fundamental concepts from the fields of model-based diagnosis and heuristic search in Sec.~\ref{sec:basics}.  
The RBF-HS algorithm is introduced and discussed in Sec.~\ref{sec:algo}, where we 
use a didactic approach which builds up RBF-HS from RBFS in a stepwise manner.
In Sec.~\ref{sec:HBF-HS} we present and describe the HBF-HS algorithm. 
We comment on related works in Sec.~\ref{sec:related}.
Finally, Sec.~\ref{sec:eval} presents our experiments and reviews the obtained results, whereas concluding remarks and pointers to future work are given in Sec.~\ref{sec:conclusion}.

\section{Preliminaries}
\label{sec:basics}
First, we briefly characterize 
model-based diagnosis concepts used throughout this work, based on the framework of \cite{Rodler2015phd,Shchekotykhin2012} which is (slightly) more general \cite{rodler17dx_reducing} than Reiter's theory \cite{Reiter87}. 
The main reason for using this more general framework is its ability \cite{rodler17dx_reducing} to capture \emph{both} classical model-based diagnosis problems involving, e.g., malfunctioning circuits or physical systems, \emph{and} alternative problem types such as faulty knowledge bases which require the expression of negative measurements (things that must \emph{not} be true for the diagnosed system) \cite{Shchekotykhin2012,DBLP:journals/ai/FelfernigFJS04,schekotihin2018protege}.
The quality assurance of knowledge-based systems is an important application domain of the algorithms presented in this work \cite{Rodler2015phd,Shchekotykhin2014,meilicke2011alignment,Horridge2011a,Kalyanpur2006a,schlobach2007debugging} and also the focus of our evaluations; however, the proposed algorithms are generally applicable to any model-based diagnosis problem. 
%
Second, we concisely review important notions from heuristic search and contrast classic path-finding with diagnosis search problems. 
This comparison should serve to facilitate the understanding of the development of the diagnosis computation procedure RBF-HS starting from the path-finding algorithm RBFS presented in Sec.~\ref{sec:algo}.

\subsection{Model-Based Diagnosis}
\label{sec:MBD}
\subsubsection{Diagnosis Problem}
\label{sec:diagnosis_problem}
We assume that the diagnosed system, consisting of a set of components $\{c_1,\dots$, $c_k\}$, is described by a finite set of logical sentences $\mo \cup \mb$, where $\mo$ (possibly faulty sentences) includes knowledge about the behavior of the system components, and $\mb$ (correct background knowledge) comprises any additional available system knowledge and system observations. More precisely, there is a one-to-one relationship between axioms
$\tax_i \in \mo$ and components $c_i$, where $\tax_i$ describes (only) the normal behavior of $c_i$ (\emph{weak fault model} \cite{feldman2008computing}).
E.g., if $c_i$ is an AND-gate in a circuit, then $\tax_i := out(c_i) = and(in1(c_i),in2(c_i))$; $\mb$ in this case might contain sentences stating, e.g., which components are connected by wires, or observed circuit outputs. 
The inclusion of a sentence $\tax_i$ in $\mo$ corresponds to the (implicit) assumption that $c_i$ is healthy. Evidence about the system behavior is captured by sets of positive ($\Tp$) and negative ($\Tn$) measurements \cite{Reiter87,dekleer1987,DBLP:journals/ai/FelfernigFJS04}. Each measurement is a logical sentence; positive ones $\tp\in\Tp$ must be true and negative ones $\tn\in\Tn$ must not be true. The former can be, depending on the context, e.g., observations about the system, probes or required system properties. 
The latter model properties that must not hold for the system, e.g., if $\mo$ is a 
knowledge base to be debugged, a negative test case might be ``every bird can fly'' (think of penguins).
We call $\tuple{\mo,\mb,\Tp,\Tn}$ a \emph{diagnosis problem instance (DPI)}. 


\begin{example} \hspace{-1em}\emph{(Diagnosis Problem)}\quad\label{ex:dpi}
Assume a DPI stated in propositional logic with $\mo := \{\tax_1: A \to \lnot B, \tax_2: A \to B, \tax_3: A \to \lnot C, \tax_4: B \to C, \tax_5: A \to B \lor C \}$.
The ``system'' (the knowledge base $\mo$ itself in this case) comprises five ``components'' $c_1, \dots,c_5$, and the ``normal behavior'' of $c_i$ is given by the respective sentence $\tax_i \in \mo$. No background knowledge ($\mb = \emptyset$) or positive measurements ($\Tp=\emptyset$) are given from the start. 
But, there is one negative measurement ($\Tn = \setof{\lnot A}$), which stipulates that $\lnot A$ must \emph{not} be an entailment of the correct system (knowledge base). Note, however, that $\mo$ (i.e., the assumption that all ``components'' are normal) in this case does entail $\lnot A$ (e.g., due to the sentences $\tax_1,\tax_2$) and thus some sentence (``component'') in $\mo$ 
must be faulty. \qed
\end{example}


\subsubsection{Diagnoses}
\label{sec:diagnoses}
Given that the system description along with the positive measurements (under the 
assumption $\mo$ that all components are healthy) is inconsistent, i.e., $\mo \cup \mb \cup \Tp \models \bot$, or some negative measurement is entailed, i.e., $\mo \cup \mb \cup \Tp \models \tn$ for some $\tn \in \Tn$, some assumption(s) about the normality of components, i.e., some sentences in $\mo$, must be retracted. We call such a set of sentences $\md \subseteq \mo$ a \emph{diagnosis} for the DPI $\tuple{\mo,\mb,\Tp,\Tn}$ iff $(\mo \setminus \md) \cup \mb \cup \Tp \not\models x$ for all $x \in \Tn \cup \setof{\bot}$. We say that a diagnosis $\md$ is a \emph{minimal diagnosis} for $\dpi$ iff there is no diagnosis $\md' \subset \md$ for $\dpi$. 
Moreover, we call a diagnosis $\md$ a \emph{minimum-cardinality diagnosis} for 
$\dpi$ iff there is no diagnosis $\md'$ with $|\md'|<|\md|$ for $\dpi$.
The set of minimal diagnoses is representative of all diagnoses under the weak fault model \cite{dekleer1992characterizing}, i.e., 
the set of all diagnoses is equal to 
the set of all supersets of minimal diagnoses.
Therefore, diagnosis approaches often 
restrict their focus to only minimal diagnoses. 
We furthermore denote by $\dt$ the (unknown) \emph{actual diagnosis} which pinpoints the actually faulty axioms, i.e., all elements of $\dt$ are in fact faulty and all elements of $\mo\setminus\dt$ are in fact correct.

\begin{example} \hspace{-1em}\emph{(Diagnoses)}\quad\label{ex:diagnoses}
	For our DPI from Example~\ref{ex:dpi} there are four minimal diagnoses, given by $\md_1:=[\tax_1,\tax_3]$, $\md_2:=[\tax_1,\tax_4]$, $\md_3:=[\tax_2,\tax_3]$, and $\md_4 := [\tax_2,\tax_5]$ (we will always denote diagnoses by square brackets).
	For instance, $\md_1$ is a 
	diagnosis as $(\mo\setminus\md_1) \cup \mb\cup \Tp = \setof{\tax_2,\tax_4,\tax_5}$ is both consistent and does not entail the given negative measurement $\lnot A$.
	That $\md_1$ is a \emph{minimal} diagnosis as well, can be seen by observing that $(\mo\setminus\md'_1) \cup \mb\cup \Tp \models \lnot A$ for any $\md'_1 \subset \md_1$, i.e., the diagnosis property is violated after removing any element from $\md_1$.
	\qed
\end{example}

\subsubsection{Diagnosis Probability Model}
\label{sec:probability_model}
\paragraph{Component and Diagnosis Probabilities} In case useful meta information is available that allows to assess the likeliness of failure for system components, the probability of diagnoses (of being the actual diagnosis) can be derived.
Specifically, given a function $\pr$ that maps each sentence (system component) $\tax \in \mo$ to its failure probability $0<\pr(\tax)<1$, the probability $\pr(X)$ of a diagnosis (candidate)
$X \subseteq \mo$ under the common assumption of independent component failure is computed \cite{dekleer1987} as the probability that all sentences in $X$ are faulty, and all others are correct, i.e., 
\begin{align}
	\pr(X) := \prod_{\tax \in X} \pr(\tax) \prod_{\tax \in \mo\setminus X} (1-\pr(\tax)) \label{eq:diag_prob}
\end{align}

\paragraph{Properties of the Probability Function} 
We call $\pr$ \emph{strictly antimonotonic} iff 
$\pr(X) > \pr(Y)$
whenever $X \subset Y$.
%
Clearly, if $\pr$ is strictly antimonotonic, then each minimal diagnosis $\md$ has a higher probability $\pr(\md)$ than any non-minimal diagnosis $\md' \supset \md$. 
That is, assuming a list that includes all subsets of $\mo$ sorted by $\pr$ in descending order, then iterating over this list 
implies that \emph{(i)}~diagnoses with higher probability are found earlier, and \emph{(ii)}~a non-minimal diagnosis can never be encountered before all minimal diagnoses that are subsets of it have been visited.
Properties (i) and (ii) are material for search-based diagnosis computation methods, like well known existing ones \cite{Reiter87,Rodler2015phd,greiner1989correction,rodler2018statichs,rodler2020ecai} and those discussed in this work, which are based on the systematic exploration of (relevant parts of) the subset space of $\mo$, and which aim at finding all and only \emph{minimal} diagnoses (cf.\ Sec.~\ref{sec:diagnoses}) in the order from high to low probability. Thus, such approaches usually rely on the strict antimonotonicity of 
$\pr$. 

For a probability function $\pr$ to be strictly antimonotonic it is sufficient that $\pr(\tax)$  $< 0.5$ for all $\tax \in \mo$. This can be easily seen from Eq.~\ref{eq:diag_prob}, where $\pr(X') < \pr(X)$ for $X' \supset X$ under this assumption since $\pr(X') = \pr(X) \prod_{\tax \in X'\setminus X} \pr(\tax)/(1-\pr(\tax))$ and each factor $\pr(\tax)/(1-\pr(\tax)) < 1$ (see also \cite[Lemma 4.14]{Rodler2015phd}).
Note that 
diagnosis applications usually involve components which are a-priori much more likely to be normal than at fault, 
cf., e.g., \cite{dekleer1987,Rodler2013,dekleer2008framework,dekleer1990usingcrude,mengshoel2010probabilistic}. Hence, strict antimonotonicity of $\pr$ will in most cases be satisfied by default. Moreover, an arbitrary function $\pr$ can be transformed to a strictly antimonotonic function $\pr'$ by choosing a fixed $c \in (0,0.5)$ and by setting $\pr'(\tax) := c \cdot\pr(\tax)$ for all $\tax \in \mo$. Observe that this transformation does not affect the relative probabilities in that $\pr'(\tax_i)/\pr'(\tax_j) = k$ whenever $\pr(\tax_i)/\pr(\tax_j) = k$, i.e., no information is lost in the sense that the mutual fault probability order and ratio between any two components will remain invariant.  

\begin{example} \hspace{-1em}\emph{(Diagnosis Probabilities)}\quad\label{ex:diag_probs}
	Reconsider the DPI from Example~\ref{ex:dpi}
	and let the fault probabilities $\langle \pr(\tax_1)$, $\dots, \pr(\tax_5)\rangle = \langle .1, .05, .1, .05, .15\rangle$. Note, since all probabilities are smaller than $.5$, we have that $\pr$ is strictly antimonotonic. 
	The probabilities of all minimal diagnoses from Example~\ref{ex:diagnoses} can be computed as $\langle \pr(\md_1), \dots,$ $\pr(\md_4)\rangle = \langle .0077,.0036,.0036,.0058\rangle$. For instance, $\pr(\md_1)$ is calculated as $0.1*(1-0.05)*0.1*(1-0.05)*(1-0.15)$. The normalized diagnosis probabilities would then be $\langle .37,.175,.175,.28\rangle$. Note, this normalization makes sense if not all diagnoses, but only \emph{minimal} diagnoses are of interest, which is usually the case in model-based diagnosis applications for complexity reasons.\qed 
\end{example}

\subsubsection{Conflicts}
\label{sec:conflicts}
Instrumental for diagnosis computation 
is the notion of a conflict \cite{Reiter87,dekleer1987}.
A conflict is a set of healthiness assumptions for components $c_i$ that cannot all hold given the current knowledge about the system. More formally, $\mc \subseteq \mo$ is a \emph{conflict} for the DPI $\tuple{\mo,\mb,\Tp,\Tn}$ iff $\mc \cup \mb \cup \Tp \models x$ for some $x \in \Tn \cup \setof{\bot}$. We call a conflict $\mc$ a \emph{minimal conflict} for $\dpi$ iff there is no conflict $\mc' \subset \mc$ for $\dpi$.

%


\begin{example} \hspace{-1em}\emph{(Conflicts)}\quad\label{ex:conflicts}
	For our $\dpi$ from Example~\ref{ex:dpi} there are four minimal conflicts, given by $\mc_1 := \langle\tax_1,\tax_2\rangle$, $\mc_2 := \langle\tax_2,\tax_3,\tax_4\rangle$, $\mc_3 := \langle\tax_1,\tax_3,\tax_5\rangle$, and $\mc_4 := \langle\tax_3,\tax_4,\tax_5\rangle$ (we will always denote conflicts by angle brackets).
	For instance, $\mc_4$, in CNF equal to $(\lnot A \lor \lnot C) \land (\lnot B \lor C) \land (\lnot A \lor B \lor C)$, is a conflict because adding the unit clause $(A)$ to this CNF yields a contradiction, which is why the negative test case $\lnot A$ is an entailment of $\mc_4$. The minimality of the conflict $\mc_4$ can be verified by rotationally removing from $\mc_4$ a single axiom at the time and controlling for each so obtained subset that this subset is consistent and does not entail $\lnot A$.\qed
\end{example}

\subsubsection{Conflict Computation}
Literature offers a variety of algorithms for conflict computation, e.g., \cite{Horridge2011a,Kalyanpur2006a,gaber2020computation,gregoire2007boosting,lagniez2013factoring,manthey2016efficient,liffiton2013enumerating,junker04,rodler2020qx,marques2013minimal,shchekotykhin2015mergexplain}. 
Among those, 
we are in this work mainly interested in so-called \emph{black-box} \cite{parsia2005debugging} techniques,
such as \textsc{QuickXplain} \cite{junker04,rodler2020qx} or \textsc{Progression} \cite{marques2013minimal}, which are independent of \emph{both} the particular used logic \emph{and} the particular used theorem prover. This independence is pivotal for the out-of-the-box applicability of diagnosis computation algorithms in domains where 
many different logics are adopted to solve problems of interest, e.g., in ontology-based intelligent applications, as studied in our evaluations (Sec.~\ref{sec:eval}). 
Given a DPI $\dpi = \tuple{\mo,\mb,\Tp,\Tn}$ as input, one execution of such a black-box algorithm 
repeatedly calls an (arbitrary) reasoner that is sound and complete for consistency checks over the logic by which 
$\dpi$
is expressed, and finally returns one minimal conflict for $\dpi$. 
None of the available black-box algorithms has a worst-case time complexity lower than $O(|\mo|)$ consistency checks \cite{marques2013minimal}. 
Since the performance of diagnosis computation methods depends largely on \emph{(i)}~the complexity of consistency checking for the used logic and \emph{(ii)}~on the number of consistency checks executed, and diagnostic algorithms have no influence on (i),
it is important to minimize (ii) by keeping the number of conflict computations at a minimum. 
\subsubsection{Relationship between Conflicts and Diagnoses}
\label{sec:relationship_conflicts_diagnoses}
Conflicts and diagnoses are closely related in terms of a hitting set and a duality property \cite{Reiter87}: 
\begin{description}[font=\normalfont\em,noitemsep]
	\item[Hitting Set Property] Let $\dpi = \tuple{\mo,\mb,\Tp,\Tn}$ be a DPI. Then $\md$ is a (minimal) diagnosis for $\dpi$ iff $\md$ is a (minimal) hitting set of all minimal conflicts for $\dpi$. \\ ($X$ is a \emph{hitting set} of a collection of sets $\mathbf{S}$ iff $X \subseteq \bigcup_{S_i \in \mathbf{S}} S_i$ and $X \cap S_i \neq \emptyset$ for all $S_i \in S$; $X$ is \emph{minimal} iff there is no other hitting set $X'$ of $\mathbf{S}$ with $X' \subset X$)
	\item[Duality Property] Given a DPI $\dpi = \tuple{\mo,\mb,\Tp,\Tn}$, $X$ is a diagnosis (or: contains a minimal diagnosis) for $\dpi$ iff $\mo \setminus X$ is not a conflict (or: does not contain a minimal conflict) for $\dpi$.
\end{description}

\begin{example} \hspace{-1em}\emph{(Conflicts vs.\ Diagnoses)}\quad\label{ex:diags+conflicts}	
	Reconsider the DPI from Example~\ref{ex:dpi}.
	Regarding the Hitting Set Property, e.g., the minimal diagnosis $\md_1$ (see Example~\ref{ex:diagnoses}) is a hitting set of all minimal conflict sets because each conflict (see Example~\ref{ex:conflicts}) contains $\tax_1$ or $\tax_3$. It is moreover a \emph{minimal} hitting set since the elimination of $\tax_1$ implies an empty intersection with, e.g., $\mc_1$, and the elimination of $\tax_3$ means that, e.g., $\mc_4$ is no longer hit. Thus, given the collection $\mC$ of all minimal conflicts, we can determine all the minimal diagnoses as the collection of minimal hitting sets of $\mC$.
	
	Concerning the Duality Property, e.g., $\md_4$ is a diagnosis as $\mo \setminus \md_4 = \{\tax_1, \tax_3$, $\tax_4\}$ is not a conflict (this can be easily verified by checking that no minimal conflict in Example~\ref{ex:conflicts} is a subset of this set), or, equivalently,
	$(\mo \setminus \md_4) \cup \mb \cup \Tp = \{\tax_1, \tax_3, \tax_4\}$ is both consistent and does not entail $\lnot A$. 
	Inversely, e.g., $\mc_2$ is a conflict since $\mo\setminus\mc_2 = \{\tax_1,\tax_5\}$ is not a diagnosis (again, this can be easily seen by verifying that no minimal diagnosis in Example~\ref{ex:diagnoses} is a subset of this set), or, equivalently,
	$(\mo\setminus(\mo\setminus\mc_2)) \cup \mb \cup \Tp = \mc_2 \cup \mb \cup \Tp= \{\tax_2,\tax_3,\tax_4\}$ entails the negative measurement $\lnot A$.
	\qed   
\end{example}

\subsection{Search}
\label{sec:search}

\subsubsection{Path-Finding Problem}
\label{sec:path-finding_problem}
A \emph{path-finding problem instance (PPI)} \cite{russellnorvig2010} can be characterized as a tuple $\langle S_0,$ $\sucs,\goal,\cost \rangle$ where $S_0$ is a distinguished \emph{initial state}, $\sucs$ is a \emph{successor function} that returns all directly reachable neighbor states of any given state, $\goal$ is a Boolean \emph{goal test} that returns $\true$ iff a given state is a goal state, and $\cost$ is a \emph{cost function} that assigns a 
real-valued cost to any given sequence of states (called \emph{path}). 
A \emph{solution} to a PPI is a path from the initial state to some goal state, and the objective is often to find an \emph{optimal solution}, i.e., one with the least costs among all solutions.

\subsubsection{Path-Finding Search Algorithms}
\label{sec:path-finding_search_algos}
\paragraph{Basic Notions and Principle} 
Algorithms that tackle PPIs usually produce a systematic search tree.
The root node 
$\node_0$ of a search tree corresponds to the state $S_0$, and from a node $\node$ corresponding to state $S$ there are $|\sucs(S)|$ emanating edges to other nodes, each of which represents one of the states in $\sucs(S)$. 
The creation of child nodes from a current leaf node $\node$ by means of $\sucs$ is called \emph{expansion} of $\node$. Inversely, the creation of a child node $\node$ when its parent is expanded is called \emph{generation} of $\node$. Importantly, \emph{each generated node $\node$ stores a pointer to its parent} to allow for the reconstruction of the path to $\node$ in case it is a goal. 
Note that one and the same state can occur multiple times in a search tree, depending on the used algorithm. In general, different ways of constructing the search tree---i.e., in which order nodes are selected for expansion, and how much about the tree construction ``history'' (e.g., already expanded nodes) is stored
---yield a variety of search methods with different properties regarding \emph{completeness} 
(will a solution be found whenever one exists?),
\emph{best-first property} or \emph{optimality}
(will the best solution be found first?), 
as well as \emph{time and space complexity} (how much time and memory will the algorithm need to find a solution?). Search algorithms that solve PPIs usually stop after the first path to a goal state is found.

\paragraph{(Un)Informed Search} 
If problem-specific information beyond the mere PPI
is (not) available to an algorithm, the problem is called \emph{(un)informed}. If applicable, such problem-specific information is normally given as a \emph{heuristic function} $h$ which assigns to each node $\node$ a non-negative real value as an estimation of the cost of the best path from $\node$'s state to some goal state. This heuristic value $h(\node)$ can then be combined with the costs $\cost(\node)$ already incurred to reach $\node$, in terms of $f(\node) := \cost(\node) + h(\node)$, which estimates the overall cost of the path from the start to some goal state via node $\node$. The cost function $f$ is called \emph{monotonic} iff $f(\node) \leq f(\node')$ for all nodes $\node,\node'$ where $\node'$ is a successor of $\node$. Some search algorithms require a monotonic function in order to guarantee optimality of the search.


\begin{example} \hspace{-1em}\emph{(Search Algorithms)}\quad \label{ex:search_algorithms}
	Important uninformed search strategies are \emph{depth-first}, \emph{breadth-first}, \emph{uniform-cost} and \emph{iterative deepening} search; popular informed search methods are \emph{A*} and \emph{IDA*} \cite{russellnorvig2010}. Each of them maintains a queue of nodes that is sorted in a specific way, where the first node of this queue is chosen for expansion at each step. Each expanded node is deleted from the queue and its generated successors are added to it in a way the defined sorting is preserved. 
	Whenever a node is expanded whose state satisfies the $\goal$ test, the respective path is returned and the search terminates. 
	
	Now, depth-first search maintains a LIFO queue, breadth-first search a FIFO queue, and uniform-cost search and A*, respectively, a queue sorted in ascending order by $\cost$ and $f$. Iterative deepening and IDA* run in iterations, executing one depth-first search per iteration.
	At this, each iteration uses an incremented depth-limit $l=1,2,\dots$ (iterative deepening) or an incremented cost-limit equal to the best known node from the last iteration that has not been expanded (IDA*).
	A depth-limit (cost-limit) $k$ means that no successors are generated for any node at tree depth $k$ (with cost $> k$).\qed
	%
\end{example}
%


%
\subsubsection{Diagnosis Search Algorithms}
\label{sec:diagnosis_search_algos}
\paragraph{Principle} Given a DPI $\langle\mo,\mb,\Tp,$ $\Tn\rangle$, a diagnosis search algorithm
is characterized by the definition of a \emph{node processing procedure}. 
The latter is divided into two parts, \emph{node labeling} and \emph{node assignment}. A \emph{generic} diagnosis search 
then works as follows:
%
\begin{itemize}[topsep=-0pt,noitemsep]
	\item Start with a queue including only the root node $\emptyset$.
	\item While the queue is non-empty and not enough minimal diagnoses have been found,
	poll the first node $\node$ from the queue and \emph{process} it. That is, \emph{compute a label} $L$ for $\node$, and \emph{assign} $\node$ (or potentially its successors) to an appropriate node class (e.g., solutions, non-solutions) based on $L$.
\end{itemize}
Different \emph{specific} diagnosis search algorithms 
are obtained by (re)defining \emph{(i)}~the sorting of the queue and \emph{(ii)}~the node processing procedure. 

\paragraph{A Prominent Example} The next example explains the workings of Reiter's seminal HS-Tree algorithm \cite{Reiter87} (and of a uniform-cost variant thereof \cite[Sec.~4.6]{Rodler2015phd}) based on the above generic characterization. 
HS-Tree is a widely used diagnosis computation technique, which is (still) the method of choice in domains where its distinguished combination of the features \emph{soundness} (computation of only minimal diagnoses), \emph{completeness} (generation of all minimal diagnoses), the \emph{best-first property} (enumeration of diagnoses in a preference order), as well as the \emph{independence of the used logic and reasoning procedure}, is vital. One such domain is the quality assurance of knowledge-based applications based on ontologies, which will also be the focus of our evaluations.

\begin{example} \hspace{-1em}\emph{(Reiter's HS-Tree)}\quad \label{ex:Reiter's_HS_Node_Processing} 
	The sorting of the queue as well as the node labeling and assignment are implemented as follows by (uniform-cost) HS-Tree:\vspace{3pt}
	
	\noindent\emph{Sorting of the queue:} Depending on the desired preference criterion to be optimized, 
	either a FIFO-queue is used (breadth-first search; \emph{minimum-cardinality diagnoses first}) or the queue is kept sorted in descending order of $\pr(\node)$, cf.\ Eq.~\ref{eq:diag_prob} (uniform-cost search; \emph{most probable diagnoses first}).\vspace{3pt}
	
	\noindent\emph{Node labeling:}
	The following checks are executed in the given order, and a label is returned as soon as the first check is positive:
	\begin{description}[topsep=3pt,noitemsep,font=\normalfont\em]
		\item[(non-minimality)] \label{enum:hstree:label:non-min}
		Is $\node$ a superset of some already found diagnosis? If yes, return $L = \closed$.
		\item[(duplicate)] \label{enum:hstree:label:duplicate} 
		Is there another node equal to $\node$ in the queue? If yes, return $L = \closed$.
		%
		\item[(reuse label)] \label{enum:hstree:label:conflict_reuse} 
		Is there a conflict $\mc$ among the already used node labels such that $\node \cap \mc = \emptyset$? If yes, return $L = \mc$.
		\item[(compute label)] \label{enum:hstree:label:getMinConflict}
		Compute a minimal conflict for $\tuple{\mo\setminus\node,\mb,\Tp,\Tn}$. If some set $\mc$ is computed, return $L = \mc$. If `no conflict' is output, return $L = \valid$. 
		%
	\end{description}\vspace{3pt}
	
	\noindent\emph{Node assignment:} If $\node$'s computed label 
	\begin{description}[topsep=3pt,noitemsep,font=\normalfont\em]
		\item[$L = \tuple{\tax_1,\dots,\tax_k}$\emph{ (a minimal conflict)},] then $k$ new successor nodes $\node_1,\dots,\node_k$ are generated and added to the queue, where $\node_i = \node \cup \{\tax_i\}$. 
		\item[$L = \valid$,] then $\node$ is a solution and added to the collection of minimal diagnoses. 
		\item[$L = \closed$,] then $\node$ is irrelevant or a proven non-solution and not added to any collection, i.e., it is discarded.
	\end{description}\vspace{3pt}
	Note, apart from guiding the node assignment, there is no purpose of a node's label $L$. Thus, in the queue, only nodes are stored, but not the labels along their paths. In a separate collection, already used node labels are recorded due to the \emph{reuse label} check. 
\vspace{3pt}
	
	\noindent\emph{Remarks:} 
	\begin{enumerate}[topsep=0pt,noitemsep]
	\item In order for this algorithm to be sound, complete and best-first
	\begin{itemize}[noitemsep]
		\item the function for conflict computation used in \emph{(compute label)} must be sound (if a set is returned, it is a conflict), complete (a conflict is returned whenever there is one), and must return only \emph{minimal}\footnote{If the minimality of computed conflicts is not guaranteed, HS-Tree becomes generally incomplete, and a directed acyclic graph version must be used to re-establish completeness, cf.\ \cite{greiner1989correction}.} conflicts, and 
		\item (for uniform-cost search) 
		$\pr$ needs to be strictly antimonotonic \cite[Sec.~4.6]{Rodler2015phd}.\footnote{If $\pr$ violates this criterion, then either the transformation of $\pr$ described in Sec.~\ref{sec:probability_model} can be applied, or breadth-first HS-Tree can be used to first compute all (or a feasible set of) minimal diagnoses which can then be ordered by $\pr$ in a post-processing step before being returned.}
	\end{itemize}
	\item \label{rem:breadth-first_can_be_simluated_by_uniform-cost} Breadth-first search can be simulated by a uniform-cost search using $\pr(\tax) := c$ for all $\tax \in \mo$ with any fixed $c \in (0,0.5)$ (cf.\ Eq.~\ref{eq:diag_prob} and Sec.~\ref{sec:probability_model}). 
	That is, the minimum-cardinality-first computation of diagnoses is equivalent to a most-probable-first computation given small uniform component probabilities.\qed  
\end{enumerate}

\end{example}
%

\subsubsection{Diagnosis Search vs.\ Path-Finding}
\label{sec:diagnosis_search_vs_path-finding}
Since the main aim of this work is to leverage ideas from classic path-finding search to derive a novel diagnosis computation approach, we next identify the main properties that distinguish diagnosis search (Sec.~\ref{sec:diagnosis_search_algos}) from path-finding (Sec.~\ref{sec:path-finding_search_algos}) algorithms:
\begin{enumerate}[itemsep=2pt,label=\emph{(\Roman*)},align=left,leftmargin=0pt,labelwidth=-0.5\parindent]
	%
	\item \label{enum:diff:PPI_formulation_not_sufficient} \emph{PPI-formulation does not suffice as an input:} Although the problem of searching for minimal diagnoses for a DPI can be stated as a PPI---where $S_0 = \emptyset$; $\sucs$ gets a \emph{labeled} node $\node$ with label $L$ and returns the successors of $\node$ if $L$ is a set, and $\emptyset$ else; $\goal(\node)$ returns $\true$ iff $\node$ is a diagnosis; and $\cost(\node) := \pr(\node)$ as per Eq.~\ref{eq:diag_prob}---this characterization is not a sufficient basis to run a diagnosis search. What is missing is the definition of a \emph{node labeling} and a \emph{node assignment} strategy (see Sec.~\ref{sec:diagnosis_search_algos}). Importantly, these missing building blocks decide over the soundness, completeness and best-first property of the diagnosis search.
	By contrast, for path-finding, the PPI includes all relevant information for the problem to be directly solved by an off-the-shelf path-finding algorithm (cf.\ Example~\ref{ex:search_algorithms}). 
	\item \label{enum:diff:states_paths_nodes_coincide} \emph{States, nodes and paths coincide:} In diagnosis search, the state of a search tree node $\node$ corresponds to $\node$ itself (i.e., to a set of $\tax_i$-elements, cf.\ Example~\ref{ex:Reiter's_HS_Node_Processing}). So, no distinction between states and nodes is made. 
	%
	When the label $\tax_i$ is assumed to be assigned to the edge 
	from any node $\node$ to its child node $\node \cup \setof{\tax_i}$ \cite{Reiter87}, nodes (and states) can be seen as representatives of the (edge labels along the) paths in the search tree.  
	\item \label{enum:diff:solutions_are_sets_not_paths} \emph{Solutions are sets, not paths:} Solutions to a diagnosis search problem are nodes (\emph{sets} of edge labels along a tree path) which are minimal diagnoses for the given DPI. Unlike in path-finding problems, the order of labels along the path does not matter. 
	\item \label{enum:diff:multiple_solution_sought} \emph{Multiple solutions are sought:} In diagnosis search, it is usually of interest to find multiple solutions, i.e., after the first solution is determined, the search must be (correctly) continuable until 
	sufficient
	solutions are found.
	\item \label{enum:diff:search_for_max_cost_solutions} \emph{Search for maximal-cost solutions:}
	In diagnosis search, one wants to calculate the maximal-cost (i.e., most probable, cf.\ Remark~\ref{rem:breadth-first_can_be_simluated_by_uniform-cost} in Example~\ref{ex:Reiter's_HS_Node_Processing})
	 solutions whereas path-finding is usually about finding a minimal-cost solution.
\item \label{enum:diff:stricter_conditions_on_cost_function} \emph{Different conditions on cost function:} Like for path-finding, the cost function used by diagnosis searches must fulfill certain criteria in order for desired properties 
to be guaranteed. 
While (informed) path-finding algorithms usually need a monotonic function $f$ (see Sec.~\ref{sec:path-finding_search_algos}) for optimality, diagnostic searches as characterized in Sec.~\ref{sec:diagnosis_search_algos} usually require the (probability) function $\pr$ used to sort the queue to be strictly antimonotonic (cf.\ Sec.~\ref{sec:probability_model}) in order to be sound, complete and best-first.
	\item \label{enum:diff:soundness_not_trivial} \emph{Soundness is not trivial:} Whereas in path-finding any path whose end state satisfies the goal test is a valid solution to the PPI, in diagnosis search an appropriate combination of suitable goal test, node labeling, node assignment and cost function is necessary to ensure soundness, i.e., that each found solution is indeed a minimal diagnosis for the given DPI.
\end{enumerate}

\begin{algorithm}[t]
	\scriptsize
	\caption{RBFS} \label{algo:RBFS}
	{\fontsize{7pt}{8pt}\selectfont
		\begin{algorithmic}[1]
			\Require 
			\textcolor{white}{.}
			PPI $\ppi := \tuple{S_0,\sucs,\goal,\cost}$ and a heuristic function $h$ (if $\ppi$ is an uninformed problem, then $h(\node) := 0$ for all nodes $\node$)
			\Ensure 
			a path from $S_0$ to some goal state, if a goal state is reachable from $S_0$ by means of successive applications of the $\sucs$ function; \emph{null} otherwise  
			
			\vspace{6pt}
			\Procedure{RBFS}{$\ppi, h$}
			\State $\mathit{solution} \gets \text{\emph{null}}$
			\State $\node_0 \gets \Call{makeNode}{S_0}$   
			\Comment{\textsc{makeNode} creates a tree node for the given state}
			\label{algoline:rbfs:generate_root_node}
			\State $\Call{RBFS'}{\node_0,f(\node_0),\infty}$ \label{algoline:rbfs:call_RBFS'} \Comment{$f(\node) := \cost(\node) + h(\node)$}
			\State \Return $\mathit{solution}$	\label{algoline:rbfs:return_solution}																															
			\EndProcedure
			\vspace{6pt}
			
			\Procedure{RBFS'}{$\mathsf{\node}, F(\mathsf{\node}), \bnd$}
			\If{$\goal(\textsc{state}(\node))$}  \Comment{\textsc{state} returns the state associated with the given node} \label{algoline:rbfs':goal_test}
			\State $\mathit{solution} \gets \Call{getPathTo}{\node}$   \Comment{\textsc{getPathTo} returns sequence of states 
			from root node to given node} \label{algoline:rbfs':getPathToNode}
			\State \textbf{exit procedure} \label{algoline:rbfs':exit_procedure}
			\EndIf
			\State $\childnodes \gets [\,]$
			\For{$S_i \in \sucs(\textsc{state}(\node))$}
			\State $\childnodes \gets \Call{add}{\textsc{makeNode}(S_i),\childnodes}$ \label{algoline:rbfs':generate_successors}\Comment{$\textsc{add}(e,C)$ adds element $e$ to collection $C$ } 
			\EndFor
			\If{$\childnodes = [\,]$}
			\State \Return $\infty$   \Comment{$\node$ is hopeless, i.e., is no goal and has no children}
			\EndIf
			\For{$\node_i \in \childnodes$}
			\If{$f(\node) < F(\node)$} \label{algoline:rbfs':f<N?} \Comment{if $\true$, $\node$ was already expanded before}
			\State $F(\node_i) \gets \max(F(\node),f(\node_i))$ \label{algoline:rbfs':F(n_i)_gets_max}
			\Else
			\State $F(\node_i) \gets f(\node_i)$ \label{algoline:rbfs':F(n_i)_gets_f(n_i)}
			\EndIf
			\EndFor 
			\If{$|\childnodes|=1$} \Comment{$\textsc{addDummyNode}$ adds a newly created ``dummy'' node $\node_d$...} 
			\State $\childnodes \gets \Call{addDummyNode}{\childnodes}$ \Comment{...with $F(\node_d) = \infty$ to the given collection}
			\EndIf
			\State $\childnodes \gets \Call{sortIncreasingByF}{\childnodes}$ \label{algoline:rbfs':sortIncreasingByF}\Comment{sort $\childnodes$ in ascending order of $F$-value} 
			\State $\node_1 \gets \Call{getAndDeleteFirstNode}{\childnodes}$ \label{algoline:rbfs':getBestChild_1} \Comment{$\node_1\dots$best child}
			\State $\node_2 \gets \Call{getFirstNode}{\childnodes}$ \Comment{$\node_2\dots$2nd-best child}
			\While{$F(\node_1) \leq \bnd \;\land\; F(\node_1) < \infty$} \label{algoline:rbfs':while}
			\State $F(\node_1) \gets \Call{RBFS'}{\node_1,F(\node_1),\min(\bnd,F(\node_2))}$ \label{algoline:rbfs':recursive_call}
			\State $\childnodes \gets \Call{insertSortedByF}{\node_1, \childnodes}$ \label{algoline:rbfs':insertSortedByF}\Comment{insert $\node_1$ s.t. sorting by $F$-value is preserved} 
			\State $\node_1 \gets \Call{getAndDeleteFirstNode}{\childnodes}$ \label{algoline:rbfs':getBestChild_2} \Comment{$\node_1\dots$best child}
			\State $\node_2 \gets \Call{getFirstNode}{\childnodes}$ \Comment{$\node_2\dots$2nd-best child}
			\EndWhile
			\State \Return $F(\node_1)$ \label{algoline:rbfs':return_F(n)}
			\EndProcedure	
		\end{algorithmic}
	}
	\normalsize
\end{algorithm}

\section{Recursive Best-First Hitting Set Search (RBF-HS)}
\label{sec:algo}

\subsection{Deriving RBF-HS from RBFS}
\label{sec:engineering_rbfhs}

\subsubsection{RBFS: The Basis}
Korf's RBFS algorithm \cite{korf1992linear,korf1993linear} provides the inspiration for RBF-HS. Historically, the main motivation that led to the engineering of RBFS was the problem that best-first searches by that time required exponential space. The idea behind RBFS is to trade (more) time for (much less) space. To this end, RBFS implements a scheme that can be synopsized as 
\begin{itemize}[noitemsep]
	\item \emph{(complete and best-first):} always expand current globally-best node while remembering current globally-second-best node,
	\item \emph{(undo and forget to keep space linear):} backtrack and explore second-best node if none of the child nodes of best node is better than second-best,
	\item \emph{(remember utility of forgotten subtrees to keep the search progressing):} 
	before deleting a subtree in the course of backtracking, store cost of subtree's best node,
	\item \emph{(restore utility at regeneration to avoid redundancy):} whenever a subtree is reexplored, use this stored cost value to update node costs in the subtree.
\end{itemize} 
%
As a result, RBFS is complete and best-first and works within linear-space bounds.

\subsubsection{RBFS: Briefly Explained}
RBFS is presented by Alg.~\ref{algo:RBFS}.
In a nutshell, it works as follows \cite{russellnorvig2010}. Initial node costs are the $f$-values computed from $\cost$ and $h$, and \emph{backed-up node costs} are named $F$-values. Initially, all backed-up node costs are the nodes' initial costs.
Starting from the root node corresponding to $S_0$, the principle is to follow the best (lowest $F$) path downwards (recursive RBFS'-calls, line~\ref{algoline:rbfs':recursive_call}). At each downward step, the variable $\bnd$ is used to keep track of the (backed-up) cost of the best alternative path available from any ancestor of the current node (note, this is the \emph{globally} best alternative path). If the current node exceeds $\bnd$, the recursion unwinds back to the alternative path. As the recursion unwinds, the cost of each node along the path is replaced with a (new) backed-up cost value, which is the best (backed-up) cost of its child nodes (cf.\ line~\ref{algoline:rbfs':return_F(n)}). In this way, RBFS always remembers the backed-up cost of the best leaf in the forgotten subtree and can therefore decide whether it is worth reexpanding the subtree at some later time (this decision is made through the condition of the while-loop). 
%
When expanding a subtree rooted at node $\node$, which has already been expanded and forgotten before (condition in line~\ref{algoline:rbfs':f<N?} is true) and whose initial cost ($f$-value) appears more promising than the algorithm knows from a previous iteration and the stored backed-up cost $F(\node)$ it actually is, the $F$-value of child nodes $\node_i$ of $\node$ is not tediously learned again by RBFS, but directly updated by means of $\node$'s $F$-value
 (see line~\ref{algoline:rbfs':F(n_i)_gets_max}).
If some node is recognized to correspond to a goal state, the path to this node is returned and RBFS' terminates (lines~\ref{algoline:rbfs':goal_test}--\ref{algoline:rbfs':exit_procedure}). 

\subsubsection{From RBFS to RBF-HS: Necessary Modifications} 
\label{sec:necessary_modifications}
In order to transform a path-finding into a diagnosis search algorithm, we have to make adequate amendments to the former with due regard to all differences between both paradigms discussed in Bullets \ref{enum:diff:PPI_formulation_not_sufficient}--\ref{enum:diff:soundness_not_trivial} in Sec.~\ref{sec:diagnosis_search_vs_path-finding}.
Next, we list and explain the main modifications necessary to derive RBF-HS from RBFS (line numbers given refer to the respective locations of the changes \emph{in the RBF-HS algorithm}, i.e., in Alg.~\ref{algo:RBF_HS}).  
\begin{enumerate}[topsep=2pt,label=\textbf{(Mod\arabic*)}, wide, labelwidth=!, labelindent=0pt,itemsep=2pt]
	\item \label{enum:mod:} A node labeling (line~\ref{algoline:rbfhs':label} and \textsc{label} procedure) and a node assignment (lines \ref{algoline:rbfhs':if_nonmin}--\ref{algoline:rbfhs':return_after_valid}) strategy have to be added. Importantly, the goal test (check, whether a node is a minimal diagnosis, lines~\ref{algoline:label:if_n_supseteq_n_i}, \ref{algoline:label:if_C_cap_node=emptyset} and \ref{algoline:label:findMinConflict}) as well as the preparation of nodes for expansion (i.e., the provision of a minimal conflict, line~\ref{algoline:label:return_C} or \ref{algoline:label:return_new_cs}) is part of these two code blocks. 
	\emph{Justification:} Bullet \ref{enum:diff:PPI_formulation_not_sufficient}.
	%
	\item \label{enum:mod:} 
	Differentiation between nodes, states and paths is no longer necessary, which is why the functions \textsc{makeNode} (generates node from state), \textsc{state} (extracts state from node), and \textsc{getPathTo} (returns path from root to node) can be omitted. This becomes evident in 
	\begin{itemize}[noitemsep,topsep=0pt]
		\item line~\ref{algoline:rbfhs:call_RBFHS'} (root node is simply equal to initial state $\emptyset$; cf.\ line~\ref{algoline:rbfs:generate_root_node} in Alg.~\ref{algo:RBFS}),
		\item line~\ref{algoline:rbfhs:add_node_to_mD} (a \emph{set} $\node$ is added to the solutions $\mD$; cf.\ line~\ref{algoline:rbfs':getPathToNode} in Alg.~\ref{algo:RBFS}),
		\item line~\ref{algoline:rbfhs':expand} and \textsc{expand} function (successors are generated directly from the \emph{node} $\node$; cf.\ line~\ref{algoline:rbfs':generate_successors} in Alg.~\ref{algo:RBFS}), and 
		\item lines~\ref{algoline:label:if_n_supseteq_n_i}, \ref{algoline:label:if_C_cap_node=emptyset} and \ref{algoline:label:'no_conflict'} (goal test performed on node, not state; cf.\ line~\ref{algoline:rbfs':goal_test} in Alg.~\ref{algo:RBFS}).
	\end{itemize}
	\emph{Justification:} Bullets \ref{enum:diff:states_paths_nodes_coincide} and \ref{enum:diff:solutions_are_sets_not_paths}.
	%
	\item \label{mod:multiple_solutions} The requirement that \emph{multiple} solutions are generally desired in diagnosis search is handled in lines~\ref{algoline:rbfhs':if_mD_geq_ld}--\ref{algoline:rbfhs':return_after_valid}. 
	Note, it is essential to return $-\infty$ (i.e., the worst possible cost) as the backed-up $F$-cost of the solution node $\node$ in order to allow the search to continue in a well-defined and correct way. 
	More precisely, this will cause the $F$-value of $\node$'s best sibling node to be propagated upwards. 
	As a consequence, the backed-up value for any subtree including $\node$ will be the so-far found best cost over all nodes in this subtree \emph{except for $\node$.}
	In fact, any backed-up value $F^* := F(\node) > -\infty$ would prevent RBF-HS' from terminating and thus would make it incomplete (intuitively, at some point all other nodes would have a value lower than $F^*$ and the algorithm would loop forever exploring $\node$ again and again).  
	\emph{Justification:} Bullet \ref{enum:diff:multiple_solution_sought}.
	%
	\item Since solutions of maximal cost are stipulated in diagnosis search, all occurrences of $<$, $\leq$, $\min$, $\max$, $\infty$, \textsc{sortIncreasingByF} have to be switched to $>$, $\geq$, $\max$, $\min$, $-\infty$, \textsc{sortDecreasingByF}, respectively.
	\emph{Justification:} Bullet \ref{enum:diff:search_for_max_cost_solutions}.
	%
	\item
	The used function $f$ (probability measure $\pr$) needs to be strictly antimonotonic. 
	\emph{Justification:} Bullet \ref{enum:diff:stricter_conditions_on_cost_function}.
	%
	\item To achieve soundness (only minimal diagnoses are added to the solutions $\mD$ in line~\ref{algoline:rbfhs:add_node_to_mD}), the 
following provisions are made. 
Successor nodes $\childnodes$ are always sorted by a strictly antimonotonic function (line~\ref{algoline:rbfhs':sortDecreasingByF}), which is why minimal diagnoses will be found prior to non-minimal ones.
Moreover, the \textsc{label} function is designed such that only nodes $\node$ can be labeled $\valid$ for which no already-found diagnosis exists which is a subset of $\node$ (goal test, part~1, line~\ref{algoline:label:if_n_supseteq_n_i}), and which is evidentially a diagnosis (goal test, part~2, line~\ref{algoline:label:'no_conflict'}). Finally, the node assignment ensures that only nodes labeled $\valid$ can be assigned to the solution list $\mD$ (line~\ref{algoline:rbfhs:add_node_to_mD}).
	\emph{Justification:} Bullet \ref{enum:diff:soundness_not_trivial}.
\end{enumerate}

\begin{algorithm}[!htbp]
	\scriptsize
	\caption{RBF-HS} \label{algo:RBF_HS}
	{\fontsize{7pt}{8pt}\selectfont
		\begin{algorithmic}[1]
			\Require 
			\textcolor{white}{.}
			tuple $\tuple{\dpi, \pr, \ld}$ comprising
			\begin{itemize}[noitemsep]
				\item a DPI $\dpi = \langle\mo,\mb,\Tp,\Tn\rangle$
				\item a function $\pr$ that assigns a failure probability $\pr(\tax) \in (0,1)$ to each $\tax \in \mo$, where $\pr$ is strictly antimonotonic (cf.\ Sec.~\ref{sec:probability_model}); \quad \emph{note:} the cost function $f(\node) := \pr(\node)$ as per Eq.~\ref{eq:diag_prob} for all tree nodes $\node \subseteq \mo$
				\item the number $\ld$ of leading minimal diagnoses to be computed 
			\end{itemize}
			\Ensure 
			list $\mD$ where $\mD$ is the list of the $\ld$ (if existent) most probable (as per $\pr$) 
			minimal diagnoses wrt.\ $\dpi$, sorted by probability in descending order  
			
			\vspace{6pt}
			\Procedure{RBF-HS}{$\dpi, \pr, \ld$}
			\State $\mD \gets [\,], \; \mC \gets [\,]$
			\label{algoline:rbfhs:initialize_mD_mC_f}
			\State $\mc \gets \Call{FindMinConflict}{\dpi}$ \label{algoline:rbfhs:findMinConflict}
			\If{$\mc =\emptyset$}  \Comment{$\mc = \emptyset \Leftrightarrow \emptyset \cup \mb \cup U_{\Tp} \models x$ for some $x \in \{\bot\} \cup \Tn$ (cf.\ Sec.~\ref{sec:diagnoses}) $\Leftrightarrow$...} 
			\label{algoline:rbfhs:mc=emptyset}
			\State \Return $\mD$ \Comment{...even if all comp.\ are assumed faulty, the problem persists $\Leftrightarrow$ no diagnosis} \label{algoline:rbfhs:return_mD_1}	
			\EndIf
			\If{$\mc = \text{`no conflict'}$} \Comment{`no conflict' $\Leftrightarrow$ even if all comp.\ are assumed normal, no problem...} \label{algoline:rbfhs:mc='no_conflict'}
			\State \Return $[\emptyset]$  \Comment{...exists $\Leftrightarrow$ the only minimal diagnosis is to assume no comp.\ faulty} \label{algoline:rbfhs:return_mD_2}	
			\EndIf
			\State $\mC \gets \Call{add}{\mc,\mC}$  \label{algoline:rbfhs:add_mc_to_mC}	
			\State $\Call{RBF-HS'}{\node_0,f(\node_0),-\infty}$
			\label{algoline:rbfhs:call_RBFHS'} \Comment{$\node_0 := \emptyset$ is the root node}
			\State \Return $\mD$	\label{algoline:rbfhs:return_mD_3}																															
			\EndProcedure
			\vspace{6pt}
			
			\Procedure{RBF-HS'}{$\mathsf{\node}, F(\mathsf{\node}), \bnd$} \label{algoline:rbfhs':procedure_RBF-HS'}
			\State $L \gets \Call{label}{\node}$ \label{algoline:rbfhs':label}
			\If{$L = \closed$} \label{algoline:rbfhs':if_nonmin}
			\State \Return $-\infty$ \label{algoline:rbfhs':return_after_closed}
			\EndIf
			\If{$L = \valid$} \label{algoline:rbfhs':if_valid}
			\State $\mD \gets \Call{add}{\node,\mD}$ \label{algoline:rbfhs:add_node_to_mD}  \Comment{new minimal diagnosis found}
			\If{$|\mD| \geq \ld$}    \label{algoline:rbfhs':if_mD_geq_ld}
			\State \textbf{exit procedure} \label{algoline:rbfhs':exit_procedure}
			\EndIf
			\State \Return $-\infty$   \label{algoline:rbfhs':return_after_valid}
			\EndIf
			\State $\childnodes \gets \Call{expand}{\node,L}$ \label{algoline:rbfhs':expand}
			\For{$\node_i \in \childnodes$} \label{algoline:rbfhs':for_node_in_childnodes}
			\If{$f(\node) > F(\node)$} \label{algoline:rbfhs':if_f(n)>F(n)} \Comment{if $\true$, $\node$ was already expanded before}
			\State $F(\node_i) \gets \min(F(\node),f(\node_i))$ \label{algoline:rbfhs':F(n_i)_gets_min}
			\Else
			\State $F(\node_i) \gets f(\node_i)$ \label{algoline:rbfhs':F(n_i)_gets_f(n_i)}
			\EndIf
			\EndFor 
			\If{$|\childnodes|=1$} \label{algoline:rbfhs':if_|childnodes|=1} \Comment{add dummy node $\node_d$ with $F(\node_d) = -\infty$}
			\State $\childnodes \gets \Call{addDummyNode}{\childnodes}$
			\EndIf
			\State $\childnodes \gets \Call{sortDecreasingByF}{\childnodes}$ \label{algoline:rbfhs':sortDecreasingByF}
			\State $\node_1 \gets \Call{getAndDeleteFirstNode}{\childnodes}$ \label{algoline:rbfhs':getBestChild_1} \Comment{$\node_1\dots$best child}
			\State $\node_2 \gets \Call{getFirstNode}{\childnodes}$ \label{algoline:rbfhs':getSecondBestChild_1} \Comment{$\node_2\dots$2nd-best child}
			\While{$F(\node_1) \geq \bnd \;\land\; F(\node_1) > -\infty$} \label{algoline:rbfhs':while}
			\State $F(\node_1) \gets \Call{RBF-HS'}{\node_1,F(\node_1),\max(\bnd,F(\node_2))}$ \label{algoline:rbfhs':recursive_call}
			\State $\childnodes \gets \Call{insertSortedByF}{\node_1, \childnodes}$ \label{algoline:rbfhs':insertSortedByF}
			\State $\node_1 \gets \Call{getAndDeleteFirstNode}{\childnodes}$ \label{algoline:rbfhs':getBestChild_2} \Comment{$\node_1\dots$best child}
			\State $\node_2 \gets \Call{getFirstNode}{\childnodes}$ \label{algoline:rbfhs':getSecondBestChild_2} \Comment{$\node_2\dots$2nd-best child}
			\EndWhile
			\State \Return $F(\node_1)$ \label{algoline:rbfhs':return_F(n)}
			\EndProcedure
			\vspace{6pt}
			
			\Procedure{\textsc{label}}{$\node$} 
			\For{$\node_i \in \mD$}\label{algoline:label:non-min_crit_start}
			\If{$\node \supseteq \node_i$}  \Comment{goal test, part 1 (is $\node$ non-minimal?)} \label{algoline:label:if_n_supseteq_n_i}  
			\State \Return $\closed$ \Comment{$\node$ is a non-minimal diagnosis}
			\EndIf
			\EndFor\label{algoline:label:non-min_crit_end}
			\For{$\mc \in \mC$}\label{algoline:label:reuse_start}
			\If{$\mc \cap \node = \emptyset$}\label{algoline:label:if_C_cap_node=emptyset}  \Comment{cheap non-goal test (is $\node$ not a diagnosis?)}   
			\State \Return $\mc$\label{algoline:label:return_C} \Comment{$\node$ is not a diagnosis; reuse $\mc$ to label $\node$}
			\EndIf 
			\EndFor\label{algoline:label:reuse_end}
			\State $L\gets \Call{findMinConflict}{\langle\mo\setminus\node,\mb,\Tp,\Tn\rangle}$\label{algoline:label:findMinConflict}  
			\If{$L$ = \text{`no conflict'}}	 \Comment{goal test, part 2 (is $\node$ diagnosis?)} \label{algoline:label:'no_conflict'}					
			\State \Return $\valid$\label{algoline:label:return_valid} \Comment{$\node$ is a minimal diagnosis}
			\Else	\Comment{$\node$ is not a diagnosis}
			\State $\mC \gets \Call{add}{L,\mC}$\label{algoline:label:add_new_cs}  \Comment{$L$ is a \emph{new} minimal conflict ($\notin \mC$)}
			\State \Return $L$\label{algoline:label:return_new_cs}
			\EndIf
			\EndProcedure
			
			\vspace{6pt}
			
			\Procedure{\textsc{expand}}{$\node, \mc$}
			\State $\mathsf{Succ\_Nodes} \gets [\,]$
			\For{$e \in \mc$} \label{algoline:expand:for_loop}
			\State $\mathsf{Succ\_Nodes} \gets \Call{add}{\node\cup\{e\},\mathsf{Succ\_Nodes}}$ \label{algoline:expand:add_successor_node}
			\EndFor
			\State \Return $\mathsf{Succ\_Nodes}$
			\EndProcedure	
		\end{algorithmic}
	}
	\normalsize
\end{algorithm}

\subsection{RBF-HS Algorithm Description}
\label{sec:algo_walkthrough}
\subsubsection{Inputs and Output}
\label{sec:inputs+outputs} 
RBF-HS is depicted by Alg.~\ref{algo:RBF_HS}. It accepts the following arguments: 
a DPI $\dpi = \tuple{\mo,\mb,\Tp,\Tn}$,
a probability measure $\pr$ (see Sec.~\ref{sec:probability_model}), and 
a stipulated number $\ld$ of minimal diagnoses (``leading diagnoses'') to be returned. It outputs the $\ld$ (if existent) minimal diagnoses of maximal probability wrt.\ $\pr$ for $\dpi$. Note, 
the computation of the $\ld$ diagnoses of minimal cardinality (instead of maximal probability) can be effectuated by specifying $\pr$ accordingly (cf.\ Remark~\ref{rem:breadth-first_can_be_simluated_by_uniform-cost} in Example~\ref{ex:Reiter's_HS_Node_Processing}).
%
%
%
%
%
%

\subsubsection{Trivial Cases}
\label{sec:trivial_cases}
At the beginning (line~\ref{algoline:rbfhs:initialize_mD_mC_f}), RBF-HS initializes the solution list of found minimal diagnoses $\mD$ and the list of already computed minimal conflicts $\mC$.
Then, two trivial cases are checked, i.e., whether no diagnoses exist for $\dpi$ (lines~\ref{algoline:rbfhs:mc=emptyset}--\ref{algoline:rbfhs:return_mD_1}), or if the empty set is the only diagnosis for $\dpi$ (lines~\ref{algoline:rbfhs:mc='no_conflict'}--\ref{algoline:rbfhs:return_mD_2}). The former case applies iff the empty set is a conflict for $\dpi$, which implies that $\mo\setminus\emptyset = \mo$ is not a diagnosis for $\dpi$ by the Duality Property (cf.\ Sec.~\ref{sec:relationship_conflicts_diagnoses}), which in turn means that no diagnosis can exist since every diagnosis is a subset of $\mo$ and all supersets of diagnoses are diagnoses as well (weak fault model, cf.\ Sec.~\ref{sec:diagnosis_problem}). The latter case holds iff there is no conflict at all for $\dpi$, i.e., in particular, $\mo$ is not a conflict, which is why $\mo\setminus\mo = \emptyset$ is a diagnosis by the Duality Property, and consequently no other \emph{minimal} diagnosis can exist. 

If none of these trivial cases is given, the call of \textsc{findMinConflict} (line~\ref{algoline:rbfhs:findMinConflict}) returns a non-empty minimal conflict $\mc$ (line~\ref{algoline:rbfhs:add_mc_to_mC} is reached), which entails by the Hitting Set Property (cf.\ Sec.~\ref{sec:relationship_conflicts_diagnoses}) that a non-empty (minimal) diagnosis will exist. For later reuse, 
$\mc$ is added to the computed conflicts $\mC$, and then the recursive sub-procedure RBF-HS' is called (line~\ref{algoline:rbfhs:call_RBFHS'}). The arguments passed to RBF-HS' are the root node $\node_0 := \emptyset$, its $f$-value, and the initial bound set to $-\infty$.

\subsubsection{Recursion: Principle}
The basic principle of the recursion (RBF-HS' procedure) is very similar as sketched above for RBFS. That is, always explore the open node with best $F$-value in a depth-first manner, until the best node has worse costs than the globally best alternative node (whose cost is always stored by $\bnd$). Then backtrack and propagate the best $F$-value among all child nodes up at each backtracking step. Based on their latest known $F$-value, the child nodes at each tree level are resorted in best-first order of $F$-value. When re-exploring an already explored but later forgotten subtree, the cost of nodes in this subtree is, if necessary, 
updated through a cost inheritance from parent to children. In this vein, 
a relearning of already learned backed-up cost-values, and thus repeated and redundant work, is avoided. Exploring a node in RBF-HS means labeling this node and assigning it to an appropriate collection of nodes based on the computed label (cf.\ Sec.~\ref{sec:diagnosis_search_algos} and Example~\ref{ex:Reiter's_HS_Node_Processing}). The recursion is executed until 
$\mD$ comprises the desired number $\ld$ of minimal diagnoses or the hitting set tree has been explored in its entirety.

\subsubsection{Recursion: Structure}
\label{sec:recursion_structure}
To get a better impression of RBF-HS' 
on an abstract, structural level, 
it is instructive to look at RBF-HS' as a succession of the following blocks. 
An algorithm walkthrough with detailed descriptions of all these blocks is given in \ref{apx:algo_walkthrough}. 
\begin{itemize}[noitemsep]
	\item node labeling (line~\ref{algoline:rbfhs':label}),
	\item node assignment (lines~\ref{algoline:rbfhs':if_nonmin}--\ref{algoline:rbfhs':return_after_valid}), 
	\item node expansion (line~\ref{algoline:rbfhs':expand}),
	\item node cost inheritance (lines~\ref{algoline:rbfhs':for_node_in_childnodes}--\ref{algoline:rbfhs':F(n_i)_gets_f(n_i)}),
	\item child node preparation (lines~\ref{algoline:rbfhs':if_|childnodes|=1}--\ref{algoline:rbfhs':sortDecreasingByF}), and
	\item recursive child node exploration (lines~\ref{algoline:rbfhs':getBestChild_1}--\ref{algoline:rbfhs':return_F(n)}).
\end{itemize}
The node labeling (function \textsc{label}) can be further split into the following blocks:
\begin{itemize}[noitemsep]
	\item \emph{non-minimality} check (lines~\ref{algoline:label:non-min_crit_start}--\ref{algoline:label:non-min_crit_end}),
	\item \emph{reuse label} check (lines~\ref{algoline:label:reuse_start}--\ref{algoline:label:reuse_end}), and
	\item \emph{compute label} operations (lines~\ref{algoline:label:findMinConflict}--\ref{algoline:label:return_new_cs}).
\end{itemize}
Note, the \textsc{label} function of RBF-HS' is equal to the one used in Reiter's HS-Tree (cf.\ Example~\ref{ex:Reiter's_HS_Node_Processing}), except that the \emph{duplicate} check is obsolete in RBF-HS'. The reason for this is that there cannot ever be any duplicate (i.e., set-equal) nodes in memory at the same time during the execution of RBF-HS. This holds because for all potential duplicates $\node_i,\node_j$, we must have $|\node_i|=|\node_j|$, but equal-sized nodes must be siblings (depth-first tree exploration) which is why $\node_i$ and $\node_j$ must contain $|\node_i|-1$ equal elements
(same path up to the parent of $\node_i,\node_j$) and one necessarily different element (label of edge pointing from parent to $\node_i$ and $\node_j$, respectively).

\begin{figure}[t!]
	\centering
	\phantom{.}
		\xygraph{
			!{<0cm,0cm>;<1cm,0cm>:<0cm,-1cm>::}
			!{(-1,0) }*+{ \stackrel{\textcolor{red}{-\infty}}{\textcircled{\scriptsize 1}\langle1,2,5\rangle^{C}} }="n1"
			!{(-4.2,.5) }*+{ \stackrel{\textcolor{red}{.25}}{\textcircled{\scriptsize 2}\langle2,4,6\rangle^{C}} }="n21"
			"n1":"n21"_{.41}^{1}
			!{(-1,1) }*+{? }="n22"
			"n1":"n22"^(0.6){2}_(0.6){.25}
			!{(1,1) }*+{ ? }="n23"
			"n1":"n23"_{.25}^(0.6){5}
			!{(-4.5,.16) }*+{ \phantom{.} }="n2a"
			!{(-5.5,1.7) }*+{ ? }="n31"
			"n21":"n31"_(0.7){.09}^(0.6){2}
			!{(-3.9,1.7) }*+{ \underset{\color{ForestGreen}-\infty}{\textcircled{\scriptsize 3}\checkmark_{(\md_1)}} }="n32"
			"n21":"n32"_(0.55){.28}^(0.6){4}
			!{(-2,1.7) }*+{ \underset{\color{ForestGreen}-\infty}{\textcircled{\scriptsize 4}\checkmark_{(\md_2)}} }="n33"
			"n21":"n33"^(0.6){.27}_(0.6){6}
			!{(-3.6,2.05) }*+{\color{ForestGreen}\underbrace{\phantom{\qquad\qquad\qquad\qquad\qquad\qquad}}  }="n41"
			!{(-0,1.9) }*+{ \boxed{\scriptstyle \mD = [ [1,4],[1,6] ] }}="key"
			!{(-3.6,2.15) }*+{\phantom{.}}="n51"
			"n51" :@/^2.5cm/^{\textcolor{ForestGreen}{.09}}@[ForestGreen] "n2a"
		}
		\vspace{-2pt}
		\begin{center}
			\scriptsize 
			backtrack \#1, discard subtree $\textcircled{\scalebox{.7}{2}}$ (${\color{ForestGreen}F(\node_1) = 0.09} < {\color{red}0.25 = \bnd}$)
		\end{center}
		\vspace{-10pt}
		
		\hdashrule[0.5ex]{\columnwidth}{1pt}{3mm} 
%
		\xygraph{
			!{<0cm,0cm>;<1cm,0cm>:<0cm,-1.2cm>::}
			!{(-2,0) }*+{ \stackrel{\textcolor{red}{-\infty}}{\textcircled{\scriptsize 1}\langle1,2,5\rangle^{C}} }="n1"
			!{(-5,.5) }*+{ ? }="n21"
			"n1":"n21"_{\textcolor{ForestGreen}{.09}}^{1}
			!{(-2,1) }*+{ \stackrel{\textcolor{red}{.25}}{\textcircled{\scriptsize 5}\langle1,3,4\rangle^{C}} }="n22"
			"n1":"n22"^(0.5){2}_(0.5){.25}
			!{(1,.5) }*+{ ? }="n23"
			"n1":"n23"_{.25}^(0.6){5}
			!{(-2.3,0.87) }*+{ \phantom{.} }="n2a"
			!{(-4,2) }*+{ ? }="n31"
			"n22":"n31"_(0.7){.09}^(0.6){1}
			!{(-2,2) }*+{ ? }="n32"
			"n22":"n32"_(0.6){.07}^(0.6){3}
			!{(0,2) }*+{ ? }="n33"
			"n22":"n33"^(0.6){4}_(0.6){.18}
			!{(-1.95,2.29) }*+{\color{ForestGreen}\underbrace{\phantom{\;\quad\quad\qquad\qquad\qquad\qquad\qquad}}  }="n41"
			!{(-1.95,2.37) }*+{\phantom{.}}="n51"
			"n51" :@/^2.8cm/^{\textcolor{ForestGreen}{.18}}@[ForestGreen] "n2a"
		}
		\vspace{-6pt}
		\begin{center}
			\scriptsize 
			backtrack \#2, discard subtree $\textcircled{\scalebox{.7}{5}}$ (${\color{ForestGreen}F(\node_1) = 0.18} < {\color{red}0.25 = \bnd}$)
		\end{center}
		\vspace{-10pt}
		
		\hdashrule[0.5ex]{\columnwidth}{1pt}{3mm} 
%
		\xygraph{
			!{<0cm,0cm>;<1cm,0cm>:<0cm,-1cm>::}
			!{(-2,0) }*+{ \stackrel{\textcolor{red}{-\infty}}{\textcircled{\scriptsize 1}\langle1,2,5\rangle^{C}} }="n1"
			!{(1.2,.5) }*+{ \stackrel{\textcolor{red}{.18}}{\textcircled{\scriptsize 6}\langle2,4,6\rangle^{R}} }="n21"
			"n1":"n21"_{.25}^{5}
			!{(-2,1) }*+{? }="n22"
			"n1":"n22"^(0.6){2}_(0.6){\textcolor{ForestGreen}{.18}}
			!{(-4,1) }*+{ ? }="n23"
			"n1":"n23"_(0.65){\textcolor{ForestGreen}{.09}}^(0.6){1}
			!{(1.62,.2) }*+{ \phantom{.} }="n2a"
			!{(-0.7,1.7) }*+{ ? }="n31"
			"n21":"n31"_(0.7){.06}^(0.6){2}
			!{(1.2,1.7) }*+{ \underset{\color{ForestGreen}-\infty}{\textcircled{\scriptsize 7}\checkmark_{(\md_3)}} }="n32"
			"n21":"n32"_(0.5){.18}^(0.5){4}
			!{(2.8,1.7) }*+{ ? }="n33"
			"n21":"n33"^(0.5){.17}_(0.5){6}
			!{(-3,1.9) }*+{ \boxed{\scriptstyle \mD = [ [1,4],[1,6],[5,4] ] }}="key"
			!{(1.25,2.05) }*+{\color{ForestGreen}\underbrace{\phantom{\qquad\qquad\qquad\qquad\qquad\qquad}}  }="n41"
			!{(1.25,2.15) }*+{\phantom{.}}="n51"
			"n51" :@/_2.8cm/^{\textcolor{ForestGreen}{.17}}@[ForestGreen] "n2a"
		}
		\vspace{-6pt}
		\begin{center}
			\scriptsize 
			backtrack \#3, discard subtree $\textcircled{\scalebox{.7}{6}}$ (${\color{ForestGreen}F(\node_1) = 0.17} < {\color{red}0.18 = \bnd}$)
		\end{center}
		\vspace{-10pt}
		
		\hdashrule[0.5ex]{\columnwidth}{1pt}{3mm} 
%
		\xygraph{
			!{<0cm,0cm>;<1cm,0cm>:<0cm,-1.2cm>::}
			!{(-2,0) }*+{ \stackrel{\textcolor{red}{-\infty}}{\textcircled{\scriptsize 1}\langle1,2,5\rangle^{C}} }="n1"
			!{(-5,.5) }*+{ ? }="n21"
			"n1":"n21"_{\textcolor{ForestGreen}{.09}}^{1}
			!{(-2,1) }*+{ \stackrel{\textcolor{red}{.17}}{\textcircled{\scriptsize 8}\langle1,3,4\rangle^{R}} }="n22"
			"n1":"n22"^(0.5){2}_(0.5){\textcolor{ForestGreen}{.18}}
			!{(1,.5) }*+{ ? }="n23"
			"n1":"n23"_{\textcolor{ForestGreen}{.17}}^(0.6){5}
			!{(-0.2,1.8) }*+{ \phantom{.} }="n2a"
			!{(-4,2) }*+{ ? }="n31"
			"n22":"n31"_(0.7){.09}^(0.6){1}
			!{(-2,2) }*+{ ? }="n32"
			"n22":"n32"_(0.6){.07}^(0.6){3}
			!{(0,2) }*+{ \stackrel{\textcolor{red}{.17}}{\textcircled{\scriptsize 9}\langle1,5,6,7\rangle^{C}} }="n33"
			"n22":"n33"^(0.5){4}_(0.5){.18}
			!{(-1,3.22) }*+{\color{ForestGreen}\aunderbrace[l8D4r]{\phantom{\hspace{180pt}}}  }="n4a"
			!{(-4,3) }*+{ ? }="n41"
			"n33":"n41"^(0.8){1}_(0.8){.06}
			!{(-1.5,3) }*+{ ? }="n42"
			"n33":"n42"^(0.7){5}_(0.8){.04}
			!{(0,3) }*+{ ? }="n43"
			"n33":"n43"^(0.6){6}_(0.6){.11}
			!{(2,3) }*+{ ? }="n44"
			"n33":"n44"^(0.6){7}_(0.6){.04}
			!{(0.05,3.33) }*+{\phantom{.}}="n5a"
			"n5a" :@/^5cm/_{\textcolor{ForestGreen}{.11}}@[ForestGreen] "n2a"
		}
		\vspace{-5pt}
		\begin{center}
			\scriptsize 
			backtrack \#4, discard subtree $\textcircled{\scalebox{.7}{9}}$ (${\color{ForestGreen}F(\node_1) = 0.11} < {\color{red}0.17 = \bnd}$)
		\end{center}
		\vspace{-10pt}
		
		\hdashrule[0.5ex]{\columnwidth}{1pt}{3mm} 
	
	\caption{RBF-HS executed on the example DPI (part I), cf.\ Example~\ref{ex:RBF-HS}.}
	\label{fig:rbfhs_example_part1}
\end{figure}

\begin{figure}[t!]
	\centering
	\phantom{.}
		\xygraph{
			!{<0cm,0cm>;<1cm,0cm>:<0cm,-1.2cm>::}
			!{(-2,0) }*+{ \stackrel{\textcolor{red}{-\infty}}{\textcircled{\scriptsize 1}\langle1,2,5\rangle^{C}} }="n1"
			!{(-5,.5) }*+{ ? }="n21"
			"n1":"n21"_{\textcolor{ForestGreen}{.09}}^{1}
			!{(-2,1) }*+{ \stackrel{\textcolor{red}{.17}}{\textcircled{\scriptsize 8}\langle1,3,4\rangle^{R}} }="n22"
			"n1":"n22"^(0.5){2}_(0.5){\textcolor{ForestGreen}{.18}}
			!{(1,.5) }*+{ ? }="n23"
			"n1":"n23"_{\textcolor{ForestGreen}{.17}}^(0.6){5}
			!{(-2.3,0.87) }*+{ \phantom{.} }="n2a"
			!{(-4,2) }*+{ ? }="n31"
			"n22":"n31"_(0.7){.09}^(0.6){1}
			!{(-2,2) }*+{ ? }="n32"
			"n22":"n32"_(0.6){.07}^(0.6){3}
			!{(0,2) }*+{ ? }="n33"
			"n22":"n33"^(0.6){4}_(0.6){\textcolor{ForestGreen}{.11}}
			!{(-1.95,2.29) }*+{\color{ForestGreen}\underbrace{\phantom{\;\quad\quad\qquad\qquad\qquad\qquad\qquad}}  }="n41"
			!{(-1.95,2.37) }*+{\phantom{.}}="n51"
			"n51" :@/^2.8cm/^{\textcolor{ForestGreen}{.11}}@[ForestGreen] "n2a"
		}
		\vspace{-4pt}
		\begin{center}
			\scriptsize 
			backtrack \#5, discard subtree $\textcircled{\scalebox{.7}{8}}$ (${\color{ForestGreen}F(\node_1) = 0.11} < {\color{red}0.17 = \bnd}$)
		\end{center}
		\vspace{-10pt}
		
		\hdashrule[0.5ex]{\columnwidth}{1pt}{3mm} 
%
		\xygraph{
			!{<0cm,0cm>;<1cm,0cm>:<0cm,-1cm>::}
			!{(-2,0) }*+{ \stackrel{\textcolor{red}{-\infty}}{\textcircled{\scriptsize 1}\langle1,2,5\rangle^{C}} }="n1"
			!{(1.2,.5) }*+{ \stackrel{\textcolor{red}{.11}}{\textcircled{\tiny 10}\langle2,4,6\rangle^{R}} }="n21"
			"n1":"n21"_{\textcolor{ForestGreen}{.17}}^{5}
			!{(-2,1) }*+{? }="n22"
			"n1":"n22"^(0.6){2}_(0.6){\textcolor{ForestGreen}{.11}}
			!{(-4,1) }*+{ ? }="n23"
			"n1":"n23"_(0.65){\textcolor{ForestGreen}{.09}}^(0.6){1}
			!{(2.9,1.4) }*+{ \phantom{.} }="n2a"
			!{(-0.7,1.7) }*+{ ? }="n31"
			"n21":"n31"_(0.7){.06}^(0.6){2}
			!{(1.2,1.7) }*+{ \underset{\color{ForestGreen}-\infty}{\textcircled{\tiny 11}\times_{(=\md_3)}} }="n32"
			"n21":"n32"_(0.5){.18}^(0.5){4}
			!{(2.8,1.7) }*+{ \stackrel{\textcolor{red}{.11}}{\textcircled{\tiny 12}\langle1,3,4\rangle^{R}} }="n33"
			"n21":"n33"^(0.5){.17}_(0.5){6}
			!{(-0.2,3) }*+{ ? }="n41"
			"n33":"n41"^(0.7){1}_(0.8){.06}	
			!{(1.6,3) }*+{ ? }="n42"
			"n33":"n42"^(0.7){3}_(0.8){.04}
			!{(3.2,3) }*+{ \underset{\color{ForestGreen}-\infty}{\textcircled{\tiny 13}\times_{(\supset\md_3)}} }="n43"
			"n33":"n43"^(0.5){.11}_(0.5){4}						
			!{(1.8,3.35) }*+{\color{ForestGreen}\aunderbrace[l7D4r]{\phantom{\;\qquad\qquad\qquad\qquad\qquad\qquad}}  }="n41"
			!{(2.2,3.5) }*+{\phantom{.}}="n51"
			"n51" :@/_2.1cm/_(0.82){\textcolor{ForestGreen}{.06}}@[ForestGreen] "n2a"
		}
		\vspace{-4pt}
		\begin{center}
			\scriptsize 
			backtrack \#6, discard subtree $\textcircled{\scalebox{.57}{12}}$ (${\color{ForestGreen}F(\node_1) = 0.06} < {\color{red}0.11 = \bnd}$)
		\end{center}
		\vspace{-10pt}
		
		\hdashrule[0.5ex]{\columnwidth}{1pt}{3mm} 
%
		\xygraph{
			!{<0cm,0cm>;<1cm,0cm>:<0cm,-1cm>::}
			!{(-2,0) }*+{ \stackrel{\textcolor{red}{-\infty}}{\textcircled{\scriptsize 1}\langle1,2,5\rangle^{C}} }="n1"
			!{(1.2,.5) }*+{ \stackrel{\textcolor{red}{.11}}{\textcircled{\tiny 10}\langle2,4,6\rangle^{R}} }="n21"
			"n1":"n21"_{\textcolor{ForestGreen}{.17}}^{5}
			!{(-2,1) }*+{? }="n22"
			"n1":"n22"^(0.6){2}_(0.6){\textcolor{ForestGreen}{.11}}
			!{(-4,1) }*+{ ? }="n23"
			"n1":"n23"_(0.65){\textcolor{ForestGreen}{.09}}^(0.6){1}
			!{(1.75,0.15) }*+{ \phantom{.} }="n2a"
			!{(-0.7,1.7) }*+{ ? }="n31"
			"n21":"n31"_(0.7){.06}^(0.6){2}
			!{(1.2,1.7) }*+{ \underset{\color{ForestGreen}-\infty}{\textcircled{\tiny 11}\times_{(=\md_3)}} }="n32"
			"n21":"n32"_(0.5){.18}^(0.5){4}
			!{(2.8,1.7) }*+{ ? }="n33"
			"n21":"n33"^(0.65){6}_(0.5){\textcolor{ForestGreen}{.06}}					
			!{(1.05,2.05) }*+{\color{ForestGreen}\aunderbrace[l4D5r]{\phantom{\quad\qquad\qquad\qquad\qquad\qquad}}  }="n41"
			!{(0.8,2.2) }*+{\phantom{.}}="n51"
			"n51" :@/_2cm/_(0.65){\textcolor{ForestGreen}{.06}}@[ForestGreen] "n2a"
		}
		\vspace{-4pt}
		\begin{center}
			\scriptsize 
			backtrack \#7, discard subtree $\textcircled{\scalebox{.57}{10}}$ (${\color{ForestGreen}F(\node_1) = 0.06} < {\color{red}0.11 = \bnd}$)
		\end{center}
		\vspace{-10pt}
		
		\hdashrule[0.5ex]{\columnwidth}{1pt}{3mm} 
%
		\xygraph{
			!{<0cm,0cm>;<1cm,0cm>:<0cm,-1.2cm>::}
			!{(-2,0) }*+{ \stackrel{\textcolor{red}{-\infty}}{\textcircled{\scriptsize 1}\langle1,2,5\rangle^{C}} }="n1"
			!{(-5,.5) }*+{ ? }="n21"
			"n1":"n21"_{\textcolor{ForestGreen}{.09}}^{1}
			!{(-2,1) }*+{ \stackrel{\textcolor{red}{.09}}{\textcircled{\tiny 14}\langle1,3,4\rangle^{R}} }="n22"
			"n1":"n22"^(0.5){\textcolor{ForestGreen}{.11}}_(0.5){2}
			!{(1,.5) }*+{ ? }="n23"
			"n1":"n23"_{\textcolor{ForestGreen}{.06}}^(0.6){5}
			!{(-0.2,1.8) }*+{ \phantom{.} }="n2a"
			!{(-1.65,0.53) }*+{ \phantom{.} }="n1.5a"
			!{(-0.8,1.3) }*+{ \phantom{.} }="n2.5a"
			!{(-4,2) }*+{ ? }="n31"
			"n22":"n31"_(0.7){.09}^(0.6){1}
			!{(-2,2) }*+{ ? }="n32"
			"n22":"n32"_(0.6){.07}^(0.6){3}
			!{(0,2) }*+{ \stackrel{\textcolor{red}{.09}}{\textcircled{\tiny 15}\langle1,5,6,7\rangle^{C}} }="n33"
			"n22":"n33"^(0.5){\textcolor{brown}{.11}}_(0.5){4}
			%
			!{(-4,3) }*+{ ? }="n41"
			"n33":"n41"^(0.8){1}_(0.8){.06}
			!{(-1.5,3) }*+{ ? }="n42"
			"n33":"n42"^(0.7){5}_(0.8){.04}
			!{(0,3) }*+{ \textcircled{\tiny 16}\checkmark_{(\md_4)} }="n43"
			"n33":"n43"^(0.6){6}_(0.6){.11}
			!{(2,3) }*+{ ? }="n44"
			"n33":"n44"^(0.6){7}_(0.6){.04}
			!{(1.6,1.3) }*+{ \boxed{\scriptstyle \mD = [ [1,4],[1,6],[5,4],[2,4,6] ] }}="key"
			"n1.5a" :@/^0.4cm/^(0.7){\text{\textcolor{brown}{inherit}}}@[brown] "n2.5a"
		}
		\vspace{-2pt}
		\begin{center}
			\scriptsize 
			exit procedure ($|\mD| = 4 \geq 4 = \ld$) $\quad\Rightarrow\quad$ return $\mD$
		\end{center}
		\vspace{-10pt}
		
		\hdashrule[0.5ex]{\columnwidth}{1pt}{3mm} 
	\caption{RBF-HS executed on the example DPI (part II), cf.\ Example~\ref{ex:RBF-HS}.}
	\label{fig:rbfhs_example_part2}
\end{figure}

\subsection{RBF-HS Exemplification}
The following example illustrates the workings of RBF-HS.
\begin{example} \hspace{-1em}\emph{(RBF-HS)}\quad\label{ex:RBF-HS}
	\noindent\emph{Inputs.} Consider a defective system with seven components, described by 
	$\dpi := \langle\mo,\mb,\Tp,\Tn\rangle$, where $\mo = \{\tax_1,\dots,\tax_7\}$ 
	and no background knowledge or any positive or negative measurements are initially given,
	i.e., $\mb,\Tp,\Tn = \emptyset$.
	Let $\langle \pr(\tax_1), \dots, \pr(\tax_7)\rangle := \langle.26,.18,.21,.41,.18,.40,.18\rangle$ (note: $\pr$ is strictly antimonotonic since the probability of each $\tax_i\in\mo$ is less than $.5$, cf.\ Sec.~\ref{sec:probability_model}). In addition, let all minimal conflicts for $\dpi$ be $\langle\tax_1,\tax_2,\tax_5\rangle$, $\langle\tax_2,\tax_4,\tax_6\rangle$, $\langle\tax_1,\tax_3$, $\tax_4\rangle$, and $\langle\tax_1,\tax_5,\tax_6, \tax_7\rangle$. Assume we want to use RBF-HS to find the $\ld := 4$ most probable diagnoses for $\dpi$ (e.g., because we surmise the actual diagnosis to be amongst the most likely candidates). To this end, 
	$\dpi$, $\pr$ and $\ld$ are passed to RBF-HS (Alg.~\ref{algo:RBF_HS}) as input arguments.\vspace{5pt}
	
	\noindent\emph{Illustration (Figures).} The way of proceeding of RBF-HS is depicted by Figs.~\ref{fig:rbfhs_example_part1} and \ref{fig:rbfhs_example_part2}. In the figures, we use the following notation. 
	Axioms $\tax_i$ are simply referred to by $i$ (in node and edge labels). Numbers $\textcircled{\scriptsize k}$ indicate the chronological node labeling (expansion) order. Recall that nodes in Alg.~\ref{algo:RBF_HS} are sets of (integer) edge labels along tree branches. E.g., node $\textcircled{\scriptsize 9}$ in Fig.~\ref{fig:rbfhs_example_part1} corresponds to the node $\node = \{\tax_2,\tax_4\}$, i.e., to the assumption that components $c_2,c_4$ are at fault whereas all others are working properly. The probability $\pr(\node)$ (i.e., the original $f$-value) of a node $\node$ is shown by the black number from the interval $(0,1)$ that labels the edge pointing to $\node$, e.g., the cost of node $\textcircled{\scriptsize 9}$ is $0.18$. 
	We tag minimal conflicts $\tuple{\dots}$ that label internal nodes by $^C$ if they are freshly computed (\emph{expensive}; \textsc{findMinConflict} call, line~\ref{algoline:label:findMinConflict}),
	and 
	by $^R$ if they result from a reuse of some already computed and stored (see list $\mC$ in Alg.~\ref{algo:RBF_HS}) minimal conflict
	(\emph{cheap}; reuse label check; lines~\ref{algoline:label:reuse_start}--\ref{algoline:label:reuse_end}).
	%
	Leaf nodes are labeled as follows:
	``$?$'' is used for open (i.e., generated, but not yet labeled) nodes;
	$\checkmark_{(\md_i)}$ for a node 
	labeled $\valid$, i.e., a minimal diagnosis named $\md_i$, that is not yet stored in $\mD$;
	$\times_{(\mathit{Expl})}$ for a node labeled $\closed$, i.e., one that constitutes a non-minimal diagnosis or a diagnosis that has already been found and stored in $\mD$; $\mathit{Expl}$ is an explanation for the non-minimality in the former, and for the redundancy of node in the latter case, i.e., $\mathit{Expl}$ names a minimal diagnosis in $\mD$ that is a proper subset of the node, or it names a diagnosis in $\mD$ which is equal to node, respectively. Whenever a new diagnosis is 
	added to $\mD$ (line~\ref{algoline:rbfhs:add_node_to_mD}), this is displayed in the figures by a box that shows the current state of $\mD$. 
	For each expanded node, the value of the $\bnd$ variable relevant to the subtree rooted at this node is denoted by a red-colored value above the node. By green color, we show the backed-up $F$-value returned in the course of each backtracking step 
	(i.e., the best known probability of any node in the respective subtree). Further, $f$-values that have been updated by backed-up $F$-values are signalized by {green-colored} edge labels, see, e.g., in Fig.~\ref{fig:rbfhs_example_part1}, the left edge emanating from the root node of the tree has been reduced from $0.41$ ($f$-value) to $0.09$ ($F$-value) after the first backtrack. Finally, $F$-values of parents inherited by child nodes 
	(line~\ref{algoline:rbfhs':F(n_i)_gets_min}) are indicated by brown color, see the edge between node $\textcircled{\tiny 14}$ and node $\textcircled{\tiny 15}$ in Fig.~\ref{fig:rbfhs_example_part2}.\vspace{5pt}
	
	\noindent\emph{Discussion and Remarks.}
	Initially, RBF-HS starts with an empty root node, labels it with the minimal conflict $\tuple{1,2,5}$ at step $\textcircled{\scriptsize 1}$, generates the three corresponding child nodes $\{1\}, \{2\}, \{5\}$ shown by the edges originating from the root node, and recursively processes the best child node 
	(left edge, $f$-value $0.41$) at step $\textcircled{\scriptsize 2}$. The $\bnd$ for the subtree rooted at node $\textcircled{\scriptsize 2}$ corresponds to the best edge label ($F$-value) of any open node other than node $\textcircled{\scriptsize 2}$, which is $0.25$ in this 
	case. In a similar manner, the next recursive step is taken in that the best child node of node $\textcircled{\scriptsize 2}$ with an $F$-value not less than $\bnd = 0.25$ is processed. This leads to the labeling of node $\{1,4\}$ with $F$-value $0.28 \geq \bnd$ at step $\textcircled{\scriptsize 3}$, which reveals the first (proven most probable) diagnosis $\md_1 := [1,4]$ with $\pr(\md_1) = 0.28$, which is added to the solution list $\mD$. Note that $-\infty$ is at the same time returned for node $\textcircled{\scriptsize 3}$ (which indicates that the node has already been explored and ensures that the next best node has now the highest $F$-value).
	After the next node has been processed and the second-most-probable minimal diagnosis $\md_2 := [1,6]$ 
	with $\pr(\md_2) = 0.27$
	has been detected, the by now best remaining child node of node $\textcircled{\scriptsize 2}$ has an $F$-value of $0.09$ (leftmost node). This value, however, is lower than $\bnd$. Due to the best-first property of RBF-HS, this node is not explored right away because $\bnd$ suggests that there are more promising unexplored nodes elsewhere in the tree which have to be checked first. To keep the memory requirements linear, the current subtree rooted at node $\textcircled{\scriptsize 2}$ is discarded before a new one is examined. Hence, the first backtrack is executed. This involves 
	the storage of the best (currently known) $F$-value of any node in the subtree as the backed-up $F$-value of node $\textcircled{\scriptsize 2}$. This newly ``learned'' $F$-value is signalized by the green number ($0.09$) that by now labels the left edge emanating from the root. 
	Analogously, RBF-HS proceeds for the other nodes, 
	whereas the used $\bnd$ value is always the best value among the $\bnd$ value of the parent and all sibling's $F$-values. 
	Please also observe the $F$-value inheritance that takes place when node $\{2,4\}$ is generated for the third time (node $\textcircled{\tiny 15}$, Fig.~\ref{fig:rbfhs_example_part2}). The reason for this is that the original $f$-value of $\{2,4\}$ is $0.18$ (see top of Fig.~\ref{fig:rbfhs_example_part1}), but the meanwhile ``learned'' $F$-value of its parent $\{2\}$ is $0.11$ and thus smaller.
	This means that $\{2,4\}$ must have already been explored and the \emph{de-facto} probability of any (minimal) diagnosis in the subtree rooted at $\{2,4\}$ must be less than or equal to $0.11$.\vspace{5pt}
	
	\noindent\emph{Output.} Finally, RBF-HS immediately terminates as soon as the $\ld$-th (in this case: fourth) minimal diagnosis $\md_4$ is located and added to $\mD$. The list $\mD$ of minimal diagnoses arranged in descending order of probability $\pr$ is returned.\qed
\end{example}

\subsection{RBF-HS Complexity}
\label{sec:algo_complexity}
The next theorem states the complexity of RBF-HS, derived in 
\ref{apx:complexity}. 
\begin{thm}[Complexity of RBF-HS]\textcolor{white}{.}\label{thm:complexity}
Let $\dpi = \langle\mo,\mb,\Tp,\Tn\rangle$ be an arbitrary DPI, $\ld$ the finite positive natural number of diagnoses to be computed, $n$ the number of nodes expanded by HS-Tree (without the duplicate criterion) for $\dpi$ and $\ld$, $t_{\mathit{CC}}$ the worst-case time of a consistency check for $\dpi$, $\mathbf{minC}$ the set of all minimal conflicts for $\dpi$, and $\mc_{\max}$ the conflict of maximal size for $\dpi$. Further, let $\mathit{TPT} := t_{\mathit{CC}} |\mo|(|\mathbf{minC}| + |\ld|)$ (theorem proving time).
Finally, assume $|\mathbf{minC}|$ is in $O(1)$, i.e., independent of the size of $\dpi$.
Then:
\begin{itemize}[noitemsep]
	\item \emph{Time Complexity:} 
	RBF-HS requires time in $O(n + \mathit{TPT})$ for the computation of the $\ld$ diagnoses of minimal cardinality for $\dpi$; and time in $O(n^2 + \mathit{TPT})$ for the computation of the $\ld$ most probable diagnoses for $\dpi$. 
	\item \emph{Space Complexity:} RBF-HS requires space in $O(|\mo|)$.
\end{itemize}
\end{thm}

%
%
%

\subsection{RBF-HS Correctness}
\label{sec:algo_correctness}
The next theorem shows that RBF-HS is correct. A proof is given in 
\ref{apx:proof}.
\begin{thm}[Correctness of RBF-HS]\label{thm:correctness}
	Let \textsc{findMinConflict} be a sound and complete method for conflict computation, i.e., given a DPI, it outputs a minimal conflict for this DPI if a minimal conflict exists, and `no conflict' otherwise.
	RBF-HS is sound, complete and best-first, i.e., it 
	computes \emph{all} and \emph{only} minimal diagnoses \emph{in descending order} \emph{of probability} as per the strictly antimonotonic probability measure $\pr$.
\end{thm}

\subsection{RBF-HS: Potential Impact and Synergies with Other Techniques}
\label{sec:impact}
Beside RBF-HS's direct usage 
\begin{itemize}[noitemsep,topsep=0pt]
	\item \emph{as a space-efficient alternative} to (exponential-space) best-first diagnosis search algorithms such as HS-Tree \cite{Reiter87}, HST tree \cite{wotawa2001variant}, DynamicHS \cite{rodler2020ecai}, GDE \cite{dekleer1987}, or StaticHS \cite{rodler2018statichs}, or
	\item \emph{as a best-first alternative} to sound and complete linear-space \emph{any-first} searches like Inv-HS-Tree \cite{Shchekotykhin2014}, or
	\item \emph{as a complete alternative} to best-first, but incomplete algorithms like CDA* \cite{williams2007conflict} or STACCATO \cite{abreu2009low}, 
\end{itemize}
several uses of RBF-HS combined with existing techniques can be conceived of. We briefly sketch some of them next, before we discuss a hybrid method that combines HS-Tree \cite{Reiter87} and RBF-HS in more detail in the next section:
\begin{enumerate}[itemsep=3pt,label=\emph{(\Alph*)},align=left,labelwidth=-0.5\parindent]
	\item \emph{Informed HS-Tree:} 
	The idea is to run RBF-HS as a preprocessor in order to provide more informed node probabilities, and to subsequently adopt HS-Tree using these ``learned'' probabilities as $f$-values.
	To this end, e.g., RBF-HS could be executed with a fixed time limit and modified to store backed-up $F$-values of 
	(a subset of) the visited nodes---not only of the ones that are kept in memory after backtracking steps. Like a heuristic for classic A*, this additional ``lookahead'' information might lead to the finding of the preferred diagnoses by expanding significantly fewer nodes.
\item \emph{RBF-HS as a Decision Heuristic:} 
The rationale is to run RBF-HS for a certain limited time and to afterwards take the ``learned'' $F$-value(s) as an estimate of the hardness or some other relevant property of the diagnosis problem. 
Depending on how the node costs are set (cf.\ Sec.~\ref{sec:inputs+outputs}), the backed-up $F$-value can provide an estimation of the least depth of the search tree, i.e., of the least size of minimum-cardinality diagnoses, or an upper bound estimate of the probability of the minimal diagnoses. 
%
Such an estimate can then be used, e.g.:
\begin{itemize}[noitemsep,topsep=0pt]
	\item \emph{To decide which algorithm to use}, e.g., whether to drop some nice-to-have requirement(s) to the adopted diagnosis computation algorithm (such as completeness or the best-first property) in order to keep performance reasonable (cf., e.g., \cite{Shchekotykhin2014}).
	\item \emph{For an informed selection of a limit for depth-limited or cost-limited search} \cite{russellnorvig2010} (cf.\ Example~\ref{ex:search_algorithms}). When using a suitable limit, 
	the latter can be powerful linear-space strategies to find the preferred diagnoses, and might be substantially faster than iterative deepening, IDA* (hitting set) searches and {RBF-HS}.
\end{itemize}
	\item \emph{RBF-HS as a Plug-In:} 
	Given a diagnosis search method that 
	uses a hitting set generation routine as a black-box, such as SDE \cite{stern2012}, RBF-HS can be used as a plug-in, e.g., in case memory issues are faced when using other best-first algorithms. 
\end{enumerate}

\section{Hybrid Best-First Hitting Set Search (HBF-HS)}
\label{sec:HBF-HS}
In this section, we propose a hybrid technique called HBF-HS, shown by Alg.~\ref{algo:HBF_HS}, that aims at combining the advantages of the more space-attractive RBF-HS with those of the more time-attractive HS-Tree.
\subsection{HBF-HS Algorithm Description}
The goal of HBF-HS is to allow for an as fast as possible diagnosis search while preserving soundness, completeness and best-firstness also in cases where state-of-the-art searches boasting these three properties (e.g., HS-Tree) run out of memory. Given a so-called switch criterion $\mathit{stop_{\mathsf{HS}}}$ and otherwise the same inputs as RBF-HS, i.e., a DPI $\dpi$, a probability measure $\pr$, and a number $\ld$ of leading diagnoses to be computed, the principle of HBF-HS is as follows: Initially (line~\ref{algoline:hbfhs:HS-Tree_call}), execute standard HS-Tree, as described in Example~\ref{ex:Reiter's_HS_Node_Processing}. 
If the switch criterion $\mathit{stop_{\mathsf{HS}}}$ (e.g., 
a maximal amount or fraction of memory consumed) is met, then the algorithm checks if sufficient (at least $\ld$) or all (indicated by an empty queue) minimal diagnoses have already been computed by HS-Tree (line~\ref{algoline:hbfhs:if_mD_HS_geq_ld}). If so, the collection of minimal diagnoses is returned (line~\ref{algoline:hbfhs:return_mD_HS}). Otherwise, a switch to RBF-HS is prompted.
The recursive procedure RBF-HS' of the latter then continues the search (line~\ref{algoline:hbfhs:call_RBFHS'}) while only consuming a linear amount of additional memory. In this vein, HS-Tree can utilize as much memory as it needs while executing (\emph{focus on time optimization}), and, before the available memory is depleted, RBF-HS takes over (\emph{focus on space optimization}) so that the problem remains solvable. 

The transfer of control between HS-Tree and RBF-HS' is rather straightforward while guaranteeing the retention of soundness, completeness and best-first properties. 
The idea is to extract the relevant information from the search tree produced by HS-Tree and use it to set up a new search tree, on which RBF-HS' can start operating. Specifically, after HS-Tree is stopped, the \emph{switch process} (lines~\ref{algoline:hbfhs:make_virtual_root}--\ref{algoline:hbfhs:call_RBFHS'}) is as follows:
\begin{enumerate}[noitemsep,leftmargin=22pt,label=\emph{(S\arabic*)}]
	\item \label{enum:switch-step:1} 
	Create a virtual root node $\node_0$ (line~\ref{algoline:hbfhs:make_virtual_root}); set the bound of $\node_0$ to $-\infty$ (line~\ref{algoline:hbfhs:call_RBFHS'}). As the $f$-value of the root node is irrelevant to RBF-HS and thus arbitrary, 
	set $f(\node_0) := 0$.
	\item \label{enum:switch-step:2} 
	View the queue $\Queue_{\mathsf{HS}}$ of open nodes of HS-Tree as child nodes $\childnodes_{\mathsf{root}}$ of the root $\node_0$. Beforehand, delete all duplicates from $\Queue_{\mathsf{HS}}$ (line~\ref{algoline:hbfhs:delete_duplicates}), e.g., if $\{1,2\}$ and $\{2,1\}$ is contained, delete one of these nodes. This prevents RBF-HS' (which does not involve duplicate checks, cf.\ Sec.~\ref{sec:recursion_structure}) from 
	doing redundant work.
	\item \label{enum:switch-step:3} Copy all minimal diagnoses ($\mD_{\mathsf{HS}}$) and minimal conflicts ($\mC_{\mathsf{HS}}$) already computed by HS-Tree to the collections $\mD$ and $\mC$ of RBF-HS, respectively (line~\ref{algoline:hbfhs:initialize_mD_mC_for_RBF-HS}).
\end{enumerate} 
Then, execute plain RBF-HS', where the procedures \textsc{label} and \textsc{expand} are slightly adapted to handle the virtual root node $\node_0$, 
see Alg.~\ref{algo:HBF_HS}. Finally, return $\mD$ (line~\ref{algoline:hbfhs:return_mD}).

\begin{algorithm}[!htbp]
	\scriptsize
	\caption{HBF-HS} \label{algo:HBF_HS}
	{\fontsize{7pt}{8pt}\selectfont
		\begin{algorithmic}[1]
			\Require 
			\textcolor{white}{.}
			tuple $\tuple{\dpi, \pr, \ld, \mathit{stop_{\mathsf{HS}}}}$ comprising
			\begin{itemize}[noitemsep]
				\item a DPI $\dpi = \langle\mo,\mb,\Tp,\Tn\rangle$
				\item a function $\pr$ that assigns a failure probability $\pr(\tax) \in (0,1)$ to each $\tax \in \mo$, where $\pr$ is strictly antimonotonic (cf.\ Sec.~\ref{sec:probability_model}); \quad \emph{note:} the cost function $f(\node) := \pr(\node)$ as per Eq.~\ref{eq:diag_prob} for all tree nodes $\node \subseteq \mo$
				\item the number $\ld$ of leading minimal diagnoses to be computed
				\item a stop criterion $\mathit{stop_{\mathsf{HS}}}$ for \textsc{HS-Tree} 
			\end{itemize}
			\Ensure 
			list $\mD$ where $\mD$ is the list of the $\ld$ (if existent) most probable (as per $\pr$) 
			minimal diagnoses wrt.\ $\dpi$, sorted by probability in descending order  
			\vspace{6pt}
			\Procedure{HBF-HS}{$\dpi, \pr, \ld, \mathit{stop_{\mathsf{HS}}}$}  \Comment{\textsc{HS-Tree} as described in Example~\ref{ex:Reiter's_HS_Node_Processing}...}
			\State $\langle\mD_{\mathsf{HS}},\mC_{\mathsf{HS}},\Queue_{\mathsf{HS}}\rangle \gets \Call{HS-Tree}{\dpi,\pr,\ld,\mathit{stop_{\mathsf{HS}}}}$ \label{algoline:hbfhs:HS-Tree_call} \Comment{...executes until 
				$\mathit{stop_{\mathsf{HS}}}$ is met}
			\Statex  
			\Comment{$\mD_{\mathsf{HS}}$ / $\mC_{\mathsf{HS}}$ / $\Queue_{\mathsf{HS}} = $ min diagnoses / min conflicts / node queue of \textsc{HS-Tree} when $\mathit{stop_{\mathsf{HS}}}$ is met}
			\If{$|\mD_{\mathsf{HS}}| \geq \ld \,\lor\, \mQ_{\mathsf{HS}} = [\,]$}  
			\label{algoline:hbfhs:if_mD_HS_geq_ld} \Comment{if sufficient ($\ld$) or all min diagnoses already computed by HS-Tree}
			\State \Return $\mD_{\mathsf{HS}}$ 
			\Comment{return diagnoses; no call of RBF-HS' needed}
			\label{algoline:hbfhs:return_mD_HS}	
			\EndIf
			\State $\node_0 \gets \Call{makeVirtualRootNode}$  \label{algoline:hbfhs:make_virtual_root} \Comment{create a virtual root node $\node_0$ and set $f(\node_0) := 0$}
			\State $\childnodes_{\mathsf{root}} \gets \Call{DeleteDuplicates}{\Queue_{\mathsf{HS}}}$ \label{algoline:hbfhs:delete_duplicates} \Comment{delete all duplicate nodes from $\Queue_{\mathsf{HS}}$} 
			\State $\mD \gets \mD_{\mathsf{HS}}, \; \mC \gets \mC_{\mathsf{HS}}$
			\label{algoline:hbfhs:initialize_mD_mC_for_RBF-HS} \Comment{initialize min diagnoses / min conflicts collections maintained by RBF-HS'}
			\State $\Call{RBF-HS'}{\node_0,f(\node_0),-\infty}$
			\label{algoline:hbfhs:call_RBFHS'} \Comment{procedure RBF-HS' in Alg.~\ref{algo:RBF_HS}}
			\State \Return $\mD$	\label{algoline:hbfhs:return_mD}							
			\EndProcedure
			\vspace{6pt}
			
			\Procedure{\textsc{label}}{$\node$}    \Comment{same as \textsc{label} in Alg.~\ref{algo:RBF_HS}, but handles special case where $\node =$ virtual root node}
			\State \label{algoline:hbfhs:comment_on_label_function} \Comment{check if $\node$ is the virtual root node; if so, return $L = \childnodes_{\mathsf{root}}$; else, execute 
				lines~\ref{algoline:label:non-min_crit_start}--\ref{algoline:label:return_new_cs}	of Alg.~\ref{algo:RBF_HS} \qquad \phantom{...}}
			\EndProcedure
			\vspace{6pt}
			\Procedure{\textsc{expand}}{$\node,\mc$}   \Comment{same as \textsc{expand} in Alg.~\ref{algo:RBF_HS}, but handles special case where $\node =$ virtual root node}
			\State $\mathsf{Succ\_Nodes} \gets [\,]$
			\For{$e \in \mc$} \label{algoline:hbfhs:expand:for_loop}
			\If{$\Call{isSet}{e}$} \label{algoline:hbfhs:expand:if_isSet} \Comment{$\node =$ virtual root node $\Rightarrow \mc = \childnodes_{\mathsf{root}}$, each element of which is \emph{a set}}
			\State $\mathsf{Succ\_Nodes} \gets \Call{add}{e,\mathsf{Succ\_Nodes}}$ \label{algoline:hbfhs:expand:add_succ_of_virtual_root} 
			\Else  \label{algoline:hbfhs:expand:else}	\Comment{$\node \neq$ virtual root node $\Rightarrow \mc$ is a min conflict, each element of which is \emph{not a set}}
			\State $\mathsf{Succ\_Nodes} \gets \Call{add}{\node\cup\{e\},\mathsf{Succ\_Nodes}}$ \label{algoline:hbfhs:expand:add_successor_node}
			\EndIf
			\EndFor
			\State \Return $\mathsf{Succ\_Nodes}$
			\EndProcedure	
		\end{algorithmic}
	}
	\normalsize
\end{algorithm}

\begin{figure}
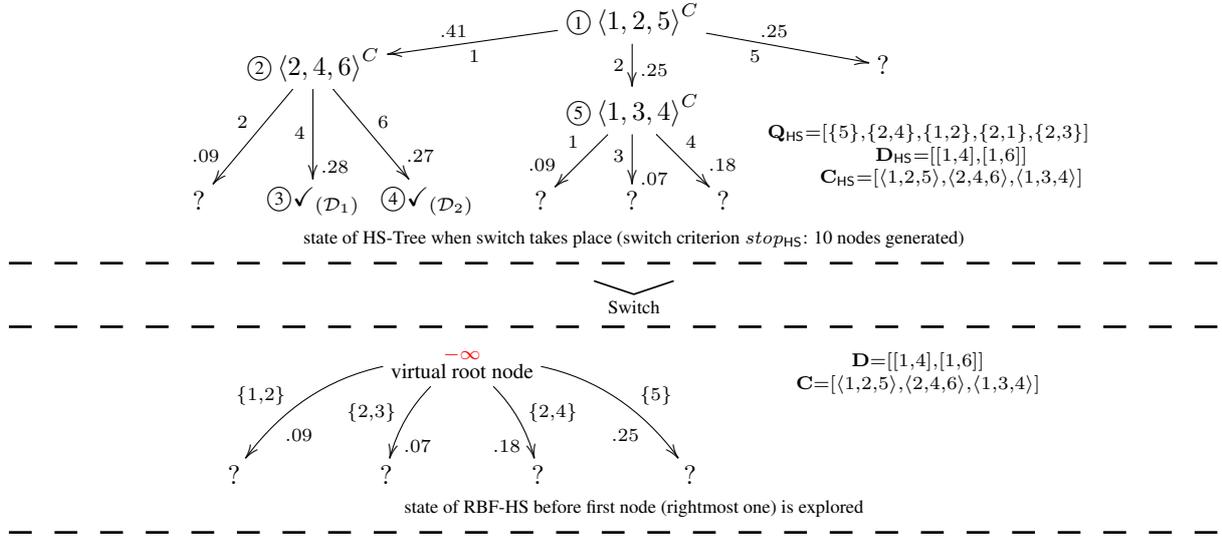

\centering
\phantom{.}
		\xygraph{
			!{<0cm,0cm>;<1.2cm,0cm>:<0cm,1.2cm>::}
			!{(4.75,4)}*+{\textcircled{\scriptsize 1}\tuple{1,2,5}^C}="c1c"
			!{(1.25,3.5) }*+{\textcircled{\scriptsize 2}\tuple{2,4,6}^C}="c2c" 
			!{(4.75,3) }*+{\textcircled{\scriptsize 5}\tuple{1,3,4}^C}="c3c" 
			!{(7.5,3.5) }*+{?}="c2r"
			!{(0,2) }*+{?}="d4"
			!{(1.25,2) }*+{\textcircled{\scriptsize 3}\checkmark_{(\md_1)}}="d1"
			!{(2.5,2) }*+{\textcircled{\scriptsize 4}\checkmark_{(\md_2)}}="d2"
			!{(3.75,2) }*+{?}="dup1"
			!{(4.75,2) }*+{?}="?2-2"
			!{(5.75,2) }*+{?}="c4c"
			"c1c":"c2c"^{1}_(0.55){.41}
			"c1c":"c3c"_{2}^(0.55){.25}
			"c1c":"c2r"_{5}^(0.55){.25}
			"c2c":"d4"_{2}_(0.75){.09}
			"c2c":"d1"_{4}^(0.75){.28}
			"c2c":"d2"^{6}^(0.75){.27}
			"c3c":"dup1"_{1}_(0.75){.09}
			"c3c":"?2-2"_{3}^(0.75){.07}
			"c3c":"c4c"^{4}^(0.75){.18}
			!{(8.25,2.5) }*+{ \scriptstyle \mD_{\mathsf{HS}} = [ [1,4],[1,6] ] }="diags"
			!{(8.25,2.25) }*+{ \scriptstyle \mC_{\mathsf{HS}} = [ \tuple{1,2,5},\tuple{2,4,6},\tuple{1,3,4} ] }="conflicts"
			!{(8,2.75) }*+{ \scriptstyle \mQ_{\mathsf{HS}} = [ \{5\},\{2,4\},\{1,2\},\{2,1\},\{2,3\} ] }="Q"
		}
		\vspace{-8pt} 
		\begin{center}
			\scriptsize state of HS-Tree when switch takes place (switch criterion $\mathit{stop_{\mathsf{HS}}}$: 10 nodes generated)
		\end{center}
	\vspace{-10pt}
	\hdashrule[0.5ex]{\columnwidth}{1pt}{3mm}
		\begin{minipage}[c]{\textwidth}
			\centering
				\rotatebox{270}{$\Bigg>$} \\ \vspace{0pt}
			\scriptsize Switch
		\end{minipage}
		\hdashrule[0.5ex]{\columnwidth}{1pt}{3mm} 
		\xygraph{
			!{<0cm,0cm>;<1cm,0cm>:<0cm,1cm>::}
			!{(4,4)}*+{\stackrel{\textcolor{red}{-\infty}}{\text{\footnotesize virtual root node}}}="c1c"
			!{(1,2.5) }*+{?}="c2c" 
			!{(3,2.5) }*+{?}="c3c" 
			!{(5,2.5) }*+{?}="c2r"
			!{(7,2.5) }*+{?}="c3r"
			!{(10,4) }*+{ \scriptstyle \mD = [ [1,4],[1,6] ] }="diags"
			!{(10,3.7) }*+{ \scriptstyle \mC = [ \tuple{1,2,5},\tuple{2,4,6},\tuple{1,3,4} ] }="conflicts"
			"c1c":@/_0.5cm/_(0.7){\{1,2\}}^(0.75){.09} "c2c"
			"c1c":@/_0.2cm/_(0.6){\{2,3\}}^(0.75){.07} "c3c"
			"c1c":@/^0.2cm/^(0.6){\{2,4\}}_(0.75){.18} "c2r"
			"c1c":@/^0.5cm/^(0.7){\{5\}}_(0.75){.25}"c3r"
		}
		\vspace{-8pt}
		\begin{center}
			\scriptsize state of RBF-HS before first node (rightmost one) is explored
		\end{center}
	\vspace{-10pt}
	\hdashrule[0.5ex]{\columnwidth}{1pt}{3mm}
	\caption{Sketch of the execution of HBF-HS on DPI from Example~\ref{ex:RBF-HS}
		\label{fig:ex_HBF-HS}
	}
\end{figure}

\subsection{HBF-HS Exemplification}
The following example illustrates the workings of HBF-HS.	
	
	\begin{example} \hspace{-1em}\emph{(HBF-HS)}\quad\label{ex:HBF-HS}
		Let us reconsider the DPI introduced in Example~\ref{ex:RBF-HS} and have a look how HBF-HS would proceed for it. Assume the switch criterion $\mathit{stop_{\mathsf{HS}}}$ is defined as ``ten generated nodes''. Specifically, this means: Execute HS-Tree until ten nodes are generated, then execute steps \ref{enum:switch-step:1}--\ref{enum:switch-step:3}, and finally run RBF-HS'. Fig.~\ref{fig:ex_HBF-HS} shows at the top the end state of HS-Tree before the switch is performed, and at the bottom the state of the transformed tree on which RBF-HS' will begin its operations. 
		Observe the following things:
		\begin{itemize}
			\item At the time the switch takes place, ten nodes 
			have been generated
			and seven nodes are currently maintained by HS-Tree, encompassing i.a.\ 
			five open nodes (``?'') stored in the queue $\Queue_{\mathsf{HS}}$. Note, two of these nodes, the leftmost and fourth-leftmost one, are equal (i.e., the path labels $\{1,2\}$ and $\{2,1\}$ coincide). Hence, one of them is a duplicate and does not need to be further considered (recall that diagnoses are \emph{sets} of edge labels). 
			Now, Step~\ref{enum:switch-step:1} of the switch process prompts the construction of a new tree through the generation of a virtual root node with $\bnd$ (red color) set to $-\infty$. Step~\ref{enum:switch-step:2} then effectuates the connection of this root node by one edge each to the four \emph{non-duplicate} open nodes in $\Queue_{\mathsf{HS}}$. The result is shown at the bottom of Fig.~\ref{fig:ex_HBF-HS}.  
			Observe that the labels of the edges \emph{emanating from the root node} are now \emph{sets of} elements from $\mo$. 
			Still, all labels for other edges non-linked to the root node are singletons, just as in plain RBF-HS
			(cf.\ Example~\ref{ex:RBF-HS}). Note, in RBF-HS we do not use set notation for edge labels simply because all these labels are \emph{single} elements from $\mo$ (cf.\ Figs.~\ref{fig:rbfhs_example_part1} and \ref{fig:rbfhs_example_part2}). 
			\item Until the switch, two minimal diagnoses have already been located by HS-Tree (nodes \textcircled{\scriptsize 3} and \textcircled{\scriptsize 4}; collection $\mD_{\textsf{HS}}$), and three minimal conflicts have been computed (node labels \textcircled{\scriptsize 1}, \textcircled{\scriptsize 2} and \textcircled{\scriptsize 5}; collection $\mC_{\mathsf{HS}}$). These are copied to the respective collections $\mD$ and $\mC$ maintained by RBF-HS in step \ref{enum:switch-step:3}, as depicted on the right in the bottom part of 
			Fig.~\ref{fig:ex_HBF-HS}. 
			\item The execution of RBF-HS' works exactly as discussed in Example~\ref{ex:RBF-HS}, with the difference that it starts with the partial hitting set tree displayed at the bottom of Fig.~\ref{fig:ex_HBF-HS}, where we have one root node, which has four elements in $\childnodes$. 
			That is, the first explored node would be the rightmost one, $\{5\}$, with the maximal $F$-value $0.25$ among $\childnodes$, and the $\bnd$ for the processing of $\{5\}$ would be $0.18$, the second-best $F$-value (of node $\{2,4\}$) among $\childnodes$. 
			Intuitively, the RBF-HS' execution in the course of HBF-HS can be regarded as a warm-start version of RBF-HS with some conflicts and open nodes, and potentially also some diagnoses, provided from the outset. \qed
		\end{itemize} 
	\end{example}

\subsection{HBF-HS Complexity}
\label{sec:hbfhs_complexity}
The complexity of HBF-HS depends largely on the used switch criterion $\mathit{stop_{\mathsf{HS}}}$, i.e., on how long HS-Tree runs until RBF-HS' continues. Unsurprisingly, the \emph{worst-case} complexity of HBF-HS corresponds to the worse complexity among RBF-HS and HS-Tree for both time and space.
In other words, the worst-case time complexity of HBF-HS equals the one of RBF-HS (if the switch takes place immediately and HS-Tree does not run at all), whereas its space complexity coincides with the one of HS-Tree (if the switch does not takes place until HS-Tree finishes executing). More formally:
\begin{thm}[Complexity of HBF-HS]
	Let $T$ be the worst-case time complexity of RBF-HS, and $S$ the worst-case space complexity of HS-Tree. Then, HBF-HS has a worst-case time complexity of $T$ and a worst-case space complexity of $S$.
\end{thm} 
Of greater practical interest than these extreme cases are more reasonable settings of $\mathit{stop_{\mathsf{HS}}}$. Since the bottleneck of HS-Tree is its (exponential) space complexity, and since it tends to exhibit better runtimes than RBF-HS (not only in the worst case, but also on average, as well shall see in Sec.~\ref{sec:eval}), 
the (only) appropriate strategy seems to be to condition the switch criterion on a measure of space consumed by HS-Tree rather than time. 
Intuitively, the longer it is affordable (wrt.\ memory) to run HS-Tree, the faster termination we might expect from HBF-HS, since its time complexity in this case will tend more towards the time complexity of HS-Tree. 
The material question then is, how much additional memory the execution of RBF-HS' will require \emph{after} the 
switch, or, how much memory one might safely concede to HS-Tree \emph{before} the switch. This question is answered by the following theorem, a direct corollary of Theorem~\ref{thm:complexity}: 
\begin{thm}[Space Complexity of HBF-HS after Switch]\label{thm:hbfhs:space_complexity_after_switch}
	Let the conditions of Theorem~\ref{thm:complexity} apply, and let $S$ be the amount of memory consumed by HS-Tree 
	until $\mathit{stop_{\mathsf{HS}}}$ is true. 
	Then, the additional memory beyond $S$ required by HBF-HS is in $O(|\mo|)$, i.e., linear. 
\end{thm}

\subsection{HBF-HS Correctness}
\label{sec:hbfhs_correctness}
The next theorem shows that HBF-HS is correct, and is proven in 
\ref{apx:hbfhs:correctness}.
\begin{thm}[Correctness of HBF-HS] \label{thm:hbfhs:correctness}
Let \textsc{findMinConflict} be a sound and complete method for conflict computation, i.e., given a DPI, it outputs a minimal conflict for this DPI if a minimal conflict exists, and `no conflict' otherwise. Further, let $\mathit{stop_{\mathsf{HS}}}$ be any predicate depending on the execution state of HS-Tree. Then, HBF-HS is sound, complete and best-first, i.e., it computes \emph{all} and \emph{only} minimal diagnoses \emph{in descending order} \emph{of probability} as per the strictly antimonotonic probability measure $\pr$.
\end{thm}

	\section{Related Work}
	\label{sec:related}
\subsection{Classifying Diagnosis Computation Methods}
\label{sec:classifying_diag_comp_methods}
Literature offers a wide variety of diagnosis computation algorithms, motivated by different diagnosis problems, domains and challenges. These algorithms can be compared along multiple dimensions, e.g.,\footnote{Citations per dimension 
	are not intended to be exhaustive. We rather tried to give \emph{some} representatives of each property and to give credit to (hopefully) most relevant works \emph{over all the discussed dimensions}.}
\begin{itemize}[noitemsep]
	\item \emph{best-first} (minimal diagnoses are output in order, most-preferred first, as per a given preference criterion)  \cite{Reiter87,dekleer1987,Rodler2015phd,greiner1989correction,rodler2018statichs,dekleer1991focusing_prob_diag} vs.\
	\emph{any-first} (no particular order on output 
	diagnoses can be guaranteed) \cite{lin2003computation,Shchekotykhin2014,pill2012optimizations},
	\item \emph{complete} (given sufficient runtime and memory, all minimal diagnoses are computed)
	\cite{Reiter87,dekleer1987,lin2003computation,greiner1989correction,Shchekotykhin2014,rodler2018statichs,stern2012,haenni1998generating,xiangfu2006method} vs.\
	\emph{incomplete} (in general, not all minimal diagnoses are found) \cite{abreu2009low,williams2007conflict,feldman2008computing,dekleer2009mininimum,li2002computing,siddiqi2007hierarchical},
	\item \emph{conflict-based} (minimal diagnoses are built as hitting sets of conflicts) \cite{Reiter87,dekleer1987,Rodler2015phd,wotawa2001variant,lin2003computation,greiner1989correction,rodler2018statichs,stern2012,xiangfu2006method,dekleer2011hitting} vs.\ 	\emph{direct} (minimal diagnoses are built without reliance on conflicts, e.g., through divide-and-conquer or compilation techniques)  \cite{Shchekotykhin2014,darwiche2001decomposable,felfernig2012efficient,metodi2014novel,torasso2006model},
	\item \emph{stateful} (state of the search data structure is maintained and reused throughout a diagnosis session, even if the diagnosis problem changes through the acquisition of new information about the diagnosed system) 
	\cite{dekleer1987,Rodler2015phd,rodler2018statichs,rodler2020ecai,Siddiqi2011} vs.\ 
	\emph{stateless} (whenever the diagnosis computation algorithm is called, it computes diagnoses by means of a fresh search data structure)  \cite{Reiter87,wotawa2001variant,greiner1989correction,xiangfu2006method,dekleer2011hitting},
	\item \emph{black-box} (the used theorem prover is seen as a pure oracle for consistency checks, which makes the diagnosis search very general in that no dependency on any particular logic or reasoning mechanism is given)  \cite{Reiter87,Rodler2015phd,wotawa2001variant,greiner1989correction,Shchekotykhin2014,rodler2018statichs} vs.\ 	\emph{glass-box} (the used theorem prover is internally optimized or modified for diagnostic purposes, which can bring performance gains, but makes the method reliant on one particular reasoning mechanism and on certain logics used to describe the diagnosed system)  \cite{Horridge2011a,Kalyanpur2006a,baader2008axiom_pinpointing,schlobach2007debugging},
	\item \emph{on-the-fly} (conflicts are computed on demand in the course of the diagnosis search) \cite{Reiter87,dekleer1987,wotawa2001variant,greiner1989correction,rodler2018statichs} vs.\ \emph{preliminary} (set of minimal conflicts must be known in advance and given as input to the diagnosis search) \cite{lin2003computation,abreu2009low,pill2012optimizations,xiangfu2006method,li2002computing,dekleer2011hitting},
	\item \emph{worst-case linear-space} (memory requirements linear in the problem size, even in the worst case) \cite{Shchekotykhin2014,felfernig2012efficient} vs.\ \emph{worst-case exponential-space} (memory requirements generally exponential in the problem size) \cite{Reiter87,dekleer1987,wotawa2001variant,lin2003computation,greiner1989correction,rodler2018statichs,rodler2020ecai,darwiche2001decomposable}. 
\end{itemize}
	\subsection{Towards Improving Existing Methods}
	\label{sec:towards_improving_existing_methods}
	Our study of these existing works suggests two different things. First, the best choice of algorithm, in general, depends largely on the particular tackled problem (domain and requirements). Consequently, there is little hope to find an algorithm that comes anywhere near improving 
	\emph{all} of the existing ones. Second, performance improvements for algorithms are often achieved at the cost of losing desirable properties (e.g., completeness or the best-first guarantee). Hence, it is particularly noteworthy 
	that RBF-HS as well as HBF-HS aim to improve existing sound, complete and best-first diagnosis search \emph{while preserving all these favorable properties}. Moreover, to the best of our knowledge, RBF-HS is the \emph{first} linear-space diagnosis computation method that ensures soundness, completeness and the best-first property. 
	
	Nevertheless, we next discuss the (classes of the) diagnosis computation methods that are most closely related to RBF-HS and HBF-HS wrt.\ the categorization above.  
	
	\subsection{Discussion of Related Works}
	In terms of the above-mentioned dimensions, RBF-HS and HBF-HS are best-first, complete, stateless, conflict-based, black-box, and on-the-fly. Moreover, RBF-HS is worst-case linear-space whereas HBF-HS is not. However, although no linear-space guarantee is given for HBF-HS, 
	the latter is nevertheless meant to be an improved variant of RBF-HS which ``is allowed'' to consume more than a linear amount of memory in order to reduce computation time.
	The diagnosis algorithms most closely related to the ones proposed in this work can be divided into \emph{compilation-based}, 
	\emph{duality-based}, 
	and \emph{best-first} ones, which we discuss in more detail next:\footnote{\ref{apx:related_works_in_heuristic_search} provides a review of other, more general memory-limited search algorithms that are related to RBF-HS and HBF-HS, but do not focus on diagnosis computation.} 
%
	
	\subsubsection{Compilation-Based Approaches}
	\noindent\emph{General differences to the proposed techniques:} These approaches are not black-box, i.e., dependent on the logic used to represent the diagnosed system. 
	They can be polynomial-space or linear-space, but only under certain circumstances.
	
	\noindent\emph{Discussion:} These techniques compile the diagnosis problem into some target representation such as SAT \cite{metodi2014novel}, OBDD \cite{torasso2006model} or DNNF \cite{darwiche2001decomposable}. Often, the generation of (minimum-cardinality; but not maximal-probability) diagnoses can be accomplished in worst-case polynomial time in the size of the respective compilation. For a polynomial-sized compilation, this implies polynomial-time diagnosis generation. However, the size of the compilation might be exponential in the size of the diagnosis problem for all these approaches, which means that no guarantee for polynomial-space (or polynomial-time), let alone linear-space, diagnosis generation can be given. 
	Second, for these compilation approaches to be applicable to a DPI, the diagnosed system must be amenable to a propositional-logic description, which is not always the case \cite{Shchekotykhin2012,Horridge2011a,Kalyanpur2006a}.
	Beyond that, compilation approaches usually do not allow to take influence on the exact order in which diagnoses are output. In summary, these methods are in general not linear-space, not best-first, 
	and not black-box. 
	
	A compilation-based approach that is based on abstraction techniques and especially suited for a sequential diagnosis \cite{dekleer1987} scenario is SDA \cite{Siddiqi2011}. One difference between RBF-HS and SDA is that only a single best diagnosis (instead of a set of best diagnoses) is output by SDA at the end of the sequential diagnosis process. Second, it 
	is questionable if similar abstraction-techniques as used in SDA are applicable to logics more expressive than propositional logic and to systems that are structurally different from typical circuit topologies.
	
	The authors of \cite{el1995diagnosing} present an approach that translates a circuit diagnosis problem into a constraint optimization problem. 
	If this constraint problem is amenable to a tree representation,
	then the minimum-cardinality diagnoses can be generated in linear time and space. However, it is unclear if and how non-circuit-problems and more expressive or other types of logics can be addressed. 
	
	%
	
	\subsubsection{Duality-Based Approaches}
	\label{sec:related:duality_based_approaches}
	\noindent\emph{General differences to the proposed techniques:} These approaches are either not best-first or not worst-case linear-space.
	
	\noindent\emph{Discussion:}
	FastDiag \cite{felfernig2012efficient} and its sequential diagnosis extension Inv-HS-Tree 
	\cite{Shchekotykhin2014} perform a linear-space depth-first diagnosis search that is grounded on the relationship between diagnoses and conflicts according to the Duality Property (cf.\ Sec.~\ref{sec:relationship_conflicts_diagnoses}). The soundness and completeness of the diagnosis computation despite the depth-first search is accomplished by interchanging the role of conflicts and diagnoses in 
	the hitting set tree. That is, in these approaches the node labels correspond to minimal diagnoses and the tree paths represent conflicts. The computation of minimal diagnoses instead of minimal conflicts during the labeling process is achieved by a suitable adaptation \cite{Shchekotykhin2014} of the QuickXplain algorithm \cite{junker04,rodler2020qx}. The main difference between these approaches and RBF-HS (and HBF-HS) is that the former cannot ensure that the diagnoses are computed in any particular (preference) order. 
	
	The authors of \cite{stern2012} present a sound and complete approach that interleaves conflict and diagnosis computation in a way that information from conflict computation aids the diagnosis computation and vice versa. However, unlike RBF-HS, this approach is not linear-space in general. In addition, it cannot compute most probable but only minimum-cardinality diagnoses.
	
	\subsubsection{Best-First-Search Approaches\protect\footnote{We restrict the discussion here to sound and complete methods. The consideration of all best-first algorithms would go beyond the scope of this work.}} 
	\noindent\emph{General differences to the proposed techniques:} 
	Whenever these approaches are sound and complete, they are worst-case exponential-space. 
	
	\noindent\emph{Discussion:}
	First and foremost, there are the seminal methods HS-Tree \cite{Reiter87}, along with its amended version HS-DAG proposed by \cite{greiner1989correction}, and GDE \cite{dekleer1987}, which are sound, complete\footnote{
	HS-Tree is complete only if no non-minimal conflicts are generated (cf.\ Example~\ref{ex:Reiter's_HS_Node_Processing}), like in our setting.
	} and best-first. 
	
	The works of \cite{Rodler2015phd,Kalyanpur2006a} describe sound and complete uniform-cost search variants of 
	HS-Tree which enumerate diagnoses in some order of preference. At this, \cite{Rodler2015phd} defines the preference order by means of a probability model over diagnoses (as characterized in Sec.~\ref{sec:probability_model}) whereas \cite{Kalyanpur2006a} relies on a heuristic model that ranks single axioms based on their ``importance''. The sum over axioms included in a diagnosis is used to determine the rank of the diagnosis. The author of \cite{meilicke2011alignment} goes one step further and incorporates a heuristic function (cf.\ Sec.~\ref{sec:path-finding_search_algos}) into the search, yielding a hitting set version of A*. 
	To the best of our knowledge, the specification of a useful heuristic function, as suggested in \cite{meilicke2011alignment} for an additive cost function, is an open problem in uniform-cost hitting set search with 
	multiplicative costs, 
	as in the case of our proposed methods.
	
	\cite{wotawa2001variant} suggests a variant of HS-DAG which builds a hitting set tree based on a subset-enumeration strategy in order improve the diagnosis computation time. The same objective is pursued by \cite{jannach2016parallel}, who propose parallelization techniques for 
	HS-Tree. 
	
	Further, there are sound, complete and best-first diagnosis searches that are particularly useful for fault isolation and sequential diagnosis \cite{dekleer1987}: 
	StaticHS \cite{rodler2018statichs} and DynamicHS \cite{rodler2020ecai}. These are stateful in that they exploit a persistently stored and incrementally adapted (search) data structure to make the diagnostic process more efficient. More specifically, DynamicHS targets the minimization of the computation time, whereas StaticHS aims at the reduction of the number of interactions necessary from a user, e.g., to make system measurements or answer system-generated queries.
	
	In contrast to RBF-HS and HBF-HS, all these best-first search approaches 
	are not ``memory-aware'' in that they
	require exponential space in general. 
	
	\subsection{Diagnosis Computation in the Knowledge-Base Debugging Domain}
	Due to their independence of the adopted system description language and theorem prover, RBF-HS and HBF-HS appear to be particularly attractive for application domains where different logics and reasoning engines are regularly used. One such field is knowledge-base debugging, which is also the focus of our evaluations. 
	
	Methods for knowledge-base debugging can be divided into \emph{model-based} and \emph{heu-ristics-based} approaches. The former, e.g., \cite{Rodler2015phd,Shchekotykhin2012,KalyanpurPSH05,SchlobachHCH07}, can be seen as principled, theorem-proving-based methods which draw on the general theory of model-based diagnosis \cite{Reiter87,dekleer1987}. 
	While these techniques usually allow for the finding of a precise and succinct explanation of all identified problems in a knowledge base, they can be comparably costly in terms of computation time and space due to the necessity of logical reasoning.
	Heuristics-based approaches, e.g., \cite{Rector2004,schulz2010pitfalls,Roussey2009,poveda2012validating},
	in contrast, can be regarded as experience-based techniques that draw on empirical knowledge such as common fault patterns, rules of thumb, or best practices. 
	They are a fast alternative 
	in case model-based techniques are too slow, but are often incomplete (i.e., they can only identify diagnoses revealing bugs for which appropriate heuristics were defined) and sometimes unsound (i.e., they might return diagnoses that comprise correct axioms). 
	
	Model-based approaches can be further classified into \emph{glass-box} and \emph{black-box} ones, based on the way how a logical reasoner is used in the debugging process (cf.\ Sec.~\ref{sec:classifying_diag_comp_methods}). 
%
	While glass-box approaches are highly optimized and performant for particular logics,
	black-box methods
	score with their 
	flexible applicability for a multitude of logics and reasoning engines \cite{Kalyanpur2006a}. 
	%
	Orthogonal to their categorization as black-box or glass-box, model-based techniques usually focus on one of two main ways of locating the faulty axioms in a knowledge base. The first class are \emph{justification-based approaches}, e.g.\ 
	\cite{Horridge2011a,Kalyanpur2006a}, which assume that users \emph{directly analyze conflicts}	to mentally reason about the faultiness of the axioms occurring in the conflicts. The second class of \emph{diagnosis-based approaches}, e.g., \cite{Shchekotykhin2012,SchlobachHCH07,kalyanpur2006repairing}, to which the methods presented in this work belong, takes the intermediate step of \emph{computing diagnoses} in order to assist users in this cognitively complex task. To this end, two main	ways of diagnosis computation were proposed for knowledge-based systems, either as hitting sets of conflicts using Reiter's HS-Tree \cite{Rodler2015phd,Horridge2011a,Kalyanpur2006a,SchlobachHCH07,meilicke2011}, or by means of duality-based techniques \cite{Shchekotykhin2014,rodler2020mbd_sampling} (cf.\ discussion in Sec.~\ref{sec:related:duality_based_approaches}). 

\section{Evaluation}
\label{sec:eval}

\subsection{Objective}
The goals of our evaluation are
\begin{itemize}[noitemsep,topsep=0pt]
	\item to demonstrate the out-of-the-box general applicability of 
	the proposed algorithms to diagnosis problems over
	different and highly 
	expressive
	knowledge representation languages,
	\item to understand their practical runtime, memory efficiency, and scalability, and
	\item to compare the suggested methods 
	against 
	one of the most widely used algorithms with the same properties in terms of the classification discussed in Sec.~\ref{sec:classifying_diag_comp_methods}. 
\end{itemize}
Importantly, the goal is \emph{not} to show that the proposed algorithms are better than all or most algorithms in literature, which is pointless as discussed in Sec.~\ref{sec:towards_improving_existing_methods}.
 Rather, we intend to show the advantage of using the proposed techniques 
in a diagnosis scenario where the properties soundness, completeness, the best-first enumeration of diagnoses, and the general applicability are of interest or even required. 

One such domain is ontology and knowledge base\footnote{
	We 
	use the terms \emph{ontology} and \emph{knowledge base} interchangeably 
	in this section. For the purposes of this paper, we consider both to be finite sets of axioms expressed in some monotonic logic, cf.\ $\mo$ in Sec.~\ref{sec:basics}.} debugging, where practitioners and experts from the field usually\footnote{
	The discussed requirements emerged from discussions with ontologists whom we interviewed (e.g., in the course
	of a tool demo at the Int’l Conference on Biological Ontology 2018) in order to develop and customize our ontology debugging tool \emph{OntoDebug} \cite{schekotihin2018protege,DBLP:conf/foiks/SchekotihinRS18} to match its users' needs.}, and especially in critical applications of ontologies such as medicine \cite{rector2011getting}, want a debugger to output 
exactly the faulty axioms that really explain the observed faults in the ontology (\emph{soundness} and \emph{completeness}) at the end of a debugging session.
In addition, experts often wish to perpetually monitor the most promising fault explanations throughout the debugging process (\emph{best-first property}) with the intention to stop the session early if they recognize the fault. As was recently studied by \cite{rodler2020mbd_sampling}, the use of best-first algorithms often also involves efficiency gains in debugging as opposed to other strategies. Apart from that, it is a big advantage for users of knowledge-based systems to have a debugging solution that works out of the box for different logical languages and with different logical reasoners (\emph{general applicability}, cf.\ \emph{black-box} property in Sec.~\ref{sec:classifying_diag_comp_methods}). The reasons for this are that \emph{(i)}~ontologies are formulated in a myriad of different (Description) logics \cite{DLHandbook} with the aim to achieve the required expressivity for each ontology domain of interest at the least cost for inference, and \emph{(ii)}~highly specialized reasoners exist for different logics (cf., e.g., \cite{baader2005tractable,kazakov2014}), and being able to flexibly switch to the most efficient reasoner for a particular debugging problem 
can bring significant performance improvements \cite{romero2012more}.

For these reasons, we use
\begin{itemize}[noitemsep,topsep=0pt]
	\item 
	real-world knowledge base debugging problems (cf.\ Sec.~\ref{sec:dataset}) formulated over a range of different logics with hard reasoning complexities
	to test our approaches, and
	\item HS-Tree \cite{Reiter87,Rodler2015phd} (cf.\ Example~\ref{ex:Reiter's_HS_Node_Processing}) to compare our approaches against,  which \emph{(i)}~has exactly the same properties---soundness, completeness, best-firstness, and general applicability---as 
	our proposed techniques, and \emph{(ii)}~appears to be the most prevalent and widely used algorithm (e.g., \cite{Rodler2015phd,meilicke2011alignment,Shchekotykhin2012,Horridge2011a,Kalyanpur2006a,Rodler2013,DBLP:conf/foiks/SchekotihinRS18,fu2016graph,baader2018weakening,ji2009radon,ribeiro2009base,moodley2011root,schlobach2007debugging}) for diagnosis computation in the area of knowledge-based systems. 
\end{itemize}

\setlength{\tabcolsep}{24pt}
\begin{table}
	\renewcommand\arraystretch{1}
	\scriptsize
	\centering
	\caption{\small Dataset used in the experiments (sorted by the number of components/axioms of the diagnosis problem, 2nd column).}
	\label{tab:dataset}
	\begin{minipage}{0.98\linewidth}
		\centering
		\begin{tabular}{@{}lrlr@{}} 
			\toprule
			KB $\mo$				& $|\mo|$& expressivity \textsuperscript{\textbf{1)}} 		& \#D/min/max \textsuperscript{\textbf{2)}} \\ \midrule
			Koala (K) 
			& 42 		& $\mathcal{ALCON}^{(D)}$& 10/1/3     \\
			University (U) 
			& 50 		& $\mathcal{SOIN}^{(D)}$& 90/3/4      \\
			IT  
			& 
			140 		& $\mathcal{SROIQ}$& 1045/3/7	  \\
			UNI  
			\phantom{\textsuperscript{\textbf{4)}}}		
			& 
			142 		& $\mathcal{SROIQ}$& 1296/5/6	  \\
			Chemical (Ch) 
			& 144 		& $\mathcal{ALCHF}^{(D)}$& 6/1/3     \\
			MiniTambis (M) 
			\phantom{\textsuperscript{\textbf{4)}}}		
			& 173 		& $\mathcal{ALCN}$ 		& 48/3/3	  \\
			
			ctxmatch-cmt-conftool (ccc) 
			& 
			458 		& $\mathcal{SIN}^{(D)}$&  934*/2/16* \\
			ctxmatch-conftool-ekaw (cce) 
			& 
			491 		& $\mathcal{SHIN}^{(D)}$&  953*/3/35* \\
			Transportation (T) 
			& 
			1300 		& $\mathcal{ALCH}^{(D)}$& 1782/6/9	  \\
			Economy (E) 
			& 1781 		& $\mathcal{ALCH}^{(D)}$& 864/4/8     \\
			DBpedia (D) 
			& 
			7228 		& $\mathcal{ALCHF}^{(D)}$& 7/1/1     \\
			Opengalen (O) 
			& 
			9664		& $\mathcal{ALEHIF}^{(D)}$& 110/2/6     \\
			CigaretteSmokeExposure (Cig) 
			& 
			26548 		& $\mathcal{SI}^{(D)}$& 1566*/4/7*	  \\
			Cton (C) 
			& 
			33203		& $\mathcal{SHF}$& 15/1/5     \\
			\bottomrule
		\end{tabular}
	\end{minipage}
	\renewcommand\arraystretch{1}
	\begin{minipage}{0.98\linewidth}
		\centering
		\setlength{\tabcolsep}{2pt}
		\begin{tabular}{@{}lp{11.4cm}@{}}
			\textbf{1):} & Description Logic expressivity: each calligraphic letter stands for a (set of) logical constructs that are allowed in the respective language, e.g., $\mathcal{C}$ denotes negation (``complement'') of concepts, for details see \cite{DLHandbook,DL_complexity};
			intuitively, the more letters, the higher the expressivity of a logic and the complexity of reasoning for this logic tends to be.
			\\
			\textbf{2):} & \#D/min/max denotes the number / the minimal size / the maximal size of minimal diagnoses for the DPI resulting from each input KB $\mo$. If tagged with a $^*$, a value signifies the number or size determined within one hour using the suite of algorithms included in the \emph{OntoDebug} tool \cite{DBLP:conf/foiks/SchekotihinRS18}	 (for problems where the finding of \emph{all} minimal diagnoses was impossible within reasonable time). 
		\end{tabular}
	\end{minipage}
\end{table}

\subsection{Dataset}
\label{sec:dataset}
The benchmark of 
inconsistent or incoherent\footnote{ A knowledge base $\mo$ is called \emph{incoherent} iff it entails that some predicate $p$ must always be false; formally: $\mo \models \forall \mathbf{X} \lnot p(\mathbf{X})$ where $p$ is a predicate with arity $k$ and $\mathbf{X}$ a tuple of $k$ variables. With regard to ontologies, incoherence means that some class (unary predicate) must not have any instances, and if so, the ontology becomes inconsistent (cf., e.g., \cite{SchlobachHCH07}).} real-world ontologies we used for our experiments 
is given in Tab.~\ref{tab:dataset}.\footnote{The benchmark problems can be downloaded from \url{http://isbi.aau.at/ontodebug/evaluation}.}
Subsets of this dataset have been investigated i.a.\ in \cite{Shchekotykhin2012,Kalyanpur2006a,rodler2019KBS_userstudy,qi2007measuring,del2010modular,Stuckenschmidt2008,Horridge2008}. 
As the table shows, the ontologies cover a spectrum of different problem sizes (number of axioms or components; column~2), logical expressivities (which determine the complexity of consistency checking; column~3),  
as well as diagnostic structures (number and size of minimal diagnoses; column~4). Note that the complexity of consistency checks (used within the \textsc{findMinConflict} procedure in Alg.~\ref{algo:RBF_HS})
over the logics in Tab.~\ref{tab:dataset} ranges from EXPTIME-complete to 2-NEXPTIME-complete \cite{DL_complexity,grau2008owl}.
Hence, from the point of view of model-based diagnosis, ontology debugging problems represent a particularly challenging domain as they usually deal with harder logics
than more traditional diagnosis problems (which often use propositional knowledge representation languages that are not beyond {NP-complete}). 

\subsection{Experiment Settings}
\label{sec:experiment_settings}

\subsubsection{Different Diagnosis Scenarios}
\label{sec:eval:different_diagnosis_scenarios}
To study the performance and robustness of our approaches under varying circumstances, we considered a range of different \emph{diagnosis scenarios} in our experiments. 
A diagnosis scenario is defined by the set of inputs given to Alg.~\ref{algo:RBF_HS}, i.e., by a DPI $\dpi$, a number $\ld$ of minimal diagnoses to be computed, as well as a setting of the component fault probabilities $\pr$.
The DPIs for our tests were defined as $\tuple{\mo,\emptyset,\emptyset,\emptyset}$, one for each $\mo$ in Tab.~\ref{tab:dataset}. That is, the task was to find a minimal set of axioms (faulty components) responsible for the inconsistency or incoherence of $\mo$, without any background knowledge or measurements initially given (cf.\ Example~\ref{ex:RBF-HS}). For the parameter $\ld$ we used the values $\{2,6,10,20\}$.
The fault probability $\pr(\tax)$ of each axiom (component) $\tax \in \mo$ was either chosen uniformly at random from $(0,1)$ (\emph{maxProb}), or specified in a way (cf.\ Remark~\ref{rem:breadth-first_can_be_simluated_by_uniform-cost} in Example~\ref{ex:Reiter's_HS_Node_Processing}) the diagnosis search returns minimum-cardinality diagnoses first (\emph{minCard}). As a Description Logic reasoner, we adopted Pellet \cite{Sirin2007}.

\subsubsection{Goal to Find Actual Diagnosis}
\label{sec:eval:goal_to_find_actual_diag}
To simulate as realistic as possible diagnosis circumstances, where the actual diagnosis (i.e., the de-facto faulty axioms) is of interest and needs to be isolated from a set
of initial minimal diagnoses (cf.\ column 4 of Tab.~\ref{tab:dataset}), we ran five sequential diagnosis
\cite{dekleer1987,Shchekotykhin2012} sessions 
executed by each RBF-HS and HS-Tree, for each diagnosis scenario defined in Sec.~\ref{sec:eval:different_diagnosis_scenarios}. 
Roughly, the idea is to use 
a diagnosis search 
as a routine called multiple times during an iterative information acquisition process
with the goal to find the actual fault \emph{with certainty}.
Specifically, a \emph{sequential diagnosis session} 
can be conceived of as having the following two alternating phases that are iterated until a single minimal diagnosis remains: 
\begin{itemize}[noitemsep,topsep=0pt]
	\item diagnosis search, and
	\item measurement conduction.
\end{itemize}
The former involves the determination of the $\ld$ most preferred minimal diagnoses $\mD$ according to $\pr$ for a given DPI. The latter subsumes the selection of an optimal system measurement 
based on $\mD$ (to rule out as many spurious diagnoses 
as possible), as well as the incorporation of the new system knowledge resulting from the measurement outcome into the DPI.

Measurement selection requires a \emph{measurement selection procedure} \cite{DBLP:journals/corr/Rodler16a,rodler17dx_activelearning,rodler2018ruleML} which gets a set of minimal diagnoses $\mD$ as input, and outputs one system measurement such that 
\emph{(i)}~this measurement is optimal wrt.\ some \emph{measurement quality criterion}, and
\emph{(ii)}~any outcome for this measurement eliminates at least one spurious diagnosis in $\mD$. As measurement quality criteria we adopted \emph{split-in-half (SPL)} \cite{Shchekotykhin2012}, which suggests 
a measurement with the highest worst-case number of spurious diagnoses in $\mD$ eliminated,
and \emph{entropy (ENT)} \cite{dekleer1987}, which selects a measurement with highest information gain. These two quality criteria appear to be the commonly adopted ones in model-based diagnosis, cf., e.g., \cite{Shchekotykhin2014,Shchekotykhin2012,Rodler2013,rodler2018ruleML,pietersma2005model,pattipati1990,feldman2009fractal,ressencourt2006hierarchical,gonzalez2011spectrum}. 
%

In our experiments, a measurement was defined as a true-false question to an oracle \cite{Rodler2015phd,Shchekotykhin2012,rodler2019KBS_userstudy,rodler_eichholzer2019},
e.g., for a biological knowledge base one such query could be $Q:=\mathsf{Bird} \sqsubseteq \exists \mathsf{hasCapability}.\mathsf{Flying}$ (``is every bird capable of flying?''). 
As a measurement selection procedure for computing optimal queries wrt.\ the criteria SPL and ENT we adopted the algorithm suggested in \cite{rodler17dx_queries,DBLP:journals/corr/Rodler2017}.
Given a positive (negative) answer for a query $Q$, the DPI is updated by assigning $Q$ to the positive (negative) measurements (cf.\ Sec.~\ref{sec:diagnosis_problem}). 
The resulting new DPI is then used in the next iteration of the sequential diagnosis session. That is, a new set $\mD$ of the $\ld$ most preferred diagnoses 
is determined for this updated DPI, an optimal measurement is calculated for $\mD$, and so on. Once there is only a single minimal diagnosis for a current DPI, the session stops and outputs the remaining diagnosis. 

To solve a different problem in each of the five executed sequential diagnosis sessions per diagnosis scenario, 
we predefined a different randomly chosen actual diagnosis as the target solution per session.\footnote{As no correct solutions are known for the knowledge-based benchmark problems in Tab.~\ref{tab:dataset}, any diagnosis might in principle be the right solution. Which diagnosis is the actual one depends on the desiderata of the stakeholders of a knowledge base and might differ depending on the particular modeled domain.  
E.g., when a knowledge base is about an academic domain, a professor might need to hold at least two courses per semester if some university X is modeled, while professors with less than two courses might legitimately exist if another university Y is described.
This motivates the random selection of the target solution in our experiments, which is also common practice in the field (cf., e.g., \cite{Shchekotykhin2014,Shchekotykhin2012,Rodler2013}).}
This preset actual diagnosis was also used to determine measurement outcomes (i.e., to answer the generated questions) in that each question was automatically answered in a way the actual diagnosis was not ruled out.



Advantages of concentrating on sequential diagnosis runs in our evaluations (instead of single-run diagnosis search executions\footnote{\ref{apx:eval_results_single-run-case} provides results for single-run tests we have executed in addition to the presented sequential diagnosis experiments. The results for both scenarios are highly consistent.}) are:
\begin{itemize}[noitemsep,topsep=0pt]
	\item Sequential diagnosis is one of the main applications of diagnosis searches.
	\item The potential impact (cf.\ \cite{rodler2020mbd_sampling}) of different measurement quality criteria on algorithms' performances can be assessed.
	\item Without the information acquisition through sequential diagnosis it is in many cases practically infeasible to find the actual fault (cf.\ the large numbers of diagnoses in the fourth column of Tab.~\ref{tab:dataset}).
	\item Multiple diagnosis searches, each for a different (updated) DPI, are executed during one sequential session and flow into the experiment results, which provides more representative evidence of the algorithms' robustness and real performance.
\end{itemize}

\subsubsection{Settings in a Nutshell}
%
We ran five sequential diagnosis sessions, each searching for a randomly specified minimal diagnosis, for each algorithm among RBF-HS
and HS-Tree, for each measurement quality criterion among ENT and SPL, for each DPI from Tab.~\ref{tab:dataset}, for each probability setting among maxProb and minCard, and for each number of diagnoses $\ld \in \{2,6,10,20\}$ to be computed (in each iteration of the session, i.e., at each call of a diagnosis search algorithm). 

\subsubsection{Scalability Tests}
\label{sec:scalabiity_tests}
In order to evaluate the scalability of the tested algorithms, 
we conducted an additional scalability experiment. To this end, we 
ran tests with $\ld := 100$ and otherwise same settings as described above on all DPIs from our dataset for which at least 100 minimal diagnoses exist (cf.\ last column of Tab.~\ref{tab:dataset}). 
Note, the parameter $\ld$ has a major influence on the hardness of the diagnosis computation task, since, given a set of minimal diagnoses, only deciding whether an additional minimal diagnosis not in this set exists is already NP-complete (even if logical consistency checking is in P) \cite{Bylander1991}.

\subsection{Experiment Results\footnotemark}
\label{sec:experiment_results}
We first explain how to read the figures, and then discuss the experiment results. 
\subsubsection{Presentation of the Results}
\label{sec:presentation_of_results}
The results for the minCard experiments are shown by 
Fig.~\ref{fig:results_card_SPL} (measurement quality criterion SPL), Fig.~\ref{fig:results_card_ENT} (measurement quality criterion ENT), Fig.~\ref{fig:results_card_scalability} (scalability tests), and Fig.~\ref{fig:results_card_hardest_cases} (analysis of the hardest cases).
Each figure compares the runtime and memory consumption we measured for RBF-HS and HS-Tree averaged over the five performed sessions (note the logarithmic scale). 
More specifically, the figures depict the factor of less memory consumed by RBF-HS (blue bars), as well as the factor of more time needed by RBF-HS (orange bars), in relation to HS-Tree.
That is, blue bars tending upwards (downwards) mean a better (worse) memory behavior of RBF-HS, whereas upwards (downwards) orange bars signify worse (better) runtime of RBF-HS. For instance, a blue bar of height 10 means that HS-Tree required 10 times as much memory as RBF-HS did in the same experiment; or a downwards orange bar representing the value 0.5 indicates that RBF-HS finished the diagnosis search task in half of HS-Tree's runtime. 
Regarding the absolute runtime and memory expenditure (not displayed in the figures),
we measured a min / avg / max runtime of 
0.02 / 25 / 2176 sec (ENT) and 0.03 / 29 / 4806 sec (SPL), as well as
a min / avg / max space consumption of
7 / 10K / 2M tree nodes (ENT) and 7 / 7K / 1.2M tree nodes (SPL). 
Fig.~\ref{fig:results_HBF-HS_card} illustrates the impact of using HBF-HS in those cases where we observed significant ($\geq$ 20\,\%) time overheads of RBF-HS (orange bars) versus HS-Tree (red horizontal line). The runtimes of HBF-HS as compared against HS-Tree are indicated by the blue bars, which are shaded in those cases where HBF-HS exhibited an even lower runtime than HS-Tree. Moreover, the gray circles along the black line plot the percental time savings achieved by HBF-HS compared to RBF-HS. 

\footnotetext{Please see \url{http://isbi.aau.at/ontodebug/evaluation} for the raw data. Our implementations of RBF-HS and HBF-HS can be accessed at \url{https://bit.ly/2Gp3XwX}.}


\subsubsection{Discussion of the Results}

We next summarize the main insights from our experiment results. 
As a preliminary point, please note that all the algorithms under test have to compute the same predefined target solution in each test run, and that all of them have the same features soundness, completeness and best-firstness. Hence, observed savings of one algorithm versus another \emph{neither} arise at the cost of losing any theoretical guarantees, \emph{nor} due to computing a different output.
\begin{enumerate}[itemsep=2pt,label=\emph{(\arabic*)},align=left,leftmargin=0pt,labelwidth=-0.5\parindent]  
	\item \emph{Favorable space-time tradeoff:}
	Whenever the diagnosis problem was non-trivial to solve, i.e., required a runtime of more than one second (which was the case in 94\,\% of the tested cases), RBF-HS traded space favorably for time. In other words, compared to HS-Tree, the factor of memory saved by RBF-HS was higher than the factor of incurred time overhead in all interesting cases (blue bar is higher than orange one; cf.\ Figs.~\ref{fig:results_card_SPL} and \ref{fig:results_card_ENT}). 
	%
	%

	\begin{figure}[t]
		\includegraphics[width=0.985\columnwidth]{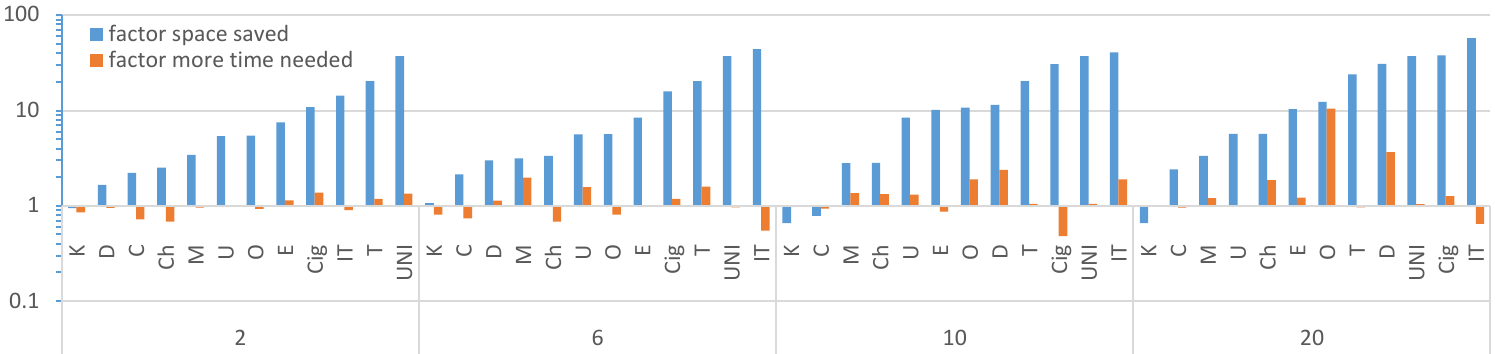}
		\caption{Results (RBF-HS vs.\ HS-Tree) for the measurement quality criterion SPL: x-axis shows the ontologies from Tab.~\ref{tab:dataset} and the number of computed diagnoses $\ld \in \{2,6,10,20\}$. Per setting of $\ld$, the ontologies along the x-axis are sorted from low to high space savings achieved by RBF-HS (blue bars).}
		\label{fig:results_card_SPL}
	\end{figure}
	
	\begin{figure}[t]
		\includegraphics[width=\columnwidth]{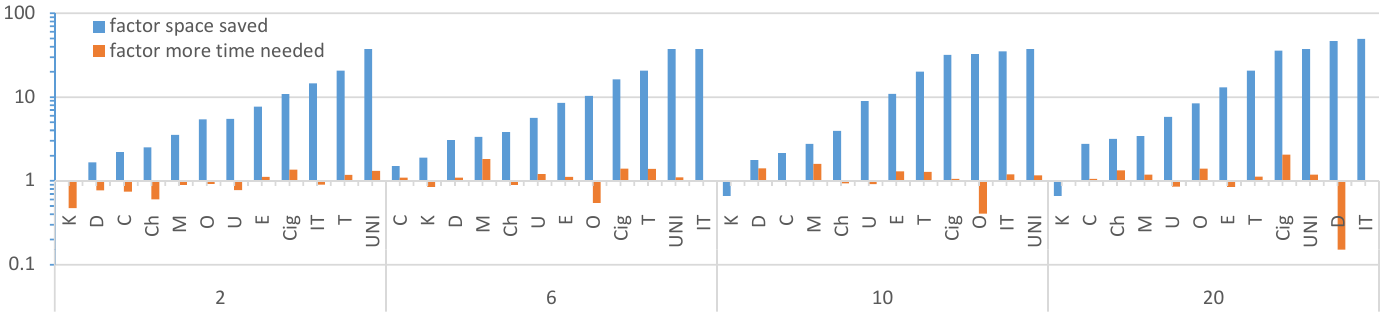}
		\caption{Results (RBF-HS vs.\ HS-Tree) for the measurement quality criterion ENT: x-axis shows the ontologies from Tab.~\ref{tab:dataset} and the number of computed diagnoses $\ld \in \{2,6,10,20\}$. Per setting of $\ld$, the ontologies along the x-axis are sorted from low to high space savings achieved by RBF-HS (blue bars). }
		\label{fig:results_card_ENT}
	\end{figure}

	\item \emph{Substantial space savings:} Space savings of RBF-HS ranged from significant to tremendous (cf.\ Figs.~\ref{fig:results_card_SPL} and \ref{fig:results_card_ENT}), and often reached factors larger than 10 (in 45\,\% of the cases) and up to 50 (ENT) and 57 (SPL). In other words, HS-Tree required up to 57 times as much memory for the same tasks as RBF-HS did. On average, the factor of memory saved amounted to 14.1 for ENT and to 13.8 for SPL, i.e., RBF-HS required an average of less than 8\,\% of the memory HS-Tree consumed. Note, in five scenarios (involving the ontologies K and C), HS-Tree required slightly less memory than RBF-HS, which however does not carry weight due to the fact that hitting set trees were very small (diagnoses of cardinality one, cf.\ Tab.~\ref{tab:dataset}) in these runs.
	\item \emph{Often even runtime improvements:} 
	In 35\,\% (ENT) and 38\,\% (SPL) of the cases, RBF-HS exhibited \emph{both} a lower 
	runtime compared with HS-Tree \emph{and} saved significant portions of memory (blue bar goes up, orange one goes down; cf.\ Figs.~\ref{fig:results_card_SPL} and \ref{fig:results_card_ENT}).
	This observation may appear surprising at first sight, since RBF-HS relies on forgetting and re-exploring, whereas HS-Tree keeps all relevant information in memory. However, also studies comparing classic (non-hitting-set) best-first searches have observed that linear-space approaches can outperform exponential-space ones in terms of runtime \cite{zhang1995performance}. One reason for this is that, at the processing of each node, the management (node insertion and removal) of an exponential-sized priority queue of open nodes requires time linear in the current tree depth. Hence, when the queue management time of HS-Tree outweighs the time for redundant node regenerations expended by RBF-HS, then the latter will outperform the former. 
	\item \label{enum:results:if_rbfhs_bad_use_hbfhs} \emph{Whenever it takes RBF-HS long, use HBF-HS:} 
	In those cases where RBF-HS manifested a significant (20\,\% or higher) time overhead versus HS-Tree, 
	the use of HBF-HS (with a mere allowance of 400 nodes in memory before the switch from HS-Tree to RBF-HS is triggered) could almost always reduce the runtime to times comparable with those of HS-Tree (cf.\ Fig.~\ref{fig:results_HBF-HS_card}). The median time overhead of HBF-HS compared to HS-Tree was some 0\,\% for ENT and 3\,\% for SPL, where HBF-HS led to even better times than HS-Tree in 44\,\% (ENT) and 42\,\% (SPL) of the cases.
	At the same time, 
	remarkably, 
	the memory consumption of HBF-HS \emph{never} exceeded 416 nodes, whereas HS-Tree required memory for up to more than half a million nodes, which amounts to a deterioration factor of over 1000 compared to HBF-HS. 	
	The time savings over RBF-HS achieved by HBF-HS were substantial in many cases, reaching up to 70\,\% for ENT and maxima of over 90\,\% for SPL (see the gray circles in Fig.~\ref{fig:results_HBF-HS_card}). For instance, in the scenario SPL, 20 for ontology O, the runtime overhead factor of 10.4 versus HS-Tree (the worst value for RBF-HS measured in all experiments, cf.\ Fig.~\ref{fig:results_card_SPL}), could be reduced to a factor of 0.98 by means of HBF-HS. 
	That is, the use of HBF-HS transformed a 940\,\% time overhead of RBF-HS to an even 2\,\% faster runtime than HS-Tree's.  
	
	This suggests that, whenever RBF-HS gets caught in redundant re-explorations of subtrees and thus requires notably more time than HS-Tree, the allowance of a relatively short run of HS-Tree (until it has generated 400 nodes) before switching to RBF-HS can already yield a runtime comparable to HS-Tree. One reason for this phenomenon is that RBF-HS can save a significant number of re-explorations through the information gained by the initial breadth-first exploration of the top of the search tree. A potential second reason might be the above-mentioned high expense of managing an increasingly large queue of open nodes required by HS-Tree, as opposed to a set of open nodes of smaller and almost fixed size in case of HBF-HS. 
	%
	
	\begin{figure}[t]
		\centering
		\includegraphics[width=\columnwidth]{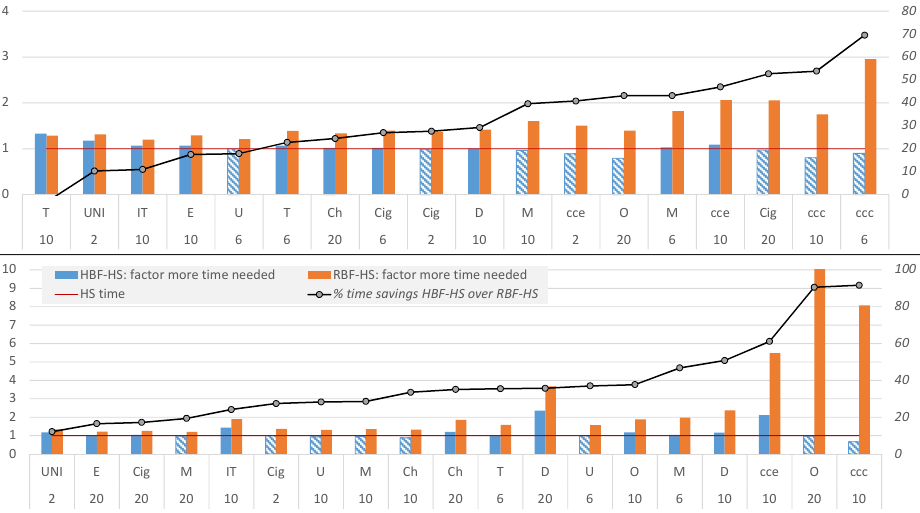}
		\caption{Results (HBF-HS with switch criterion ``400 nodes in memory'' vs.\ RBF-HS) for the measurement quality criteria ENT (\emph{top chart}) and SPL (\emph{bottom chart}): x-axis shows ontologies from Tab.~\ref{tab:dataset} and the parameter $\ld \in \{2,6,10,20\}$. Shaded blue bars highlight cases where the runtime of HBF-HS was even better than that of HS-Tree. The value of the gray circle for T, 10 that is not shown (for clarity) in the top chart, amounts to -3\,\%. Features plotted wrt. the left / right y-axis are written in normal / italic font.}
		\label{fig:results_HBF-HS_card}
	\end{figure}
	
	\item \label{enum:results:HBFHS_caps_memory}\emph{HBF-HS allows to almost ``cap'' the used memory:} The number of nodes in memory additionally consumed by HBF-HS after the switch (at 400 nodes) to RBF-HS was less than 2\,\% on average, and never more than 4\,\%, compared to the number of nodes in memory at the time the switch was executed. 
	Similar and only slightly higher values could be observed for HBF-HS performing the switch at 200 (3\,\% exceedence on average) and 100 (7\,\%) generated nodes.
	This suggests that the consumed amount of memory can practically be more or less arbitrarily limited by the definition of a suitable switch condition (which sets a memory limit that is not much lower than the---very low---memory requirement of standalone RBF-HS). 
	These findings are theoretically supported by Theorem~\ref{thm:hbfhs:space_complexity_after_switch}.
	\item \emph{Performance independent of number of computed diagnoses and measurement quality criteria:} The relative performance of RBF-HS versus HS-Tree appears to be largely independent of the number $\ld$ of computed minimal diagnoses as well as of the used measurement quality criterion (cf.\ Figs.~\ref{fig:results_card_SPL} and \ref{fig:results_card_ENT}).
	\item \emph{Performance improves for harder diagnosis problems:} 
	The gain of using RBF-HS instead of HS-Tree gets the larger, the harder the considered diagnosis problem is. This tendency can be clearly seen in Figs.~\ref{fig:results_card_SPL} and \ref{fig:results_card_ENT}, where the ontologies on the x-axis are sorted in ascending order of RBF-HS's memory reduction achieved, for each value of $\ld$. Note that roughly the same group of (more difficult / easy to solve) diagnosis problems ranks high / low for all values of $\ld$.
	\item \label{enum:results:performance_depends_on_diagnosis_pref_criterion}\emph{Performance dependent on diagnosis preference criterion:} The discussion of the results so far concentrated on the consistently good results attained by RBF-HS for the minCard 
	setting. In case of the maxProb setting, we see a pretty different picture, where time was more or less traded one-to-one for space, i.e., $k$ orders of magnitude savings in space against HS-Tree required approximately $k$ orders of magnitude more runtime of RBF-HS (blue and orange bars roughly equal). The reason for this performance degradation in case of maxProb is a known property of Korf's RBFS algorithm to perform relatively poorly when original $f$-values (in our case: probabilities) of nodes vary only slightly \cite{hatem2015recursive} 
	(cf.\ \ref{sec:time_complexity}). 
	As a result, 
	RBF-HS suffers from too many ``mind shifts'' and spends most of the time doing backtracking and re-exploration steps while making very little progress in the search tree. 
	
	However, as in the case of minCard, when we allowed for the utilization of a small amount of more memory than RBF-HS used, this problem was remedied to a great extent. In fact, adopting HBF-HS with a switch at 400 generated nodes mostly led to a runtime comparable to HS-Tree's, and in 43\,\% of the scenarios to an even lower one. 
	Only in a single scenario, i.e., O, SPL, 20, 
	HBF-HS (with a switch at 400 nodes) still required substantially more time than HS-Tree did. Obviously, this exact combination represented a particularly demanding case for HBF-HS and RBF-HS (cf.\ Bullet \ref{enum:results:if_rbfhs_bad_use_hbfhs}). 
	
	As additional tests turned out, the answer to this problem is the employment of HBF-HS equipped with a \emph{relative switch criterion} (instead of an absolute one). Concretely, we allowed HS-Tree to consume 60\,\% of the available memory before handing over to RBF-HS. 
	Runtimes as for HS-Tree could be achieved in this way (while making use of only marginally more than 60\,\% 
	of the disposable memory, cf.\ Bullet \ref{enum:results:HBFHS_caps_memory}).
	\begin{figure}[t]%
	\centering
	\includegraphics[width=0.9\textwidth]{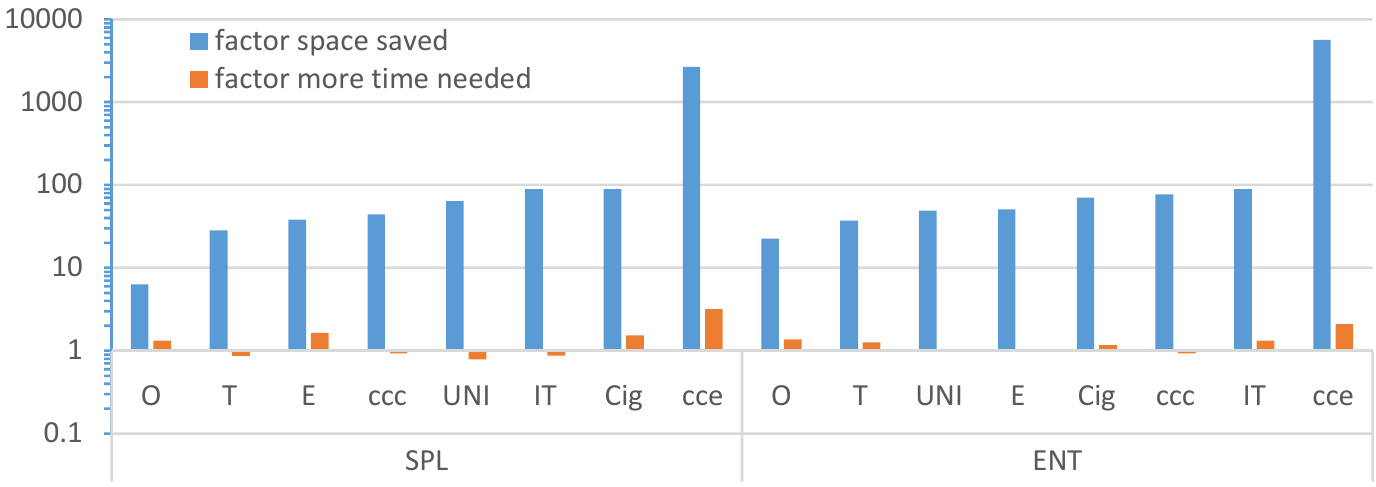}
	\caption{
		Scalability results (RBF-HS vs.\ HS-Tree) for $\ld = 100$: 
		x-axis shows ontologies from Tab.~\ref{tab:dataset} and indicates the used measurement quality criterion (SPL or ENT).
		Per measurement quality criterion, the ontologies are sorted from low to high space savings achieved by RBF-HS (blue bars).} 
	\label{fig:results_card_scalability}%
\end{figure}

	\begin{figure}[t]%
	\centering
	\includegraphics[width=0.95\textwidth]{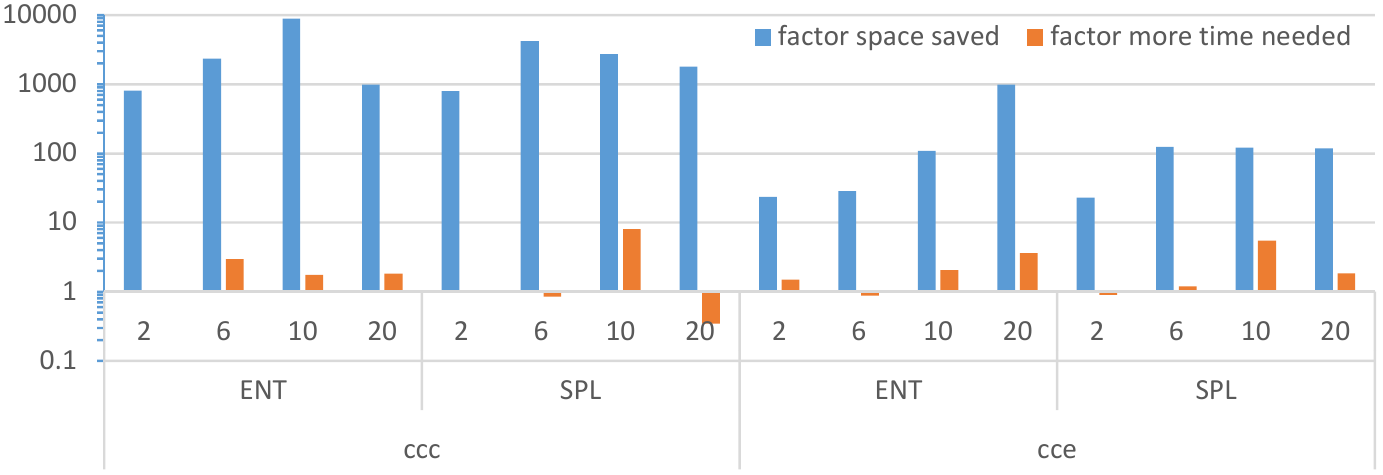}
	\caption{Results (RBF-HS vs.\ HS-Tree) for the two hardest cases ccc, cce from Tab.~\ref{tab:dataset}: 
	x-axis shows the used measurement quality criterion (SPL or ENT) and the number of computed diagnoses $\ld \in \{2,6,10,20\}$.} 
	\label{fig:results_card_hardest_cases}%
\end{figure}

	%
	%
	\item \emph{Scalability tests:} 
	The observations discussed so far have brought to light that DPIs with thousands of components (axioms) and diagnoses (cf.\ columns 2 and 4 in Tab.~\ref{tab:dataset}) could be well handled by RBF-HS in our tests (Figs.~\ref{fig:results_card_SPL} and \ref{fig:results_card_ENT}), and even led to a better relative performance in comparison to HS-Tree than problems with fewer components and possible faults. In our scalability experiments (cf.\ Sec.~\ref{sec:scalabiity_tests}), we additionally tested the algorithm performance when a large number of diagnoses is computed. The results for the minCard setting are presented in Fig.~\ref{fig:results_card_scalability}. 
%
	It displays that enormous space savings (in all cases) oppose
	\begin{itemize}[noitemsep,topsep=0pt]
	\item reasonable runtime overheads (9 cases), which were always lower than a factor of 1.65 except for the ontology cce with factors 3.18 (SPL) and 2.09 (ENT),
	\item roundly equal runtimes (2 cases), and
	\item even runtime savings (5 cases) ranging between 7\,\% and 22\,\%.
	\end{itemize}
	Space savings achieved by RBF-HS ranged from 84\,\% (case O, SPL) to 
	more than 99.9\,\% (case cce, SPL+ENT) and exceeded 95\,\% in all but a single case. Even the combination of the measurement quality criterion SPL and ontology O, which proved to be a particularly unfavorable case as regards runtime in the normal experiments (cf.\ Bullets~\ref{enum:results:if_rbfhs_bad_use_hbfhs} and \ref{enum:results:performance_depends_on_diagnosis_pref_criterion}), turned out to be unproblematic in the scalability tests. This shows that RBF-HS scales very well when minimum-cardinality diagnoses are of interest.
	
	Note, for three of the five runs for the case cce with measurement quality criterion ENT, we observed that HS-Tree ran out of memory (after a runtime of three minutes or less) whereas RBF-HS could successfully solve these problems in an average time of two and a half minutes while requiring only a negligible amount of memory for no more than 102 tree nodes.
	
	For the maxProb setting, the insight was that RBF-HS, in general, does not scale to large numbers of computed diagnoses like $\ld = 100$, as it required up to several hours of computation time per executed sequential session. HS-Tree as well as HBF-HS (with a relative switch criterion of 60\,\% consumption of the available memory, cf.\ Bullet~\ref{enum:results:performance_depends_on_diagnosis_pref_criterion}), on the other hand, could finish the same tasks in the range of few minutes. The conclusion is that, for the computation of most probable diagnoses, HBF-HS with a relative switch criterion should be used rather than RBF-HS.    
	
	\item \emph{Results for the hardest cases:} For the purpose of clarity of Figs.~\ref{fig:results_card_SPL} and \ref{fig:results_card_ENT}, we excluded the results for the two DPIs ccc and cce.
	These two DPIs result from the integration (\emph{alignment} \cite{euzenat2011ontology}) of two ontologies describing a common domain (in this case: a conference management system) in a different way. As a consequence of the automatized alignment process, a multitude of independent
	issues in terms of (minimal) conflicts
	emerge at once in the resulting ontology. This leads to large sizes of minimal diagnoses (cf.\ Tab.~\ref{tab:dataset}, column 4), which causes a high depth and thus enormous size of the hitting set tree. The runtime and memory measurements for these hard cases are demonstrated by Fig.~\ref{fig:results_card_hardest_cases}.
	We detect gigantic space savings up to nearly four orders of magnitude achieved by RBF-HS while runtime still remained in most cases comparable with HS-Tree; in 38\,\% of the cases RBF-HS's runtime was even better. For instance, for the case ccc, ENT, 2, we observed that HS-Tree required more than 800 times the memory used by RBF-HS, while RBF-HS exhibited also a 3\,\% lower runtime. Even more impressingly, RBF-HS reduced the memory consumption by a factor of more than 4200 while at the same time decreasing the computation time by 15\,\% in the case ccc, SPL, 6. Moreover, in the case ccc, SPL, 20, we registered substantial time savings of 65\,\% coupled with a 99.9\,\% memory reduction achieved by RBF-HS. 
	Finally, note that in one run for the cce, SPL, 20 setting, HS-Tree ran out of memory after running 37 minutes while RBF-HS solved the same problem in less than 11 minutes and required memory for merely 125 tree nodes.  
	Again, as discussed above,
	the use of HBF-HS allows to level any significant time overheads of RBF-HS while consuming a limited amount of memory.
	
\end{enumerate}  

\section{Conclusions and Future Work}
\label{sec:conclusion}


In this work, we introduced two new diagnostic search techniques, RBF-HS and HBF-HS, which borrow ideas from Korf's seminal RBFS algorithm \cite{korf1993linear}. The unique characteristic of RBF-HS is that it requires only linear space for the computation of an arbitrary fixed finite number of minimal diagnoses (fault explanations) while preserving the desired features soundness (\emph{only} actual fault explanations are computed), completeness (\emph{all} fault explanations can be computed), and the best-first property (fault explanations are computed \emph{in order} based on a given preference criterion).
HBF-HS is a hybrid strategy that aims at leveraging synergies between Reiter's HS-Tree \cite{Reiter87} and RBF-HS in a way that problems can be solved in reasonable time without depleting the required memory. Both suggested algorithms are generally applicable to any diagnosis problem according to Reiter's theory of model-based diagnosis \cite{Reiter87}; in particular, they are independent of the (monotonic) knowledge representation language used to describe the diagnosed system and of the adopted inference engine.



In comprehensive experiments on a corpus of real-world knowledge-based diagnosis problems of various size, diagnostic structure and reasoning complexities beyond NP-complete,
we compared our approaches against HS-Tree, a 
widely used
diagnosis computation algorithm with the same properties (soundness, completeness, best-firstness, general applicability) as the proposed methods.
The results testify that RBF-HS, when computing minimum-cardinality diagnoses, scales to large numbers of computed leading diagnoses and achieves a significant memory reduction up to several orders of magnitude for all non-easy problem instances while in addition reducing also the runtime by up to 90\,\% in more than a third of the cases. When used to determine the most probable diagnoses, RBF-HS trades space for time more or less one-to-one compared to HS-Tree. Moreover, 
for both minimum-cardinality and most probable diagnoses, whenever the runtime of RBF-HS was significantly higher than that of HS-Tree, the use of HBF-HS could level this overhead while still reasonably limiting the used memory. Overall, this demonstrates that the suggested techniques allow for \emph{memory-aware model-based diagnosis}, which can contribute, e.g., to the successful diagnosis of memory-restricted devices or memory-intensive problem instances.

Since our approaches are not restricted to diagnosis problems, but applicable to best-first hitting set computation in general, and since a multitude of real-world problems can be formulated as hitting set problems, our methods 
have the potential to impact 
research and application domains beyond the frontiers of model-based diagnosis.

Future work topics include 
the integration of RBF-HS and HBF-HS into the ontology debugging plug-in \emph{OntoDebug}\footnote{See \url{http://isbi.aau.at/ontodebug}.} \cite{schekotihin2018protege,DBLP:conf/foiks/SchekotihinRS18} for \emph{Protégé}\footnote{See \url{https://protege.stanford.edu/}.} \cite{Noy2000}, closer investigations of applications of RBF-HS discussed in Sec.~\ref{sec:impact}, as well as further research 
on hitting set variants of other heuristic search approaches. 

\section*{Acknowledgments}
This work was supported by the Austrian Science Fund
(FWF), contract P-32445-N38. I am grateful to Dietmar Jannach
for many valuable comments that helped me improve this manuscript. 


\appendix

\section{RBF-HS: Algorithm Walkthrough}
\label{apx:algo_walkthrough}
In this appendix, we provide a detailed explication of the workings of (the recursive part of) RBF-HS, based on the recursion structure discussed in Sec.~\ref{sec:recursion_structure}. To this end, we first explain the sub-procedures called by RBF-HS in \ref{apx:sub-procedures}, and then walk through the recursive process in \ref{apx:recursion_details}. 

\subsection{Sub-Procedures}
\label{apx:sub-procedures}
We next explain the workings of the sub-procedures called throughout RBF-HS:
\begin{itemize}[noitemsep]
	\item $\textsc{findMinConflict}(\dpi)$ receives a DPI $\dpi=\tuple{\mo,\mb,\Tp,\Tn}$ and outputs a minimal conflict $\mc \subseteq \mo$ if one exists, and `no conflict' else. A well-known algorithm that can be used to implement this function is \textsc{QuickXplain} \cite{junker04,rodler2020qx}.
	\item $\textsc{add}(x,L)$ takes an object $x$ and a list of objects $L$ as inputs, and returns the list obtained by appending the element $x$ to the end of the list $L$.
	\item $\textsc{addDummyNode}(L)$ takes a list of nodes $L$, appends an artificial node $\node$ with $f(\node) := -\infty$ to $L$, and returns the result.
	\item $\textsc{getAndDeleteFirstNode}(L)$ accepts a sorted list $L$, deletes the first element from $L$ and returns this deleted element.
	\item $\textsc{getFirstNode}(L)$ accepts a sorted list $L$ and returns $L$'s first element.
	\item $\textsc{sortDecreasingByF}(L)$ accepts a list of nodes $L$, sorts $L$ in descending order of $F$-value, and returns the resulting sorted list.
	\item $\textsc{insertSortedByF}(\node,L)$ accepts a node $\node$ and a list of nodes $L$ sorted by $F$-value, and inserts $\node$ into $L$ in a way the sorting of $L$ by $F$-value is preserved.
\end{itemize}

\subsection{Recursion: Details}
\label{apx:recursion_details}
The first argument passed to RBF-HS' (line~\ref{algoline:rbfhs:call_RBFHS'} or \ref{algoline:rbfhs':recursive_call}) is the node $\node$ it will process. \vspace{3pt}

\noindent\emph{Node Labeling.} As a first step, $\node$ 
is labeled by the \textsc{label} function (line~\ref{algoline:rbfhs':label}).
\vspace{5pt}

\noindent\emph{Node Assignment.}
The computed label is then handled 
very similarly as in case of Reiter's HS-Tree (cf.\ 
Example~\ref{ex:Reiter's_HS_Node_Processing}), i.e., $\closed$ nodes are 
discarded, $\valid$ ones added to $\mD$, and those labeled by a conflict $L$ are expanded by the \textsc{expand} function (line~\ref{algoline:rbfhs':expand}). In addition, since a value has to be returned 
by each recursive RBF-HS'-call (cf.\ line~\ref{algoline:rbfhs':recursive_call})
in order for the recursion to be properly resumed, the (worst possible) backed-up $F$-value $-\infty$ is returned for nodes without successors (labels $\closed$ and $\valid$). Intuitively, the value $- \infty$ can be interpreted as ``this node is hopeless or already explored''. The rationale behind this
is to avoid a misleading
of the algorithm towards re-exploring such nodes once their costs would be better than those of all other nodes. In fact, any $F$-value larger than $-\infty$ would even imply the algorithm's non-termination and thus incorrectness, cf.\ \ref{mod:multiple_solutions} on page~\pageref{mod:multiple_solutions}.  

Notably, nodes with $F$-value equal to $-\infty$ \emph{can} be considered again (given their parent nodes are expanded again), but, if so, they are directly labeled $\closed$ in line~\ref{algoline:label:non-min_crit_end} because 
they are either equal to or proper supersets of some node in $\mD$.
Equality holds for nodes originally labeled $\valid$, which are therefore in $\mD$; the superset property is given in case of nodes originally labeled $\closed$, 
for which there was already a proper subset in $\mD$ and thus there still must be one (note: no elements are ever deleted from $\mD$ in Alg.~\ref{algo:RBF_HS}).
This (\emph{inexpensive}) catching of re-explored nodes at the very beginning of \textsc{label} is critical since the \textsc{findMinConflict} operation later in \textsc{label} involves costly theorem prover calls, 
and must thus be performed as rarely as possible (cf.\ Sec.~\ref{sec:conflicts}).\vspace{5pt}

\noindent\emph{Node Expansion.}
Whenever $\node$ is neither a $\closed$ nor a $\valid$ node, it is labeled by a minimal conflict $L$ and its successors $\childnodes$ are created via a call of the \textsc{expand} function (line~\ref{algoline:rbfhs':expand}). The result of this node expansion are $|L|$ nodes, generated as $\node \cup \{\tax_i\}$ for each $\tax_i \in L$ (line~\ref{algoline:expand:add_successor_node}).\vspace{5pt}

\noindent\emph{Node Cost Inheritance.}\footnote{For a detailed argumentation why the assertions about the $f$- and $F$-costs of nodes made in this paragraph hold, please consider the proof of Theorem~\ref{thm:correctness}.}
Next, the $F$-value of each of the newly-generated child nodes $\node_i$ is set (lines~\ref{algoline:rbfhs':for_node_in_childnodes}--\ref{algoline:rbfhs':F(n_i)_gets_f(n_i)}). Note, this is necessary at each node expansion since a (child) node's $F$-value exists only as long as the node is in memory; it is no longer stored after a node is discarded through a backtracking step of the algorithm. 
%
Intuitively, the ideal $F$-value would be: \emph{(a)}~the original $f$-value for child nodes never explored before, for which there cannot be a ``learned'' $F$-value yet, \emph{(b)}~the last known $F$-value for child nodes already explored before.

Basically, there are two possibilities how RBF-HS may specify the $F$-value of a child node $\node_i$: either the $F$-value of the parent $\node$ is inherited to the child node, 
or $\node_i$'s (original) $f$-value is used.
%
In fact, the algorithm first checks whether $\node$ has already been explored before, which is true if $f(\node) > F(\node)$ (line~\ref{algoline:rbfhs':if_f(n)>F(n)}). 

In case $f(\node) > F(\node)$, the child nodes can be partitioned into those that have been explored before, and those that have not. 
For the latter class, we have $F(\node) \geq f(\node_i)$, which involves that each non-explored child node keeps its original $f$-cost ($\min$ in line~\ref{algoline:rbfhs':F(n_i)_gets_min}).
For the former class, it indeed holds that $F(\node) < f(\node_i)$, which is why all already-explored nodes inherit the $F$-value of the parent $\node$ ($\min$ in line~\ref{algoline:rbfhs':F(n_i)_gets_min}). 
Note, the child nodes' last known $F$-value (before they were discarded) might have been lower than the inherited $F(\node)$ because only \emph{one} $F$-value is remembered by the algorithm when a subtree is forgotten; however, $F(\node)$ is at least to some extent lower than $f(\node_i)$ which implies that at least some ``fraction'' of $\node_i$'s already learned backed-up cost is restored by the inheritance.

Alternatively, given $f(\node) = F(\node)$ (note that $f(\node) \leq F(\node)$ for all nodes $\node$ is an invariant throughout RBF-HS'), $\node$ can, but does not need to, have been explored already. If $\node$ has not yet been explored, then clearly none of its child nodes $\node_i$ can have been explored either, which is why it is reasonable to set the $F$-value of all children to their $f$-value (line~\ref{algoline:rbfhs':F(n_i)_gets_f(n_i)}).
Otherwise, i.e., if $\node$ has already been explored before, then the latest backed-up value $F(\node)$ (which was necessarily less than $f(\node)$) must have been forgotten in the course of backtracking steps (which \emph{is} possible, e.g., if one of $\node$'s siblings had a greater $F$-value than $\node$ at the point where RBF-HS' backtracked after exploring $\node$'s parent node). 
Now, since the $f$-value of each node is greater than the $f$-value of any of its successors (anti-monotonicity of $f$, cf.\ Bullet \ref{enum:diff:stricter_conditions_on_cost_function} on page \pageref{enum:diff:stricter_conditions_on_cost_function}), it must hold that \emph{(a)}~$F(\node) = f(\node) > f(\node_i)$ for all child nodes $\node_i$ of $\node$, and \emph{(b)}~any solution in a subtree rooted at some $\node_i$ will have cost lower than or equal to $f(\node_i)$. 
Since the ``learned'' $F$-value for any node should not be a worse estimate of the cost of a solution in the respective subtree than the original estimation given by the node's $f$-value, it does not make sense to set the $F$-value of any child node $\node_i$ to the value $F(\node)$ ($> f (\node_i)$). Hence, it is most plausible also in this case to set the $F$-value of all children to their original $f$-value (line~\ref{algoline:rbfhs':F(n_i)_gets_f(n_i)}). 
\vspace{5pt}

\noindent\emph{Child Node Preparation.}
Once all nodes in $\childnodes$ have been assigned their $F$-value, $\childnodes$ is prepared for node exploration (while-loop, line~\ref{algoline:rbfhs':while}) in the following way: First, if there is only a single node in $\childnodes$, then a second ``dummy'' node is added. The reason for this is that lines~\ref{algoline:rbfhs':getSecondBestChild_1} and \ref{algoline:rbfhs':getSecondBestChild_2} require a second node to be present in $\childnodes$. In order not to compromise the correctness of RBF-HS, the $F$-value of this dummy node has to be set to the worst possible value $-\infty$ (cf.\ argumentation for \emph{Node Assignment} above).
Second, the nodes in $\childnodes$ are sorted in descending order of $F$-value, such that exactly the nodes with the highest and second-highest $F$-value are extracted from $\childnodes$ in lines~\ref{algoline:rbfhs':getBestChild_1} and \ref{algoline:rbfhs':getBestChild_2}, respectively.\vspace{5pt}

\noindent\emph{Recursive Child Node Exploration.} 
Now, as the child nodes have been generated, their $F$-costs have been set, and the list $\childnodes$ has been prepared for being processed, the final block of RBF-HS' involves the best-first exploration of nodes in $\childnodes$ by means of the algorithm's while-loop. Throughout the iteration of the loop, the variables $\node_1$ and $\node_2$ always comprise the best and second-best node, respectively, among $\childnodes$, according to their (backed-up) $F$-value. This is guaranteed by lines~\ref{algoline:rbfhs':insertSortedByF}, \ref{algoline:rbfhs':getBestChild_2}, and \ref{algoline:rbfhs':getSecondBestChild_2}, where \textsc{insertSortedByF} inserts a node to a list such that the sorting of the list according to $F$ is preserved. The while-loop is iterated by always exploring the best node $\node_1$ through a recursive call of RBF-HS' (line~\ref{algoline:rbfhs':recursive_call}) as long as the current $\node_1$'s $F$-value is better than $\bnd$. The latter stores the maximal $F$-value over all child nodes of all ancestors of $\node_1$ (see the $\max$ which determines the bound at each recursive downward step in line~\ref{algoline:rbfhs':recursive_call}). This value at the same time corresponds to the maximal $F$-value of \emph{any} alternative node in the entire hitting set tree, which in turn is greater than or equal to the $f$-value (i.e., the probability $\pr$) of \emph{any} existing solution other than $\node_1$ (see the proof of Theorem~\ref{thm:correctness} for a precise argumentation why these things hold). 
Hence, the use of $\bnd$ as a ruler of backtracking actions guarantees that the most probable (remaining) solution is always found first (next).
At the point where all nodes in $\childnodes$ have an $F$-value lower than $\bnd$, the while-loop is exited and the currently best $F$-value among the nodes in $\childnodes$ is returned, i.e., propagated upward to their parent node $\node$. 
Note, in the course of the recursive explorations of the subtrees rooted at nodes in $\childnodes$ throughout the iteration of the while-loop, solutions might be located and added to $\mD$.\vspace{5pt}

\noindent\emph{Termination.} 
Whenever $\mD$ is extended, a check is run which tests if the list of solutions $\mD$ has already reached the stipulated size $\ld$ (line~\ref{algoline:rbfhs':if_mD_geq_ld}). If so, the RBF-HS' procedure terminates (line~\ref{algoline:rbfhs':exit_procedure}). Otherwise, i.e., if there are fewer than $\ld$ minimal diagnoses existent for the tackled DPI, RBF-HS' terminates once all nodes in the hitting set tree have been explored and assigned the backed-up value $-\infty$, which is why all recursive while-loops must stop (condition in line~\ref{algoline:rbfhs':while}). In any case, RBF-HS finally returns $\mD$ (line~\ref{algoline:rbfhs:return_mD_3}).

\section{RBF-HS: Derivation of Time and Space Complexity}
\label{apx:complexity}
In this appendix, we provide the argumentation that proves Theorem~\ref{thm:complexity} in Sec.~\ref{sec:algo_complexity}. 
\subsection{Time Complexity} 
\label{sec:time_complexity}
We can distinguish between two sources of time complexity inherent in RBF-HS: 
\begin{enumerate}[noitemsep,label=\emph{(t\arabic*)},align=left,leftmargin=0pt,labelwidth=-0.5\parindent] %
	\item \label{enum:time_complexity:reasoning} logical consistency checking, and
	\item \label{enum:time_complexity:tree_management} tree construction and management.
\end{enumerate}

As to \ref{enum:time_complexity:reasoning}, both the \emph{hardness} and the \emph{number of} 
performed consistency checks are of relevance. 

First, the \emph{hardness of consistency checks} executed by RBF-HS depends on the knowledge representation language adopted to model the diagnosed system and thus cannot be generally assessed. It might range from polynomial in the case of Horn logic over NP-complete for propositional system descriptions to even much harder, such as (2)NEXPTIME-complete for some Description Logics \cite{grau2008owl} (cf.\ our evaluation dataset in Sec.~\ref{sec:eval}). Note, despite these somewhat discouraging theoretical complexities, experience with real-world diagnosis cases has shown that practical runtimes for consistency checks are often reasonable, even for interactive scenarios and very expressive logics \cite{Shchekotykhin2014,Shchekotykhin2012,Horridge2011a,Kalyanpur2006a,rodler2019KBS_userstudy,Shearer2008}.

Regarding the \emph{number of consistency checks}, in contrast, we are able to derive the upper bound $O(|\mo| (|\mathbf{minC}| + |\ld|))$ where $\mathbf{minC}$ denotes the set of all minimal conflicts for the DPI dealt with.
To see why this holds, observe that 
\begin{itemize}[noitemsep]
	\item the only place where RBF-HS issues consistency checks is in line~\ref{algoline:label:findMinConflict} (\textsc{findMin-} \textsc{Conflict}),
	\item each \textsc{findMinConflict} call either yields a minimal conflict (line~\ref{algoline:label:return_new_cs}) or a minimal diagnosis (line~\ref{algoline:label:return_valid}),
	\item RBF-HS terminates once the desired $\ld$ minimal diagnoses have been found,
	\item each minimal conflict is actually computed \emph{only once} (but it might be reused multiple times by means of the stored list of conflicts $\mC$), and 
	\item one call of \textsc{findMinConflict} requires $O(|\mo|)$ consistency checks in the worst case \cite{marques2013minimal}
	if a minimal conflict $\mc$ is returned, and only a single check if a minimal diagnosis is found (i.e., `no conflict' is output).
\end{itemize}
Hence, no more than $|\mathbf{minC}| + |\ld|$ calls of \textsc{findMinConflict}, each issuing no more than $|\mo|$ consistency checks, can be made throughout the execution of RBF-HS.

Factor \ref{enum:time_complexity:tree_management} is somewhat harder to estimate, as one and the same node might be explored multiple times (cf., e.g., node $\{2,4\}$, which is processed three times in Example~\ref{ex:RBF-HS}). 
Essentially, there are two main aspects that affect this factor: 
\begin{enumerate}[noitemsep,label=\emph{(\roman*)}]
	\item The larger the number of \emph{different} $f$-values among all nodes is, and
	\item the higher the distribution of promising nodes in the search tree is,
\end{enumerate}
the more backtrackings and node re-explorations RBF-HS will do \cite{hatem2015recursive}. 
%
In the worst case, each node has a different $f$-value and, when sorting all nodes according to their $f$-value, any two neighbors in this sorting are in different subtrees of the root node. In such scenario, $O(n)$ node explorations 
have to be executed per newly expanded node, where $n$ is the number of all nodes in the complete hitting set tree (as constructed by HS-Tree). The reason for this is that each node expansion requires forgetting the entire last explored subtree of the root and expanding another one until the newly expanded node is reached. Since $n$ nodes will be explored overall (as many as HS-Tree explores\footnote{This holds under the assumption that HS-Tree does not close duplicate nodes, i.e.,  the \emph{(duplicate)} criterion is left out, cf.\ Example~\ref{ex:Reiter's_HS_Node_Processing}. In this case, it will explore exactly the same nodes as RBF-HS (which, by construction, cannot eliminate duplicates, cf.\ Sec.~\ref{sec:recursion_structure}),
	except that the latter might explore nodes more than once. Note that we have observed in diverse experiments with HS-Tree that it usually runs faster if the duplicate-check is omitted, because the latter has to explore a potentially exponential-sized collection of nodes at (almost) each processing of a node. The correctness of HS-Tree is not harmed by this modification.\label{footnote:complexity_HS-tree_no_duplicate_check}}), we have a resulting complexity of $O(n^2)$ (cf.\ the analogue argumentation in \cite{hatem2015recursive} for RBFS). 
However, this scenario is only possible when RBF-HS is used to compute diagnoses in the order of decreasing probability (\emph{most probable first}).

If diagnoses should be generated in the order of increasing cardinality (\emph{minimal cardinality first}), in contrast,
we can deduce\footnote{\cite{korf1992linear} derived this for RBFS in comparison to breadth-first search. We can transfer this result to RBF-HS and HS-Tree (without duplicate check, cf.\ Footnote~\ref{footnote:complexity_HS-tree_no_duplicate_check}) for the following reasons: First, HS-Tree performs exactly a breadth-first search when minimum-cardinality diagnoses are sought, due to the $f$-cost of any node $\node$ being reciprocal to its cardinality (tree-depth) $|\node|$ in this case, cf.\ Sec.~\ref{sec:inputs+outputs}. Second, the fact that RBF-HS and HS-Tree usually execute until multiple solutions are found (while RBFS and breadth-first search terminate with the finding of the first solution) is not detrimental to the analysis in \cite{korf1992linear} as its result is independent of the $\goal$ function. In other words, if $k$ diagnoses should be found, the $k$-th found diagnosis is interpreted as the first goal node (and the first $k-1$ diagnoses are simply interpreted as non-goal nodes without successors).} 
from the findings of \cite{korf1992linear} that RBF-HS explores $O(n)$ nodes, i.e., for sufficiently large problem size, no more than a constant number 
as many as HS-Tree does. Intuitively, the plausibility of this can be verified by considering (i) and (ii) above. As to (i), we have only $d$ different node costs (i.e., cardinalities) 
where $d$ is the size of the minimal
diagnosis with maximal cardinality. Regarding (ii), it is straightforward to see that the next explored node of any node $\node$ will be the sibling of $\node$'s closest ancestor\footnote{$\node$ itself is defined to belong to the set of ancestors of $\node$.} which has not been processed in the current iteration.\footnote{Like \cite{korf1992linear}, we define an \emph{iteration of RBF-HS} as ``the interval of time when those nodes being expanded for the first time are all of the same cost.''} Thus, each next-best node will be ``close'' to the current node and a minimum number of backtracking steps will have to be performed to reach the next-best node from the current one. 
\subsection{Space Complexity}
\label{sec:space_complexity}
First, the space complexity of Korf's original RBFS algorithm, that acts as a basis for RBF-HS, is linear \cite{korf1992linear}, i.e., in $O(bd)$ where $b$ is the maximal number of successor states of any state (a.k.a.~branching factor) and $d$ the maximal length of any path in the search space. 
Second, no amendments to the recursive (depth-first) nature of RBFS have been made while deriving RBF-HS (cf.\ Sec.~\ref{sec:necessary_modifications}). 
Third, RBF-HS stores computed minimal conflicts and minimal diagnoses, information RBFS does not need. In RBF-HS, recorded conflicts allow for a more efficient labeling of nodes (reuse instead of recalculation), whereas the storage of diagnoses is essential for the algorithm's correctness and moreover trivially necessary as diagnoses constitute exactly the solutions which should finally be returned.

Hence, the space complexity of RBF-HS is affected by three factors: 
\begin{enumerate}[noitemsep,label=\emph{(s\arabic*)},align=left,leftmargin=0pt,labelwidth=-0.5\parindent] %
	\item \label{enum:space_complexity:mD} $|\mD|$ (number of stored minimal diagnoses),
	\item \label{enum:space_complexity:mC} $|\mC|$ (number of stored minimal conflicts), and
	\item \label{enum:space_complexity:tree} the space required to store the search tree.
\end{enumerate}
Factor \ref{enum:space_complexity:mD} is bounded by the fixed input argument $\ld$, which is arbitrarily preset by the user of RBF-HS, and thus in $O(1)$.\footnote{Note, if $\ld := \infty$ is specified, 
expressing the intention 
to find \emph{all} minimal diagnoses for the 
DPI, then $\ld$ is not a constant, i.e., not in $O(1)$, but conditioned by the problem size (number of existing minimal diagnoses). 
Obviously, the existence of a generally linear algorithm to accomplish that task is theoretically impossible since the mere maintenance of the collection of (potentially exponentially many) solutions $\mD$ might require more than linear space.}
Factor~\ref{enum:space_complexity:mC} is bounded by $|\mathbf{minC}|$ where $\mathbf{minC}$ is the set of all minimal conflicts for the considered DPI. Analogously to RBFS, factor~\ref{enum:space_complexity:tree} is bounded by $|\mc_{\max}|*|\mathbf{minC}|$ where $\mc_{\max}$ is the minimal conflict for DPI with maximal cardinality. The explanation for this is that 
\begin{itemize}[noitemsep]
	\item no node can have more than $|\mc_{\max}|$ child nodes (reason: exactly $k$ successors result from a node-labeling conflict of size $k$, cf.\ \textsc{expand} function in Alg.~\ref{algo:RBF_HS}; no other ways of successor generation exist in RBF-HS, cf.\ Alg.~\ref{algo:RBF_HS}),
	\item no node (set of edge labels along tree path) can include more than $|\mathbf{minC}|$ elements (reason: any node including $|\mathbf{minC}|$ elements must hit all minimal conflicts and thus must be a diagnosis; diagnoses are labeled $\valid$ or $\closed$ and never further expanded by RBF-HS), and
	\item at any tree depth, only a single node can be expanded at one particular point in time (reason: depth-first recursion, line~\ref{algoline:rbfhs':recursive_call}).
\end{itemize}
All in all, given finite $\ld$, we thus have a space complexity of $O(|\mc_{\max}|*|\mathbf{minC}|)$ which can be interpreted as branching factor ($b$) times maximal depth ($d$), equivalently as for RBFS.

Experience in the diagnosis field suggests that generally
the number of minimal conflicts does not depend on or grow with the size of the diagnosed system.\footnote{Exceptions do however exist, see, e.g., \cite{dekleer1991focusing_prob_diag}.} There are small systems with a higher number of minimal conflicts, as well as there are huge systems with negligible numbers of minimal conflicts. So, from an empirical perspective it appears to be
in many cases justified to interpret $|\mathbf{minC}|$ to be in $O(1)$, i.e., to be independent of the size of the given DPI. This assumption implies that RBF-HS requires space linear in the size of the DPI $\tuple{\mo,\mb,\Tp,\Tn}$, because clearly $|\mc_{\max}| \leq |\mo|$ due to $\mc_{\max} \subseteq \mo$ (cf.\ Sec.~\ref{sec:conflicts}). Note, if both $b$ and $d$ are assumed to be not in $O(1)$ (i.e., are dependent on the problem size), then also the original RBFS algorithm loses its linear space bounds.

\subsection{Summary}
The argumentations provided in \ref{sec:time_complexity} and \ref{sec:space_complexity} prove Theorem~\ref{thm:complexity} in Sec.~\ref{sec:algo_complexity}, quoted below:\vspace{3pt}\\ 
\textbf{Theorem~\ref{thm:complexity}} (Complexity of RBF-HS)\textbf{.}\hspace{4pt}
\emph{
Let $\dpi = \langle\mo,\mb,\Tp,\Tn\rangle$ be an arbitrary DPI, $\ld$ the finite positive natural number of diagnoses to be computed, $n$ the number of nodes expanded by HS-Tree (without the duplicate criterion) for $\dpi$ and $\ld$, $t_{\mathit{CC}}$ the worst-case time of a consistency check for $\dpi$, $\mathbf{minC}$ the set of all minimal conflicts for $\dpi$, and $\mc_{\max}$ the conflict of maximal size for $\dpi$. Further, let $\mathit{TPT} := t_{\mathit{CC}} |\mo|(|\mathbf{minC}| + |\ld|)$ (theorem proving time).
Finally, assume $|\mathbf{minC}|$ is in $O(1)$, i.e., independent of the size of $\dpi$.
Then:
\begin{itemize}[noitemsep]
	\item \emph{Time Complexity:} 
	RBF-HS requires time in $O(n + \mathit{TPT})$ for the computation of the $\ld$ diagnoses of minimal cardinality for $\dpi$; and time in $O(n^2 + \mathit{TPT})$ for the computation of the $\ld$ most probable diagnoses for $\dpi$. 
	\item \emph{Space Complexity:} RBF-HS requires space in $O(|\mo|)$.
\end{itemize}}

\section{RBF-HS: Proof of Correctness}
\label{apx:proof}
We next show the validity of the theorem below, which is stated in Sec.~\ref{sec:algo_correctness}:\vspace{3pt}\\ 
\textbf{Theorem~\ref{thm:correctness}}(Correctness of RBF-HS)\textbf{.}\hspace{4pt}
\emph{Let \textsc{findMinConflict} be a sound and complete method for conflict computation, i.e., given a DPI, it outputs a minimal conflict for this DPI if a minimal conflict exists, and `no conflict' otherwise. RBF-HS is sound, complete and best-first, i.e., it 
computes \emph{all} and \emph{only} minimal diagnoses \emph{in descending order} \emph{of probability} as per the strictly antimonotonic probability measure $\pr$.}\vspace{4pt}\\
Before we are able to state the proof of Theorem~\ref{thm:correctness}, we formulate some useful definitions and lemmas that will help us keep the proof relatively concise.
\subsection{Preparation for the Proof}
Throughout this section, we will often say that RBF-HS generates or processes a node. The following definition makes precise what we mean by this:
\begin{definition}
	We say that \emph{(i)}~a \emph{node $\node$ is generated} by RBF-HS iff $\node$ is an element of $\childnodes$ in line~\ref{algoline:rbfhs':expand}, 
	and \emph{(ii)}~a \emph{node $\node$ is processed} by RBF-HS iff a call of $\text{RBF-HS'}(\node,\_,\_)$ (line~\ref{algoline:rbfhs:call_RBFHS'} or \ref{algoline:rbfhs':recursive_call}) is executed. (The ``$\_$'' signifies that the other input arguments to RBF-HS' do not matter.)
\end{definition}
\begin{lemma}\label{lem:only_diags_can_be_added_to_mD}
	In RBF-HS, only diagnoses can be added to the collection $\mD$.
\end{lemma}
\begin{proof}
	Let us start backwards from line~\ref{algoline:rbfhs:add_node_to_mD}, which is the only place in RBF-HS where elements are added to $\mD$. The condition that must be fulfilled for this line to be reached is that $L = \valid$ must be returned for the currently processed node $\node$ that is added to $\mD$. Considering the \textsc{label} function, we find that it must return in line~\ref{algoline:label:return_valid} which in turn requires that $\textsc{findMinConflict}(\langle\mo\setminus\node,\mb,\Tp,\Tn\rangle)$ before must have returned `no conflict'. This means that $\mo \setminus \node$ does not contain a minimal conflict, or, equivalently, is not a conflict. By the Duality Property (cf.\ Sec.~\ref{sec:MBD}), we obtain that $\node$ is a diagnosis.
\end{proof}
\begin{lemma}\label{lem:if_recursion_is_entered_then_a_diag_exists}
	If line~\ref{algoline:rbfhs:call_RBFHS'} is executed, then a non-empty minimal diagnosis exists.
\end{lemma}
\begin{proof}
	The statement of this lemma follows from the algorithm's analysis (lines~\ref{algoline:rbfhs:mc=emptyset} and \ref{algoline:rbfhs:mc='no_conflict'}) of the output of the \textsc{findMinConflict} call in line~\ref{algoline:rbfhs:findMinConflict} along with the Duality Property (cf.\ Sec.~\ref{sec:MBD}). See 
	Sec.~\ref{sec:trivial_cases} for a more detailed argumentation. 
\end{proof}
\begin{lemma}\label{lem:if_diag_processed_then_added_to_mD} 
	If a node $\node$ corresponding to a minimal diagnosis $\md$ is processed for the first time by RBF-HS, 
	then $\node$ will be (directly) added to $\mD$ in line~\ref{algoline:rbfhs:add_node_to_mD}.\\
	(Equivalently: After any call of RBF-HS' which processes a node $\node$ corresponding to $\md$ returns, $\md$ is an element of $\mD$.)  
\end{lemma}
\begin{proof}
	%
	Assume that, for the first time throughout the execution of RBF-HS, a node $\node$ equal to $\md$ is processed, where $\md$ is a minimal diagnosis. Initially, in line~\ref{algoline:rbfhs':label}, a label $L$ is computed for $\node$. Within the $\textsc{label}$ function, the first thing executed is the non-minimality check in lines~\ref{algoline:label:non-min_crit_start}--\ref{algoline:label:non-min_crit_end}, where a node $\node_i$ is sought in $\mD$ which is a subset of $\node$. Since \emph{(1)}~only diagnoses can be in $\mD$ as per Lemma~\ref{lem:only_diags_can_be_added_to_mD}, \emph{(2)}~$\node = \md$ is a minimal diagnosis, and \emph{(3)}~it is the first time that a node equal to $\md$ is processed, there cannot be any subset $\node_i$ of $\node$ in $\mD$. Hence, line~\ref{algoline:label:reuse_start} is reached. Due to the Hitting Set Property (cf.\ Sec.~\ref{sec:MBD}) and the fact that $\node$ is a (minimal) diagnosis, there cannot be any (minimal) conflict $\mc$ such that $\mc \cap \node = \emptyset$. Consequently, line~\ref{algoline:label:findMinConflict} is reached. The \textsc{findMinConflict} call in line~\ref{algoline:label:findMinConflict} will return `no conflict' due to the Duality Property and because $\node$ is a diagnosis.
	As a result, \textsc{label} will return in line~\ref{algoline:label:return_valid}, which means that $\node$ will be added to $\mD$ in line~\ref{algoline:rbfhs:add_node_to_mD}.\\
	(The equivalent statement of the lemma holds since no element once added to $\mD$ can ever be removed from it, for the simple reason that there is no statement in RBF-HS that modifies $\mD$ except for the one that adds elements to $\mD$ in line~\ref{algoline:rbfhs:add_node_to_mD}.)
\end{proof}

\begin{lemma}\label{lem:for_each_RBF-HS'_call_a_value_lower_than_F(n)_is_returned} 
	For any call RBF-HS'($\node,F(\node),\bnd$), a value $X < F(\node)$ is returned (unless the RBF-HS'-procedure is exited in line~\ref{algoline:rbfhs':exit_procedure} before a return takes place).
\end{lemma}
\begin{proof}
	Assume an execution of some call of RBF-HS'($\node,F(\node),\bnd$) throughout which no exit of the RBF-HS'-procedure takes place in line~\ref{algoline:rbfhs':exit_procedure}.
	Observe that there are three spots where RBF-HS'might return, i.e., in any of the lines~\ref{algoline:rbfhs':return_after_closed}, \ref{algoline:rbfhs':return_after_valid} or \ref{algoline:rbfhs':return_F(n)}. For the returns in lines~\ref{algoline:rbfhs':return_after_closed} and \ref{algoline:rbfhs':return_after_valid},
	$-\infty$ is returned. However, $F(\node) > -\infty$ must hold. To prove this, let us consider the two places where the RBF-HS'-call can have been issued, i.e., lines~\ref{algoline:rbfhs:call_RBFHS'} or \ref{algoline:rbfhs':recursive_call}. In the former case, $F(n)$ is equal to $f(\emptyset)$, which can only attain values in $(0,1)$ (cf.\ Sec.~\ref{sec:MBD}). In the latter case,
	$\node$ is equal to a child node $\node_1$ of some node and $F(\node) = F(\node_1) >-\infty$ due to the while-condition in line~\ref{algoline:rbfhs':while}. Therefore, the statement of the lemma holds for the returns in lines~\ref{algoline:rbfhs':return_after_closed} and \ref{algoline:rbfhs':return_after_valid}.
	
	For the return in line~\ref{algoline:rbfhs':return_F(n)}, we first point out that, for any call RBF-HS'($\node,F(\node)$, $\bnd$), $F(\node) \geq \bnd$ must hold. To see this, consider again lines~\ref{algoline:rbfhs:call_RBFHS'} and \ref{algoline:rbfhs':recursive_call}, where RBF-HS' can be invoked. In the former case, $\bnd = -\infty$ and $F(\node) > \bnd$ follows from the argumentation in the previous paragraph. In the second case, as explained above, $\node$ is equal to a child node $\node_1$ of some node. Through the while-condition, we thus know that $F(\node) = F(\node_1)$ is larger than or equal to the old value of the bound. Moreover, we know by the sorting of $\childnodes$ and the fact that $\node_1$ is the node in $\childnodes$ with the largest $F$-value (due to lines~\ref{algoline:rbfhs':sortDecreasingByF}, \ref{algoline:rbfhs':getBestChild_1}, \ref{algoline:rbfhs':insertSortedByF} and \ref{algoline:rbfhs':getBestChild_2}), that $F(\node) = F(\node_1) \geq F(\node_2)$ for the node $\node_2$ with second-largest $F$-value in $\childnodes$ (cf.\ lines~\ref{algoline:rbfhs':getSecondBestChild_1} and \ref{algoline:rbfhs':getSecondBestChild_2}). Since $\bnd$ is defined as the maximum among the old value of bound and $F(\node_2)$, $F(\node) \geq \bnd$ must be true.
	
	Finally, note that a return in line~\ref{algoline:rbfhs':return_F(n)}, which is our current assumption, can only take place if the condition of the while-loop is violated. This implies that the returned value ($F(\node_1)$) is either equal to $-\infty$ or strictly less than $\bnd$. As $F(\node)$ must be greater than $-\infty$, as demonstrated in the first paragraph of this proof, we deduce that the 
	statement of the lemma also holds for the return in lines~\ref{algoline:rbfhs':return_F(n)}.
	%
\end{proof}

\begin{lemma}\label{lem:invariant_F(n)<=f(n)_holds_throughout_entire_execution} 
	Throughout the entire execution of RBF-HS and for any node $\node$, the following invariant holds: $F(\node) \leq f(\node)$.
\end{lemma}
\begin{proof}
	First, note that each node's $f$-value remains constant throughout the entire execution of RBF-HS. We now prove that \emph{(1)}~when set, each node's $F$-value is smaller than or equal to its $f$-value, and that \emph{(2)}~as long as a node (and thus its $F$-value) remains in memory, its $F$-value can never increase.
	
	Proof of (1): Here, we consider the root node and all other nodes separately. First, the root node's $F$-value is set to its $f$-value in line~\ref{algoline:rbfhs:call_RBFHS'}, i.e., $F(\node) = f(\node)$ holds.
	Second, each other node's $F$-value is set in line~\ref{algoline:rbfhs':F(n_i)_gets_min} or \ref{algoline:rbfhs':F(n_i)_gets_f(n_i)} after it is (re)constructed in line~\ref{algoline:rbfhs':expand}. At this stage, the new node is refereed to as $\node_i$ in the Alg.~\ref{algo:RBF_HS}. If line~\ref{algoline:rbfhs':F(n_i)_gets_min} applies, then clearly $F(\node_i) = \min(F(\node),f(\node_i)) \leq f(\node_i)$. If $\node_i$'s $F$-value is defined in line~\ref{algoline:rbfhs':F(n_i)_gets_f(n_i)}, then obviously $F(\node_i) = f(\node_i)$. This completes the proof of (1).
	
	Proof of (2): Here, we need to demonstrate that the $F$-value of any node $\node$ constructed in line~\ref{algoline:rbfhs':expand} is never increased until $\node$ is discarded by a backtracking step. The relevant backtracking step is issued by the execution of line~\ref{algoline:rbfhs':return_F(n)} of the \emph{same} RBF-HS'-call where $\node$ was constructed. Therefore, we only need to show that $F(\node)$ cannot increase between lines~\ref{algoline:rbfhs':if_|childnodes|=1} and \ref{algoline:rbfhs':return_F(n)}. The only place where $F$-values of nodes are modified in this part of the algorithm is line~\ref{algoline:rbfhs':recursive_call}, where the $F$-value of a node $\node_1$ is adapted by processing $\node_1$. By means of Lemma~\ref{lem:for_each_RBF-HS'_call_a_value_lower_than_F(n)_is_returned}, we infer that $\node_1$'s new $F$-value must be strictly lower than its old $F$-value, for arbitrary nodes $\node_1$.  
\end{proof}
\begin{lemma}\label{lem:only_nodes_that_are_diags_can_be_labeled_valid_or_nonmin} 
	RBF-HS labels a node $\closed$ or $\valid$ iff it corresponds to a diagnosis.
\end{lemma}
\begin{proof}
	To show the bi-implication, we show both implications $\Rightarrow$ and $\Leftarrow$.
	
	$\Rightarrow$: Assume a node $\node$ that is labeled $\closed$ or $\valid$ by RBF-HS. This implies that $\node$ is processed and that the call $\textsc{label}(\node)$ returns either in line~\ref{algoline:label:non-min_crit_end} or \ref{algoline:label:return_valid}. In the former case, due to Lemma~\ref{lem:only_diags_can_be_added_to_mD}, we have that there is some diagnosis $\node_i$ such that $\node \supseteq \node_i$, which entails that $\node$ is a diagnosis. 
	In the latter case, the call of the (sound and complete) \textsc{findMinConflict} function in line~\ref{algoline:label:findMinConflict} must have returned `no conflict' in order for line~\ref{algoline:label:return_valid} to be reached, which guarantees that $\node$ is a diagnosis.
	
	$\Leftarrow$: Assume a node $\node$ equal to a diagnosis is labeled, i.e., $\textsc{label}(\node)$ is executed. 
	First, due to the Hitting Set Property (cf.\ Sec.~\ref{sec:MBD}) and the fact that $\node$ is a diagnosis, there cannot be any (minimal) conflict $\mc$ such that $\mc \cap \node = \emptyset$. Moreover, all elements (if any) in the collection $\mC$ must be minimal conflicts due to the soundness and completeness (wrt.\ the computation of minimal conflicts) of the \textsc{findMinConflict} function. Hence, \textsc{label} cannot return in line~\ref{algoline:label:return_C}. 
	Second, if \textsc{findMinConflict} is called for the DPI $\langle\mo \setminus \node, \mb, \Tp, \Tn \rangle$, then it must return `no conflict'.
	This holds due to the soundness of \textsc{findMinConflict}, the Duality Property (cf.\ Sec.~\ref{sec:MBD}), and the fact that $\node$ is a diagnosis. As a conclusion, \textsc{label} cannot return in line~\ref{algoline:label:return_new_cs}.
	Overall, since there are exactly four possible lines where $\textsc{label}(\node)$ might return, and
	lines~\ref{algoline:label:return_C} as well as \ref{algoline:label:return_new_cs} are impossible, we obtain that a return must take place in either line~\ref{algoline:label:non-min_crit_end} or \ref{algoline:label:return_valid}.
\end{proof}
\begin{lemma}\label{lem:before_node_n_processed_for_first_time_F(n)=f(n)} 
	Let $\node$ be an arbitrary node. Before $\node$ is processed for the very first time throughout the execution of RBF-HS, $F(\node) = f(\node)$ holds whenever $\node$ is generated. \\
	(Note: By contraposition of this statement, along with Lemma~\ref{lem:invariant_F(n)<=f(n)_holds_throughout_entire_execution}, we obtain:  If $F(\node) < f(\node)$, then $\node$ was already processed at least once.)
\end{lemma}
\begin{proof}
	We prove this lemma by induction based on the tree depth $d = |\node|$ of $\node$.
	
	\emph{Induction Base:} Let $d=1$. First, observe that the $F$-value of the root node $\emptyset$ is equal to its $f$-value due to lines~\ref{algoline:rbfhs:call_RBFHS'} and \ref{algoline:rbfhs':procedure_RBF-HS'} (cf.\ the second argument of RBF-HS' in both lines). Therefore, at the (very first) RBF-HS'-call that processes the root node $\node$, the if-condition in line~\ref{algoline:rbfhs':if_f(n)>F(n)} is false for all child nodes $\node_i$ of $\node$. Hence, line~\ref{algoline:rbfhs':F(n_i)_gets_f(n_i)} is executed for all $\node_i$, which is why $F(\node_i) = f(\node_i)$ for all $\node_i$. However, the nodes $\node_i$ are exactly the nodes at depth $d = 1$ because $|\node_i| = 1$. Consequently, the proposition of the lemma holds for $d=1$.
	
	\emph{Induction Assumption:} Assume the proposition of the lemma holds for $d = k$.
	
	\emph{Induction Step:} Let $\node$ be a node at depth $d = k+1$, and let $\node$ be generated (line~\ref{algoline:rbfhs':expand}). Let us denote the parent node of $\node$ by $\node_p$. That is, $\node_p$ is the node that is currently being processed when line~\ref{algoline:rbfhs':expand}, at which $\node$ is generated, is executed. There are now two cases: either \emph{(a)}~$\node_p$ is currently being processed for the very first time during the execution of RBF-HS, or \emph{(b)}~$\node_p$ has already been processed before. 
	
	Assume (a): Since $\node_p$ is a node at depth $k$, we obtain by the Induction Assumption that $F(\node_p) = f(\node_p)$. This implies that the if-condition in line~\ref{algoline:rbfhs':if_f(n)>F(n)} is false. Thus, line~\ref{algoline:rbfhs':F(n_i)_gets_f(n_i)} is executed for $\node$ and $F(\node) = f(\node)$.
	
	Now suppose (b): Here, we know that $\node$ must have already been generated and subsequently discarded in the past since its parent $\node_p$ was already processed. By the argumentation for case (a), we know that $F(\node) = f(\node)$ was true at the very first processing of $\node_p$. 
	When the execution of this respective RBF-HS'-call (the one that processed $\node_p$ for the very first time) ended, the backed-up $F$-value $X$ returned and set as the new $F$-value of $\node_p$ (i.e., $F(\node_p) = X$) was the maximal $F$-value of any child node of $\node_p$ at this time (lines~\ref{algoline:rbfhs':sortDecreasingByF}, \ref{algoline:rbfhs':getBestChild_1}, \ref{algoline:rbfhs':insertSortedByF} and \ref{algoline:rbfhs':getBestChild_2}).
	Since $\node$ was never processed so far by assumption, and since $F(\node) = f(\node)$ was true when $\node$ was first generated, this must still have been true when $X$ was returned. Hence, $X \geq F(\node) = f(\node)$ was true after the termination of the said RBF-HS'-call. Since the maximal $F$-value over all child nodes is returned whenever the processing of a node for which children were generated terminates, a value greater than or equal to $X$ is backed-up if the processing of $\node_p$'s parent node ends. The same holds recursively for any other ancestor of $\node_p$ until ancestors of depth $1$. Note, all nodes at depth $1$ remain in memory throughout the entire execution of RBF-HS and RBF-HS cannot terminate since $\node$ is generated again by assumption.
	
	Since $\node$ is generated again by assumption, each of these ancestors must be processed again. Whenever child nodes for any of these ancestors $\node_a$ are (re)generated, $F(\node_a) \geq X$ and each child node's $F$-value is either set to its $f$-value or to $F(\node_a)$, due to lines~\ref{algoline:rbfhs':F(n_i)_gets_min} and \ref{algoline:rbfhs':F(n_i)_gets_f(n_i)}. Hence, $F(\node_p) \geq X$ or $F(\node_p) = f(\node_p)$ when $\node_p$ is (re)generated. Now, when $\node$ is (re)generated, either line~\ref{algoline:rbfhs':F(n_i)_gets_min} or \ref{algoline:rbfhs':F(n_i)_gets_f(n_i)} is executed to set $\node$'s $F$-value. If line~\ref{algoline:rbfhs':F(n_i)_gets_f(n_i)} applies, we have that $F(\node) = f(\node)$. Therefore, suppose line~\ref{algoline:rbfhs':F(n_i)_gets_min} is the one executed. As we have shown that $F(\node_p) \geq X \geq f(\node)$, we can deduce $F(\node) = \min(F(\node_p),f(\node)) = f(\node)$ also in this case. 
	
	Regardless of how often the parent $\node_p$ is processed, the same argumentation can be applied to derive that $F(\node) = f(\node)$ will hold whenever $\node$ is (re)generated. This completes the inductive proof.
\end{proof}

\subsection{Proof of Theorem~\ref{thm:correctness}}
We demonstrate the correctness of RBF-HS in the following order: termination, completeness, best-first property, and finally soundness.

\subsubsection{Termination}
Assume the RBF-HS does not terminate. The only possibilities for non-termination are that \emph{(i)}~one of the for-loops (lines~\ref{algoline:rbfhs':for_node_in_childnodes}, \ref{algoline:label:non-min_crit_start}, \ref{algoline:label:reuse_start}, and \ref{algoline:expand:for_loop}) is iterated forever, \emph{(ii)}~the recursion is iterated forever (i.e., infinitely many calls of RBF-HS' are made), or \emph{(iii)}~the while-loop during some RBF-HS'-call is iterated forever.

Assume (i): 
Let us consider the four for-loops in turn in chronological order by the point in time at which they are executed by one call of RBF-HS'.\\ 
\emph{(Line~\ref{algoline:label:non-min_crit_start}):} Since $\md \subseteq \mo$ holds for any diagnosis $\md$ and $|\mo|$ is finite (cf.\ Sec.~\ref{sec:MBD}), the for-loop must terminate.\\
\emph{(Line~\ref{algoline:label:reuse_start}):} Since $\mc \subseteq \mo$ holds for any conflict $\mc$ and $|\mo|$ is finite (cf.\ Sec.~\ref{sec:MBD}), the for-loop must terminate.\\
\emph{(Line~\ref{algoline:expand:for_loop}):} Each label $L$ other than $\closed$ or $\valid$ output by the \textsc{label} function must 
have been computed (either freshly or at an earlier stage) by the \textsc{findMinConflict} function. 
By the soundness of \textsc{findMinConflict}, $L$ must be a conflict in this case. Hence, the argument $L$ forwarded to the \textsc{expand} function must be a conflict. Since $\mc \subseteq \mo$ holds for any conflict $\mc$, $|\mo|$ is finite (cf.\ Sec.~\ref{sec:MBD}), and the for-loop iterated by the \textsc{expand} function processes each element of $L$ once, this for-loop must terminate.\\
\emph{(Line~\ref{algoline:rbfhs':for_node_in_childnodes}):} This for-loop iterates once through the set of child nodes output by the \textsc{expand} function, which must be finite by the argumentation above. Hence, this for-loop must terminate as well.
%

Assume (ii): For each recursive call in line~\ref{algoline:rbfhs':recursive_call}, the node $\node_1$, for which the call is made, is one of the child nodes of the old node $\node_1$, for which the call in line~\ref{algoline:rbfhs':recursive_call} was made one recursion level higher. Due to the construction of child nodes (line~\ref{algoline:expand:add_successor_node}), and because lines~\ref{algoline:label:if_C_cap_node=emptyset} and \ref{algoline:label:findMinConflict} guarantee that the conflict used to label the parent node is disjoint with the parent node, 
it follows that each child node has exactly one more element than its parent. 
Moreover, each element $e$ added to a node, and thus each element $e$ of any node, is an element of some conflict $\mc \subseteq \mo$, which entails that $e \in \mo$. This holds due to line~\ref{algoline:expand:add_successor_node} and the fact that the argument $L$ passed to \textsc{expand} must be a conflict, as argued above. 
So far, we have shown that each node along any tree branch is a subset of $\mo$ and each recursive downward step along the branch adds exactly one element to a node. 

Additionally, due to Lemma~\ref{lem:if_recursion_is_entered_then_a_diag_exists} and since line~\ref{algoline:rbfhs':recursive_call} was executed by assumption, we know that a diagnosis exists. By definition, each (minimal) diagnosis is a subset of $\mo$. In particular, this means that $\mo$ is necessarily a diagnosis. 
By Lemma~\ref{lem:only_nodes_that_are_diags_can_be_labeled_valid_or_nonmin}, any processed node corresponding to a diagnosis will be labeled $\valid$ or $\closed$, which prompts a return in either line~\ref{algoline:rbfhs':return_after_closed} or \ref{algoline:rbfhs':return_after_valid} and thus prevents any further recursive RBF-HS'-calls along this tree branch. 
Now, our assumption of an infinite sequence of recursive RBF-HS'-calls means that no node along this branch can be labeled $\closed$ or $\valid$. Due to $|\mo|< \infty$ (cf.\ Sec.~\ref{sec:MBD}), this yields a contradiction 
since any such infinite branch must 
at some stage process the node $\node = \mo$ which is a diagnosis. 

Assume (iii):
%
First, note that $\bnd$ occurring in the while-condition is fixed throughout the execution of one and the same while-loop. 
Second, by Lemma~\ref{lem:for_each_RBF-HS'_call_a_value_lower_than_F(n)_is_returned}, each of the (infinitely many) calls of RBF-HS' executed during the while-loop execution decreases the $F$-value of the processed node.
Third, each $F$-value returned by RBF-HS' in line~\ref{algoline:rbfhs':recursive_call} is equal to either $-\infty$ or to some original $f$-value of some node (see all return-statements throughout RBF-HS' and lines~\ref{algoline:rbfhs':F(n_i)_gets_min} and \ref{algoline:rbfhs':F(n_i)_gets_f(n_i)}). Because there are only finitely many possible nodes (each node is a subset of $\mo$ where $|\mo|<\infty$), there can also be only finitely many different $f$-values $\{f_1,\dots,f_q\}$ of nodes, which is why we can find a fixed $\epsilon > 0$ such that for all $f_i \neq f_j$ we have $|f_i - f_j| > \epsilon$. Hence, for any node $\node$, the reduction of the F-value by means of one RBF-HS'-call must be greater than $\epsilon$. 
Fourth, once its $F$-value is below $\bnd$, a node cannot be processed again during this while-loop execution (while-condition). 
Fifth, there are always finitely many child nodes which are processed by the while-loop (see above). 
From these five points, we conclude that, after a finite number of iterations, the while-condition must be violated. Contradiction.

\subsubsection{Completeness}
%
%
%
Let $\ld := \infty$ (all existing minimal diagnoses should be found) and let there be a minimal diagnosis $\md'$ such that $\md' \notin \mD$ for the collection $\mD$ returned by RBF-HS. The return can take place in lines~\ref{algoline:rbfhs:return_mD_1}, \ref{algoline:rbfhs:return_mD_2} or \ref{algoline:rbfhs:return_mD_3}. 
Line~\ref{algoline:rbfhs:return_mD_1} cannot apply since, in this case, the \textsc{findMinConflict} call in line~\ref{algoline:rbfhs:findMinConflict} returns $\emptyset$, which means that there cannot be any diagnosis by the Duality Property---this is a contradiction to our assumption that $\md'$ is a diagnosis. If line~\ref{algoline:rbfhs:return_mD_2} applies, then `no conflict' was output by \textsc{findMinConflict} in line~\ref{algoline:rbfhs:findMinConflict}, which implies that $\emptyset$ is the only diagnosis, again by the Duality Property. Hence, $\md' = \emptyset$ must hold. Since $\mD = [\emptyset]$ is returned, we have a contradiction to the assumption that $\md'$ is not returned. 

Finally, let the return of $\mD$ be in line~\ref{algoline:rbfhs:return_mD_3}. This means that RBF-HS' must have been called in line~\ref{algoline:rbfhs:call_RBFHS'}. By Lemma~\ref{lem:if_diag_processed_then_added_to_mD}, our assumption from above can be stated as: No node corresponding to $\md'$ is processed throughout the execution of RBF-HS'. 
First, note that, for each minimal diagnosis, there is a possible path from the root to that diagnosis, due to the Hitting Set Property (i.e., each diagnosis, in particular $\md'$, includes some element of every minimal conflict) and the fact that RBF-HS' can generate a node equal to $\md'$ by starting with the empty (root) node (cf.\ line~\ref{algoline:rbfhs:call_RBFHS'}), labeling it with a minimal conflict $\mc_1$ (see \textsc{label} function, line~\ref{algoline:label:return_C} or \ref{algoline:label:return_new_cs}), and by selecting a child node equal to $\setof{x}$ for some element $x \in \mc_1 \cap \md'$, and labeling this child again with a conflict $\mc_2 \cap \setof{x} = \emptyset$, and so on. 
We next show that each node $\node \subseteq \md'$ along some path from the root to $\md'$ will be processed.

First, let us assume that some node $\node' \subseteq \md'$ of cardinality $k \geq 1$ is generated, but never processed. By Lemma~\ref{lem:before_node_n_processed_for_first_time_F(n)=f(n)}, it follows that $F(\node') = f(\node') > 0 > -\infty$ will hold throughout the entire execution of RBF-HS'. 
Since RBF-HS terminates, any RBF-HS'-call for the parent node $\node'_p$ of $\node'$ must return, and since $\node'$ (i.e., a child node) was generated it must return exactly in line~\ref{algoline:rbfhs':return_F(n)}. 
(Note that $\node'_p (\subset \md')$ can be processed multiple times; however, each time the respective RBF-HS'-call that processes $\node'_p$ will return in line~\ref{algoline:rbfhs':return_F(n)} since \emph{(1)}~$\md'$ is a \emph{minimal} diagnosis by assumption, \emph{(2)}~only diagnoses can be labeled $\valid$ or $\closed$ by Lemma~\ref{lem:only_nodes_that_are_diags_can_be_labeled_valid_or_nonmin}, and \emph{(3)} $\ld = \infty$ ensures that line~\ref{algoline:rbfhs':exit_procedure} can never be executed.)
Thus, for any call that processes $\node'_p$, the returned value $F(\node'_p) \geq F(\node') > -\infty$ (due to the sorting of $\childnodes$, see lines~\ref{algoline:rbfhs':sortDecreasingByF} and \ref{algoline:rbfhs':insertSortedByF}, and due to the fact that the child node with maximal F-value is always returned, see lines~\ref{algoline:rbfhs':getBestChild_1} and \ref{algoline:rbfhs':getBestChild_2}). 
The same argumentation can be applied along the branch from $\node'_p$ to the root node, until the new $\node'_p$ is equal to the root. Finally, we can derive that $F(\node_1) \geq F(\node') > -\infty$ will hold throughout the entire execution of the first call of RBF-HS' made in line~\ref{algoline:rbfhs:call_RBFHS'}, which means that the condition of the while-loop is satisfied forever (recall that $\bnd = -\infty$ at the first RBF-HS'-call in line~\ref{algoline:rbfhs:call_RBFHS'}). This is a contradiction to the fact that RBF-HS always terminates. Thus, we have demonstrated that, for $k \in\{1,\dots,|\md'|\}$, if some $\node' \subseteq \md'$ with $|\node'|=k$ is generated, it will also be processed. In particular, this implies that $\md'$ will be processed, given that it is generated.

It remains to be shown that $\md'$ will be generated. To this end, observe that the root $\emptyset$ is trivially processed (see line~\ref{algoline:rbfhs:call_RBFHS'}) and must be labeled with a non-empty minimal conflict (as line~\ref{algoline:rbfhs:call_RBFHS'} was reached, see above), which entails by line~\ref{algoline:rbfhs':expand} (\textsc{expand} function) that all tree nodes of cardinality $k=1$ are generated, among them one subset $\node'$ of $\md'$. Since $\node'$ must be processed (note: maybe not immediately, but definitely at some stage of the algorithm's execution), as proven, some $\node' \cup \setof{x} \subseteq \md'$ of cardinality $k+1$ is generated. The same inductive argument can be applied to all nodes $\node' \subset \md'$. Consequently, $\md'$ will be eventually generated---and processed, as argued above. This is a contradiction to the assumption that $\md'$ is never processed, which finalizes the completeness proof.
\subsubsection{Best-First Property} 
%
%
%
%
We already know that RBF-HS is complete, i.e., that all minimal diagnoses for the given DPI will be in the returned list $\mD$. We now have to show that this list is sorted in descending order by $f$-value.
Since any node corresponding to a minimal diagnosis that is processed by RBF-HS will be (directly) added to $\mD$ by Lemma~\ref{lem:if_diag_processed_then_added_to_mD}, it suffices to demonstrate that, for any two minimal diagnoses $\md',\md''$ with $f(\md') < f(\md'')$, some node equal to $\md''$ is processed prior to all nodes equal to $\md'$.

To this end, let $\ld = \infty$ (the algorithm does not terminate before all minimal diagnoses have been found) and assume the opposite, i.e., some node corresponding to $\md'$ is processed earlier than all nodes equal to $\md''$. Take the (first ever) call RBF-HS'($\node,F(\node),\bnd$) with $\node = \md'$ (i.e., the first call that processes $\md'$). Then we have that $F(\node) \geq \bnd$ (while-condition) and $\bnd = \max\{F(\node^{1}_{\mathit{2bst}}), F(\node^{2}_{\mathit{2bst}}), \dots$, $F(\node^{k}_{\mathit{2bst}})\}$ with $k = |\md'|-1$ where $\node^{r}_{\mathit{2bst}}$ denotes the best alternative node (according to $F$-value) at tree depth $r$.
(Note that, at any time during its execution, RBF-HS' involves only one expanded node at each tree level; amongst the generated nodes at one level $r$, the best one is expanded and the second best one is precisely $\node^{r}_{\mathit{2bst}}$. To see that $\bnd$ is equal to the maximum of the stated set of best alternative nodes, observe that $\bnd = -\infty$ at the very first call of RBF-HS' in line~\ref{algoline:rbfhs:call_RBFHS'}, and for each node that is expanded, the new $\bnd$ is the maximum of the current $\bnd$ and the current best alternative node, cf.\ line~\ref{algoline:rbfhs':recursive_call}).

Now, let $\node^*$ be the deepest common ancestor node of $\md'$ and $\md''$ in the tree, i.e., $\node^* = \md' \cap \md''$. Since both $\md'$ and $\md''$ are \emph{minimal} diagnoses, 
$\node^* \subset \md'$ and $\node^* \subset \md''$.
Moreover, let $\node^*_{r,\md''}$ denote the $r$-th successor node of $\node^*$ along a path to a node equal to $\md''$. 
E.g., $\node^*_{1,\md''}$ describes the child node of $\node^*$ along the path to $\md''$; note that $\node^*_{r,\md''} = \md''$ for $r = |\md''|-|\node^*|$ and that $\node^*_{r,\md''}$ is a node at tree depth $|\node^*|+r$. 

For $s = |\node^*|+1$, we know from above ($F(\node) \geq \bnd$) that $F(\node) \geq F(\node^{s}_{2bst})$ and, since $\node^{s}_{2bst}$ is the best alternative node at level $s$, that $F(\node^{s}_{2bst}) \geq F(\node^*_{1,\md''})$. Furthermore, by Lemma~\ref{lem:invariant_F(n)<=f(n)_holds_throughout_entire_execution}, $f(\node) \geq F(\node)$ must hold. Overall, since $\node = \md'$, we so far have $f(\md') \geq F(\node^*_{1,\md''})$. If $|\md''|-|\node^*| = 1$, i.e., $\node^*_{1,\md''} = \md''$, then (*) $F(\node^*_{1,\md''}) = f(\node^*_{1,\md''}) = f(\md'')$ must be true. The reason for this is Lemma~\ref{lem:before_node_n_processed_for_first_time_F(n)=f(n)} and that no node corresponding to $\md''$ can have been processed yet, as this would be a contradiction to our assumption that we are considering the \emph{first} call that processes a node equal to $\md'$ and that 
this one is processed earlier than any node equal to $\md''$.
Thus, we have deduced that $f(\md') \geq f(\md'')$, which gives a contradiction to our assumption. 

So, let $|\md''|-|\node^*| \geq 2$, i.e., $\node^*_{1,\md''} \subset \md''$. By Lemma~\ref{lem:invariant_F(n)<=f(n)_holds_throughout_entire_execution}, there are now two cases: \emph{(a)}~$F(\node^*_{1,\md''}) = f(\node^*_{1,\md''})$, or \emph{(b)}~$F(\node^*_{1,\md''}) < f(\node^*_{1,\md''})$. 

Assume (a) first. Since $f(X) > f(Y)$ whenever $X \subset Y$ due to the fact that $f$ is strictly antimonotonic, we can 
derive that $f(\md') \geq F(\node^*_{1,\md''}) = f(\node^*_{1,\md''}) > f(\node^*_{j,\md''})$ for $j = 2,\dots,|\md''|-|\node^*|$. 
Hence, $f(\md') \geq f(\md'')$, a contradiction to our assumption.

Finally, assume (b). From Lemma~\ref{lem:before_node_n_processed_for_first_time_F(n)=f(n)}, we know that $\node^*_{1,\md''}$ must already have been processed. In addition, since $\md''$ is a minimal diagnosis and $\node^*_{1,\md''} \subset \md''$, we have that 
$\node^*_{1,\md''}$ can never be labeled $\valid$ or $\closed$ when it is processed, due to Lemma~\ref{lem:only_nodes_that_are_diags_can_be_labeled_valid_or_nonmin}. Therefore, and because $\ld = \infty$, every (and, in particular, the last) call of RBF-HS' that processed $\node^*_{1,\md''}$ must have returned in line~\ref{algoline:rbfhs':return_F(n)}. From this, we infer that $F(\node^*_{1,\md''}) = \max_{\node \in \childnodes}(F(\node))$ where $\childnodes$ refers to the child nodes of $\node^*_{1,\md''}$. Since $\node^*_{2,\md''} \subseteq \md''$ is one node among $\childnodes$, we obtain that $F(\node^*_{1,\md''}) \geq F(\node^*_{2,\md''})$. If $|\md''|-|\node^*| = 2$, i.e., $\node^*_{2,\md''} = \md''$, then the same argumentation as in (*) above can be applied to show that $f(\md') \geq f(\md'')$, a contradiction.

Otherwise, we consider $|\md''|-|\node^*| \geq 3$, i.e., $\node^*_{2,\md''} \subset \md''$, and can again discern two analogous cases (a) and (b) for $\node^*_{2,\md''}$. In this vein, we can consecutively derive $f(\md') \geq F(\node^*_{j,\md''})$ for $j = 3,\dots,|\md''|-|\node^*|$ and use (*) to obtain the contradiction.
This completes the proof of the best-first property. 


\subsubsection{Soundness}
We have to prove that every node that is added to $\mD$ is a minimal diagnosis. 
To this end, assume that some $\md' \in \mD$ is not a minimal diagnosis. That is, $\md'$ is \emph{(a)} not a diagnosis or \emph{(b)} a diagnosis, but not minimal. Suppose (a). Here we immediately get a contradiction to Lemma~\ref{lem:only_diags_can_be_added_to_mD}. 


Now, suppose (b). That is, $\md'$ is a non-minimal diagnosis, or, in other words, there is a minimal diagnosis $\md'' \subset \md'$. By the fact that $f$ is strictly antimonotonic, 
$f(\md'') > f(\md')$ must hold. Further, $\md'$ must have been added to $\mD$ in line~\ref{algoline:rbfhs:add_node_to_mD} as node $\node$ because this is the only place in RBF-HS where $\mD$ is extended. Thus, the \textsc{label} function must have been executed for $\node$, in particular lines~\ref{algoline:label:non-min_crit_start}--\ref{algoline:label:non-min_crit_end}. 
However, no return can have taken place in line~\ref{algoline:label:non-min_crit_end} due to the fact that $\node$ was assigned the label $\valid$ which implies that line~\ref{algoline:label:return_valid} must have been reached. 
As a consequence, the test $\node \supseteq \node_i$ in line~\ref{algoline:label:if_n_supseteq_n_i} must have been negative for all $\node_i \in \mD$. Hence, no node in $\mD$ is a subset of $\node = \md'$, which means that, in particular, $\md'' \notin \mD$ at the time $\md'$ is processed. Now, since $\md''$ is a minimal diagnosis and has a higher $f$-value than $\md'$, we obtain a contradiction to the completeness and best-first properties shown above. This completes the soundness proof.\qed  

\section{HBF-HS: Proof of Correctness}
\label{apx:hbfhs:correctness}

%
%

We next prove the following theorem, which is stated in Sec~\ref{sec:hbfhs_correctness}:
\vspace{3pt}\\ 
\textbf{Theorem~\ref{thm:hbfhs:correctness}}(Correctness of HBF-HS)\textbf{.}\hspace{4pt}
\emph{Let \textsc{findMinConflict} be a sound and complete method for conflict computation, i.e., given a DPI, it outputs a minimal conflict for this DPI if a minimal conflict exists, and `no conflict' otherwise. Further, let $\mathit{stop_{\mathsf{HS}}}$ be any predicate depending on the execution state of HS-Tree. Then, HBF-HS is sound, complete and best-first, i.e., it computes \emph{all} and \emph{only} minimal diagnoses \emph{in descending order} \emph{of probability} as per the strictly antimonotonic probability measure $\pr$.}\vspace{4pt}
\begin{proof}
	First, \textsc{HS-Tree} is executed (line~\ref{algoline:hbfhs:HS-Tree_call}). There are three possible reasons for the termination of \textsc{HS-Tree}: \emph{(i)}~the queue of open nodes maintained by \textsc{HS-Tree} is empty, \emph{(ii)}~$\ld$ minimal diagnoses have been computed, or \emph{(iii)}~$\mathit{stop_{\mathsf{HS}}}$ is true. In case (i), we have that $\Queue_{\mathsf{HS}} = [\,]$, which implies (due to line~\ref{algoline:hbfhs:if_mD_HS_geq_ld}) that HBF-HS will return $\mD_{\mathsf{HS}}$, the set of all minimal diagnoses computed by \textsc{HS-Tree}. By \cite[Prop.~4.15]{Rodler2015phd}, $\mD_{\mathsf{HS}}$ is the set of all minimal diagnoses for the given DPI, and, by \cite[Corollary~4.7]{Rodler2015phd}, the diagnoses in $\mD_{\mathsf{HS}}$ are sorted in descending order of their probability as per $\pr$. Hence, the theorem holds if case (i) applies.
	In case (ii), we have that $|\mD_{\mathsf{HS}}| \geq \ld$, which is why HBF-HS will return $\mD_{\mathsf{HS}}$ (due to line~\ref{algoline:hbfhs:if_mD_HS_geq_ld}). By \cite[Corollary~4.7]{Rodler2015phd}, $\mD_{\mathsf{HS}}$ comprises the $\ld$ most probable minimal diagnoses for the given DPI as per $\pr$, which are sorted by $\pr$ in descending order. Consequently, the theorem is true for case (ii).
	If case (iii) holds (and case (i), already discussed, does not hold),
	 it must be true that $\Queue_{\mathsf{HS}}$ is not empty. The reasons for this are that the node queue of \textsc{HS-Tree} comprises one node (the root node $\emptyset$) at the time \textsc{HS-Tree} starts executing (cf.\ Sec.~\ref{sec:diagnosis_search_algos}), that \textsc{HS-Tree} terminates as soon as the node queue becomes empty, and that case (i) above does not hold by assumption. Therefore, the condition $\Queue_{\mathsf{HS}} = [\,]$ in line~\ref{algoline:hbfhs:if_mD_HS_geq_ld} is false. Hence, the other condition, $\mD_{\mathsf{HS}} \geq \ld$, in line~\ref{algoline:hbfhs:if_mD_HS_geq_ld}, determines whether line~\ref{algoline:hbfhs:return_mD_HS} will be executed. If this condition is true, then case~(ii) applies and the theorem holds, as argued above. 
	Otherwise, the algorithm continues its execution after line~\ref{algoline:hbfhs:return_mD_HS},
	which means that RBF-HS' is called given the virtual root node $\node_0$ created in line~\ref{algoline:hbfhs:make_virtual_root}. Note that the same $\bnd := -\infty$ is used for the root node as in the original RBF-HS algorithm (cf.\ line~\ref{algoline:rbfhs:call_RBFHS'} in Alg.~\ref{algo:RBF_HS}), and that the f-value of the root, set as $f(\node_0) := 0$ in this case, is irrelevant to the correctness of RBF-HS (cf.\ \ref{apx:proof}). The only differences to the original RBF-HS algorithm are that RBF-HS' within HBF-HS starts with an initially given
	\begin{enumerate}[noitemsep,label=(\arabic*),topsep=0pt]
		\item \label{enum:proof:hbfhs:diff_mD} (possibly empty) collection $\mD := \mD_{\mathsf{HS}}$ of minimal diagnoses (line~\ref{algoline:hbfhs:initialize_mD_mC_for_RBF-HS}),
		\item \label{enum:proof:hbfhs:diff_mC} (possibly empty) collection $\mC := \mC_{\mathsf{HS}}$ of minimal conflicts (line~\ref{algoline:hbfhs:initialize_mD_mC_for_RBF-HS}), and
		\item \label{enum:proof:hbfhs:diff_mQ} (non-empty) collection $\childnodes_{\mathsf{root}}$ of child nodes of the root $\node_0$ (lines~\ref{algoline:hbfhs:comment_on_label_function} and \ref{algoline:hbfhs:expand:if_isSet}--\ref{algoline:hbfhs:expand:add_succ_of_virtual_root}).
	\end{enumerate}
We next argue why these differences do not affect the correctness of RBF-HS'.
	
	Ad \ref{enum:proof:hbfhs:diff_mD}: First, $\mD$ is correct in that it comprises the $|\mD|$ most probable minimal diagnoses as per $\pr$ due to \cite[Corollary~4.7]{Rodler2015phd}. Second, a non-empty collection $\mD$ given from the outset does not harm the correctness of RBF-HS'. To see this, observe that, beside allowing the check if $\ld$ diagnoses have been computed and the algorithm can stop (line~\ref{algoline:rbfhs':if_mD_geq_ld} in Alg.~\ref{algo:RBF_HS}), 	
	the only function of $\mD$ in RBF-HS' is the detection and closing of redundantly computed or non-minimal diagnoses (lines~\ref{algoline:label:non-min_crit_start}--\ref{algoline:label:non-min_crit_end} in Alg.~\ref{algo:RBF_HS}). More specifically, no node can be wrongly closed since each element of $\mD$ is a minimal diagnosis; and,  
	every node that must be closed because it is a redundant minimal diagnosis or a non-minimal diagnosis will actually be closed due to the strict antimonotonicity of $\pr$ (which implies that minimal diagnoses must be found before non-minimal ones).	
	
	Ad \ref{enum:proof:hbfhs:diff_mC}: First, $\mC$ is correct in that it contains only minimal conflicts, 
	since \textsc{HS-Tree} uses \textsc{findMinConflict} for conflict computation which is sound and complete by assumption. Second, a non-empty set $\mC$ given from the beginning does not affect the correctness of RBF-HS'. To realize this, note that the only function of $\mC$ in RBF-HS' is the avoidance of redundant conflict computations (lines~\ref{algoline:label:reuse_start}--\ref{algoline:label:reuse_end} in Alg.~\ref{algo:RBF_HS}).
	
	Ad \ref{enum:proof:hbfhs:diff_mQ}: By line~\ref{algoline:hbfhs:delete_duplicates}, $\childnodes_{\mathsf{root}}$ corresponds exactly to the open nodes $\Queue_{\mathsf{HS}}$ returned by \textsc{HS-Tree} when it was stopped due to $\mathit{stop_{\mathsf{HS}}}$, after all duplicate nodes have been deleted from $\Queue_{\mathsf{HS}}$. Due to the completeness of \textsc{HS-Tree} (\cite[Prop.~4.15, Lemma~4.18]{Rodler2015phd}), $\Queue_{\mathsf{HS}}$ must, for each minimal diagnosis $\md$ not already in $\mD_{\mathsf{HS}}$, include some node that is a subset of $\md$. Hence, all remaining minimal diagnoses not in $\mD_{\mathsf{HS}}$ can be constructed by extending the nodes in $\Queue_{\mathsf{HS}}$. Note also that neither the soundness nor the completeness of RBF-HS' is harmed by the deletion of duplicates (i.e., set-equal nodes) from $\Queue_{\mathsf{HS}}$ because RBF-HS' treats nodes as sets. Consequently, starting with $\childnodes_{\mathsf{root}}$ as the initial nodes, RBF-HS' can find all minimal diagnoses not in $\mD_{\mathsf{HS}}$. Finally, observe 
	that the correctness proofs of RBF-HS (cf.\ \ref{apx:proof}) do not make any assumptions about the size $|\node|$ of the nodes $\node$ in $\childnodes$. Hence, the possible inclusion of non-singleton nodes in $\childnodes_{\mathsf{root}}$ does not affect the correctness of RBF-HS'. This completes the proof.
\end{proof}
\section{Related Works in the Heuristic Search Domain}
\label{apx:related_works_in_heuristic_search}
In this appendix, we discuss \emph{general} memory-limited search algorithms that are related to RBFS (and thus to RBF-HS), 
aim at improving RBFS (and thus might be used to improve RBF-HS), or 
are conceptionally related to HBF-HS.

\subsection{Memory-Limited Search Algorithms}
Beside RBFS, there is a range of alternative linear-space heuristic search techniques. 
Some examples are  
IDA* \cite{korf1985depth},
MREC \cite{sen1989fast},
MA* \cite{chakrabarti1989heuristic}, 
DFS* \cite{rao1991depth},
IDA*-CR \cite{sarkar1991reducing},
MIDA* \cite{Wah1991MIDAAI},
ITS \cite{nau1992efficient},
IE \cite{russell1992efficient}, and
SMA* \cite{russell1992efficient}.  
In contrast to RBFS, these algorithms generally do not expand nodes in best-first order if the given cost function is non-monotonic. This property, however, does not pose a problem in the hitting set computation scenario. The reason for this is that the cost function in hitting set search \emph{has to be} anti-monotonic (cf.\ \ref{enum:diff:stricter_conditions_on_cost_function} in Sec.~\ref{sec:search}) to find solutions in
best-first order.
Recall that
anti-monotonicity (for maximal-cost solutions) in hitting set search is the equivalent to monotonicity (for minimal-cost solutions) in classic heuristic search.
%
Hence, in principle, any of these algorithms could have been used as a basis for this work, i.e., for being ``translated'' to a hitting set version. 

The causes for choosing RBFS as a foundation for our presented algorithms are twofold: First, RBFS is particularly well-understood and covered by a rich collection of literature including both theoretical and empirical analyses of the algorithm. Second, and more importantly, RBFS is asymptotically optimal\footnote{An algorithm $\mathsf{algo}$ is called \emph{asymptotically optimal} for some problem class $C$ iff it is (for the problem size $n$ growing to infinity) not more than a constant factor worse than the best achievable running time $\mathsf{best}$ 
	on problems of class $C$. Formally: 
	$\mathit{time}(\mathsf{algo},C) \in O(\mathit{time}(\mathsf{best},C))$.}, requiring $O(b^d)$ time ($b$ branching factor, $d$ maximal search tree depth) 
when being used for minimum-cardinality diagnosis computation \cite{korf1992linear}, 
one of the most central and fundamental problems in model-based diagnosis. 

Compared to IDA* (cf.\ Example~\ref{ex:search_algorithms}),
which is the most prominent\footnote{When we judge ``prominence'' by the citation tally on Google Scholar (as of April 2020).} linear best-first search algorithm and also asymptotically optimal for minimum-cardinality hitting set search, RBFS exhibits a better practical (empirical) time complexity
\cite{zhang1995performance} (note, the theoretical number of expanded nodes is $O(n^2)$ for both algorithms where $n$ is the number of nodes expanded by A* \cite{hatem2014bounded}). 
This can be intuitively explained by the fact that RBFS, unlike IDA*, does not discard the entire search tree between any two iterations \cite{korf1992linear}.
This runtime advantage of RBFS over IDA* holds especially when the cost for node expansion is high \cite{hatem2014bounded}. This is absolutely the case in diagnosis search 
where node expansion requires a conflict, which must either be sought in a maintained list of conflicts (reuse case) or must be newly generated using expensive theorem proving (computation case), see the \textsc{label} function in Alg.~\ref{algo:RBF_HS}. This is why RBFS appeared to be a more appropriate base for constructing a hitting set search than IDA*. 

Finally, there is CDA* \cite{williams2007conflict}, a version of A*, originally proposed for solving optimal constraint satisfaction problems, which is also employable for diagnosis search.
It incorporates an any-space algorithm that generates the most preferred diagnoses first. The two important differences to RBF-HS are that CDA* is not black-box, i.e., appears to be not as flexibly usable with arbitrary logics and reasoners as RBF-HS, and that CDA* is generally incomplete \cite{stern2012}.

\subsection{Works towards Improving RBFS}
The price to pay for the guaranteed linearity of RBFS in terms of space consumption is that nodes have to be forgotten each time a backtracking step is made. Whenever an already explored subtree becomes attractive again (because all other better subtrees have been explored), it will be re-examined. This scheme results in a potentially large number of node re-explorations. In the worst case, when every node has a unique $f$-value and the node with next-best $f$-value is always located in a different subtree of the root,
$O(n^2)$ nodes will be expanded 
where $n$ is the number of nodes A* would expand \cite{hatem2015recursive}.
Addressing this problem, \cite{hatem2015recursive} have proposed three techniques for controlling the overhead caused by excessive backtracking in RBFS, at the cost of generating suboptimal solutions in general. These techniques are called RBFS$_\epsilon$, RBFS$_{\mathit{kthrt}}$ and RBFS$_{\mathsf{CR}}$. The idea of RBFS$_\epsilon$ is to allow the algorithm to explore a little (as ruled by the choice of the parameter $\epsilon$) further than suggested by $\bnd$, i.e., $\bnd$ in line~\ref{algoline:rbfs':while} of Alg.~\ref{algo:RBFS} is replaced by $\bnd + \epsilon$. While this slight change yields good results in practice under an adequate setting of $\epsilon$, it does not lower the quadratic worst-case time complexity. RBFS$_{\mathit{kthrt}}$ goes one step further by loosening both $\bnd$ and the $f$-function, thereby achieving fewer backtrackings \emph{and} fewer node expansions, albeit still without theoretical time complexity savings. Finally, RBFS$_{\mathsf{CR}}$ adopts a concept originally introduced by \cite{sarkar1991reducing} for IDA* in order to reduce re-expansions. The idea is to track the distribution of $f$-values under each node along the currently explored path, which allows to adapt the backed-up $F$-value in a way it can be guaranteed that, each time a node is re-explored, twice as many successor nodes will be investigated than when this node was last explored. In this vein, the number of explored nodes can be shown to be in $O(n)$, i.e., asymptotically maximally by a constant worse than for A*.

All of these three techniques are applicable to RBF-HS as well, and we expect the (positive) implications on practical  performance in the hitting-set case to be in line with what was observed in \cite{hatem2015recursive} for classic search problems. A clear shortcoming of such an approach, however, will be the potential non-minimality of the returned diagnoses (unsoundness) and the potential violation of the preference order on the output diagnoses (best-first property not given). Whereas the soundness problem can be taken care of by postprocessing the returned diagnoses, e.g., by means of Inv-QX \cite{Shchekotykhin2014}, it is not straightforward how to handle the best-first violation, i.e., how to ensure that the returned collection $\mD$ includes exactly the $|\mD|$ best diagnoses. Both the implementation of these suboptimal RBF-HS variants as well as the study of this latter question will be part of our future work.
%

If only one solution is demanded, i.e., only the single most probable or single minimum-cardinality diagnosis is to be found, then techniques discussed in \cite{hansen2007anytime} can be applied to RBF-HS. However, this is useful only if a reasonable heuristic function (for non-additive, probabilistic costs) can be expressed for hitting set searches, which is to date still an open problem.

\subsection{Works Related to HBF-HS} 
HBF-HS follows a similar principle for RBF-HS as MREC \cite{sen1989fast} does for IDA*.  
MREC 
balances the time and space complexity by a 
parameter that 
rules
how much memory is available for use by the algorithm. In the same way that MREC behaves equally to IDA* for minimal available memory and similarly to A* for a large amount of conceded memory, HBF-HS resembles RBF-HS and HS-Tree in these two cases. 

Two other strategies that attempt to optimally exploit and exhaust the available memory in order to increase search speed are MA* \cite{chakrabarti1989heuristic} and SMA* \cite{russell1992efficient}. Their underlying principle is to store every node until the memory limit is reached, and to then purge the least promising node(s) in order to make room for the next node to be explored. Whenever the search problem is solvable within the given amount of memory, these algorithms will not run out of memory and return a best solution. Theoretically, this property cannot be proven for HBF-HS as it acts like RBF-HS from the point where the (memory-dependent) switch criterion is triggered. In other words, if the switch takes place too late (such that very little memory remains which cannot hold the linear number of nodes additionally explored after the switch), then HBF-HS can run out of memory. However, first, we observed in our experiments (cf.\ Sec.~\ref{sec:experiment_results}) that the number of additional nodes stored by HBF-HS after the switch was always minor (small single digit percentage) relative to those
generated before the switch took place.
Second, (S)MA*'s concept of on-demand node pruning can be integrated into HBF-HS as well in order to resolve this problem. Still, as a future work, we plan to carry over 
these algorithms to the diagnosis domain as well, and to study their hitting set versions. 

Moreover, \cite{russell1992efficient} suggested the IE algorithm, which however behaves the same as RBFS for a monotonic cost-function (and thus, a hitting set version of it would act identically to RBF-HS, cf.\ bullet~\ref{enum:diff:stricter_conditions_on_cost_function} on page~\pageref{enum:diff:stricter_conditions_on_cost_function}).
\section{Evaluation Results for the Single-Run Use Case}
\label{apx:eval_results_single-run-case}

\noindent \emph{Experiment settings (single-run use case):}
We ran one diagnosis search using each algorithm among RBF-HS and HS-Tree, for each DPI from Tab.~\ref{tab:dataset}, for each probability setting among maxProb and minCard, and for each number of diagnoses $\ld \in \{2,6,10,20\}$ to be computed (cf.\ Sec.~\ref{sec:experiment_settings}).

\begin{figure}[t]
	\includegraphics[width=1\columnwidth]{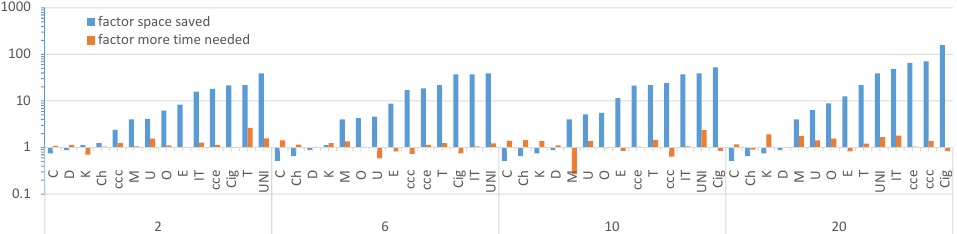}
	\caption{Results (normal tests) for the single-run experiment (RBF-HS vs.\ HS-Tree): x-axis shows ontologies from Table~\ref{tab:dataset} and number of computed diagnoses $\ld \in \{2,6,10,20\}$. Per setting of $\ld$, the ontologies along the x-axis are sorted from low to high space savings achieved by RBF-HS (blue bars).}
	\label{fig:results_card_single_run}
\end{figure}

\begin{figure}[t]
	\centering
	\includegraphics[width=0.42\columnwidth]{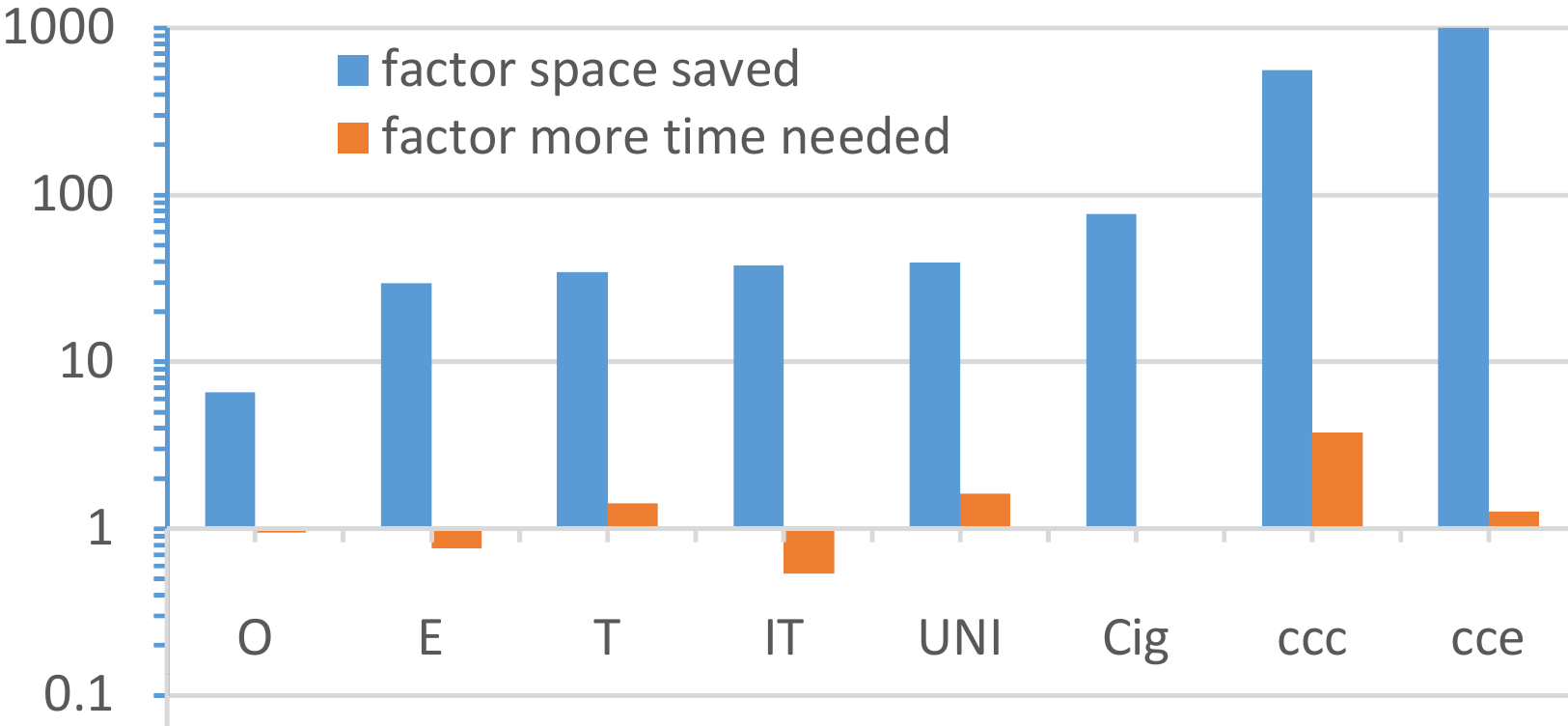}
	\caption{Results (scalability tests) for the single-run experiment (RBF-HS vs.\ HS-Tree) with setting $\ld:=100$: x-axis shows ontologies from Table~\ref{tab:dataset}. 
		The ontologies along the x-axis are sorted from low to high space savings achieved by RBF-HS (blue bars).}
	\label{fig:results_card_single_run_scalability}
\end{figure}

\noindent \emph{Presentation of the results:} The results for the single-run tests under the minCard setting, i.e., when diagnoses are computed in the order from low to high cardinality, are shown by Fig.~\ref{fig:results_card_single_run} (normal experiment) and Fig.~\ref{fig:results_card_single_run_scalability} (scalability experiment). The figures compare the runtime and memory consumption we measured for RBF-HS and HS-Tree, as described in Sec.~\ref{sec:presentation_of_results}. Regarding the absolute runtime and memory expenditure (not displayed in the figures), we measured a min / avg / max runtime of 0.003 / 0.7 / 17 sec and a min / avg / max space consumption of 3 / 560 / 37.5K tree nodes.

\noindent \emph{Normal experiments (Fig.~\ref{fig:results_card_single_run}):} In the normal experiments, we observed a positive space-time tradeoff in three out of four cases; i.e., the amount of memory savings of RBF-HS in these cases exceeded the amount of its additional time required compared to HS-Tree. As a closer analysis revealed, all scenarios where the space-time tradeoff was not positive were easy cases involving diagnoses of cardinality one that lead to very small 
search trees where a depth-first strategy (RBF-HS) can require more space than a breadth-first one (HS-Tree).
%
In most cases, space savings achieved by RBF-HS were significant; they averaged to a factor of 18, reached factors larger than 10 / 20 / 30 in 45\,\% / 33\,\% / 21\,\% of the scenarios, and attained a maximum factor of 158. Time overheads of RBF-HS, in contrast, remained below a factor of 2 in more than 96\,\% of the scenarios, and yielded an average and maximum factor of 1.2 and 2.7, respectively. In 23\,\% of the scenarios, we even registered savings of RBF-HS wrt.\ \emph{both} time \emph{and} space. When computing diagnoses 
in the order from low to high probability, 
on the other hand, the space-time tradeoff was by and large neutral, i.e., RBF-HS traded $k$ times less space for approximately $k$ times more time in most cases. In numbers, we found that RBF-HS's space-time tradeoff was positive in 53\,\% of the cases, and the median factors of space saved and more time required by RBF-HS versus HS-Tree were $3.75$ and $3.56$, respectively. In other words, RBF-HS saved a median of 73\,\% of memory against HS-Tree, whereas the latter saved a median of 72\,\% of computation time compared to the former.
We discuss the reasons for these performance differences depending on the setting minCard vs.\ maxProb in Sec.~\ref{sec:experiment_results}.

\noindent \emph{Scalability experiments (Fig.~\ref{fig:results_card_single_run_scalability}):} For the minCard setting, we observed high space savings of RBF-HS against HS-Tree in all cases, with a median of 97.4\,\% and a maximum of 99.8\,\%. Regarding computation time, RBF-HS was even equally fast or faster (up to 46\,\%) than HS-Tree in four cases, whereas HS-Tree was (up to 74\,\%) faster in the other four cases. For the maxProb setting, we find that RBF-HS generally does not scale to three-digit numbers of computed diagnoses, which results from a well-known property of the original RBFS algorithm (cf.\ discussion in Sec.~\ref{sec:experiment_results}). Only in two cases (UNI, M), we could observe a positive space-time tradeoff of RBF-HS versus HS-Tree. The median factors of space saved and more time required by RBF-HS as compared to HS-Tree were $20$ and $79$, respectively. Hence, it is recommendable to use HBF-HS if the computation of a larger number of diagnoses is required (cf.\ Sec.\ref{sec:experiment_results}). 


\end{document}